\documentclass[manuscript,screen]{acmart}

\AtBeginDocument{%
  \providecommand\BibTeX{{%
    \normalfont B\kern-0.5em{\scshape i\kern-0.25em b}\kern-0.8em\TeX}}}

\setcopyright{acmcopyright}
\copyrightyear{2018}
\acmYear{2018}
\acmDOI{XXXXXXX.XXXXXXX}

\acmConference[Conference acronym 'XX]{Make sure to enter the correct
  conference title from your rights confirmation emai}{June 03--05,
  2018}{Woodstock, NY}
\acmPrice{15.00}
\acmISBN{978-1-4503-XXXX-X/18/06}

\usepackage[utf8]{inputenc} 
\usepackage[T1]{fontenc}    
\usepackage{hyperref}       
\usepackage{url}            
\usepackage{booktabs}       
\usepackage{amsfonts}       
\usepackage{nicefrac}       
\usepackage{microtype}      

\usepackage{amsthm}
\usepackage{amsmath}
\usepackage{cleveref}

\usepackage[figuresright]{rotating}
\usepackage{caption}
\usepackage{subcaption}
\usepackage[ruled,vlined]{algorithm2e}

\usepackage{xcolor}         
\usepackage{float}

\usepackage{graphics}
\usepackage{subcaption}
\usepackage{multirow}
\usepackage{ctable}

\usepackage{xcolor}
\usepackage{textcomp}
\usepackage{multicol}

\definecolor{blueannoback}{RGB}{234,242,250}
\definecolor{greenannoback}{RGB}{230,244,214}
\definecolor{redannoback}{RGB}{255, 230, 230}
\definecolor{yellowannoback}{RGB}{252, 251, 202}
\definecolor{orangeannoback}{RGB}{247, 189, 114}
\definecolor{pinkannoback}{RGB}{238, 172, 252}

\usepackage{listings}

\lstdefinelanguage{prompt}{
    morecomment=[l][\textcolor{BurntOrange}]{@}
}

\lstdefinestyle{prompt-style}{
    language=prompt,
    escapeinside={\%*}{*)}
}

\newcommand{\lpm}{FM}  

\newcommand{\llm}{{LLM}}  

\newcommand{\gpttwo}{{GPT2}}
\newcommand{\gpttwom}{{GPT2-Medium}}
\newcommand{\gpttwol}{{GPT2-Large}}
\newcommand{\gpttwoxl}{{GPT2-XL}}
\newcommand{\gptthree}{{GPT-3}}
\newcommand{\instructgpt}{{InstructGPT}}
\newcommand{\chatgpt}{{ChatGPT}}
\newcommand{\chatgptr}{{\chatgpt~ (Raw.)}}
\newcommand{\chatgptc}{{\chatgpt~ (Con.)}}

\newcommand{\alexnet}{{AlexNet}}
\newcommand{\rsnet}{{ResNet}}
\newcommand{\dsnet}{{DenseNet}}

\newcommand{\openclip}{{OpenCLIP}}
\newcommand{\openclipl}{{\openclip-L}}
\newcommand{\openclipb}{{\openclip-B}}

\newcommand{\blip}{{BLIP}}
\newcommand{\flamingo}{{OpenFlamingo-9B}}

\newcommand{\vgap}{\vspace{-0.3cm}}  

\newcommand{\svs}{{SingaporeSVI579}}  
\newcommand{\aid}{{AID}} 
\newcommand{\ori}{(Origin)}
\newcommand{\chg}{(Updated)}
\newcommand{\poidata}{{UrbanPOI5K}}

\newcommand{\placevec}{Place2Vec}
\newcommand{\hgi}{HGI}

\begin{document}

\title[Foundation Models for GeoAI]{On the Opportunities and Challenges of Foundation Models for Geospatial Artificial Intelligence}

\author{Gengchen Mai}
\email{gengchen.mai25@uga.edu}
\orcid{0000-0002-7818-7309}
\affiliation{%
  \institution{SEAI Lab, Department of Geography, University of Georgia}
  \city{Athens}
  \state{Georgia}
  \country{USA}
  \postcode{30605}
}

\author{Weiming Huang}
\email{weiming.huang@ntu.edu.sg}
\orcid{0000-0002-3208-4208}
\affiliation{%
  \institution{School of Computer Science and Engineering, Nanyang Technological University}
  \city{}
  \state{}
  \country{Singapore}
 }

\author{Jin Sun}
\email{jinsun@uga.edu}
\affiliation{%
  \institution{School of Computing, University of Georgia}
  \city{Athens}
  \state{Georgia}
  \country{USA}
 }
 
\author{Suhang Song}
\email{suhang.song@uga.edu}
\orcid{0000-0003-1934-632X}
\affiliation{%
  \institution{College of Public Health, University of Georgia}
  \city{Athens}
  \state{Georgia}
  \country{USA}
 }

 \author{Deepak Mishra}
\email{dmishra@uga.edu}
\orcid{0000-0001-8192-7681}
\affiliation{%
  \institution{Department of Geography, University of Georgia}
  \city{Athens}
  \state{Georgia}
  \country{USA}
 }

 \author{Ninghao Liu}
\email{ninghao.liu@uga.edu}
\orcid{0000-0002-9170-2424}
\affiliation{%
  \institution{School of Computing, University of Georgia}
  \city{Athens}
  \state{Georgia}
  \country{USA}
 }

 \author{Song Gao}
\email{song.gao@wisc.edu}
\orcid{0000-0003-4359-6302}
\affiliation{%
  \institution{Geospatial Data Science Lab, Department of Geography, University of Wisconsin-Madison}
  \city{Madison}
  \state{Wisconsin}
  \country{USA}
 }

 \author{Tianming Liu}
\email{tliu@uga.edu}
\orcid{0000-0002-8132-9048}
\affiliation{%
  \institution{School of Computing, University of Georgia}
  \city{Athens}
  \state{Georgia}
  \country{USA}
 }

 \author{Gao Cong}
\email{gaocong@ntu.edu.sg}
\orcid{0000-0002-4430-6373}
\affiliation{%
  \institution{School of Computer Science and Engineering, Nanyang Technological University}
  \city{}
  \state{}
  \country{Singapore}
 }
 
\author{Yingjie Hu}
\email{yhu42@buffalo.edu}
\orcid{0000-0002-5515-4125}
\affiliation{%
	\institution{GeoAI Lab, Department of Geography, University at Buffalo}
	\city{Buffalo}
      \state{New York}
      \country{USA}
}

\author{Chris Cundy}
\email{cundy@cs.stanford.edu}
\orcid{0000-0002-4608-4110}
\affiliation{%
  \institution{Department of Computer Science, Stanford University}
  \city{Stanford}
  \state{California}
  \country{USA}
 }

\author{Ziyuan Li}
\email{ziyuan.2.li@uconn.edu}
\orcid{0009-0001-1835-3336}
\affiliation{%
  \institution{School of Business, University of Connecticut}
  \city{Storrs}
  \state{Connecticut}
  \country{USA}
 }

 \author{Rui Zhu}
\email{rui.zhu@bristol.ac.uk}
\orcid{0000-0002-8910-9445}
\affiliation{%
	\institution{School of Geographical Sciences, University of Bristol}
	\city{Bristol}
      \country{United Kingdom}
}

\author{Ni Lao}
\email{nlao@google.com}
\orcid{0000-0002-4034-7784}
\affiliation{%
  \institution{Google}
  \city{Mountain View}
  \state{California}
  \country{USA}
 }

\renewcommand{\shortauthors}{Mai et al.}

\begin{abstract}

Large pre-trained models, also known as \emph{foundation models} ({\lpm}s), are trained in a task-agnostic manner on large-scale data and can be adapted to a wide range of downstream tasks by fine-tuning, few-shot, or even zero-shot learning. Despite their successes in language and vision tasks, we have yet seen an attempt to develop foundation models for geospatial artificial intelligence (GeoAI). In this work, we explore the promises and challenges of developing multimodal foundation models for GeoAI. We first investigate the potential of many existing {\lpm}s by testing their performances on seven tasks across multiple geospatial subdomains including Geospatial Semantics, Health Geography, Urban Geography, and Remote Sensing. Our results indicate that on several geospatial tasks that only involve text modality such as toponym recognition, location description recognition, and US state-level/county-level dementia time series forecasting, these task-agnostic {\llm}s can outperform task-specific fully-supervised models in a zero-shot or few-shot learning setting. However, on other geospatial tasks, especially tasks that involve multiple data modalities (e.g., POI-based urban function classification, street view image-based urban noise intensity classification, and remote sensing image scene classification), existing foundation models still underperform task-specific models. Based on these observations, we propose that one of the major challenges of developing a {\lpm} for GeoAI is to address the multimodality nature of geospatial tasks. After discussing the distinct challenges of each geospatial data modality, we suggest the possibility of a multimodal foundation model which can reason over various types of geospatial data through geospatial alignments. We conclude this paper by discussing the unique risks and challenges to develop such a model for GeoAI.

\end{abstract}



\received{20 February 2007}
\received[revised]{12 March 2009}
\received[accepted]{5 June 2009}

\maketitle

\vgap
\section{Introduction} \label{sec:intro}

Recent trends in machine learning (ML) and artificial intelligence (AI) speak to the unbridled powers of data and computing. Extremely large models trained on Internet-scale datasets have achieved state-of-the-art (SOTA) performance on a diverse range of learning tasks.
In particular, their unprecedented success has spurred a \emph{paradigm shift} in the way that modernday ML models are trained. 
Rather than learning task-specific models from scratch \cite{gritta2018camcoder,wang2020neurotpr,lam2018xview}, such pre-trained models (so-called ``foundation models ({\lpm}s)'' \cite{bommasani2021opportunities}) are \emph{adapted} via fine-tuning or few-shot/zero-shot learning strategies and subsequently deployed on a wide range of domains \cite{brown2020gpt3,radford2021clip}.  
Such {\lpm}s 
allow for the transfer and sharing of knowledge across domains, and
mitigate the need for task-specific training data. 
Examples of foundation models are 1) large language models ($\llm$) such as PaLM \cite{wei2022palm}, LLAMA \cite{touvron2023llama}, GPT-3 \cite{brown2020gpt3}, InstrucGPT \cite{ouyang2022instructgpt}, and ChatGPT \cite{openai2022chatgpt}; 2) large vision foundation models such as Imagen \cite{saharia2022imagen}, Stable Diffusion \cite{rombach2022stablediffusion}, DALL$\cdot$E2 \cite{ramesh2022dalle2}, and SAM \cite{kirillov2023sam}; 3) large multimodal foundation models such as CLIP \cite{radford2021clip}, OpenCLIP \cite{gabriel2021openclip}, BLIP \cite{li2022blip},
  OpenFlamingo \cite{awadalla2023OpenFlamingo}, KOSMOS-1 \cite{huang2023kosmos1}, and GPT-4 \cite{openai2023gpt4}; and 4) large reinforcement learning foundation models such as Gato \cite{reed2022gato}.

Despite their successes, there exists very little work exploring the development of an analogous foundational model for geospatial artificial intelligence (GeoAI), which lies at the intersection of geospatial scientific discoveries and AI technologies~\citep{janowicz2020geoai, goodchild2021replication, mai2022symbolic}.
The key technical challenge here is the inherently \emph{multimodal} nature of GeoAI.
The core data modalities in GeoAI include text, images (e.g., remote sensing or street view images), 
trajectory data, 
knowledge graphs, and geospatial vector data (e.g., map layers from OpenStreetMap), all of which contain important geospatial information (e.g., geometric and semantic information).
Each modality exhibits special structures that require its own unique representation -- effectively combining all these representations with appropriate inductive biases in a single model requires careful design.
The \emph{multimodal} nature of GeoAI hinders a straightforward application of existing pre-trained 
{\lpm}s across all GeoAI tasks.

In this paper, we lay the groundwork for developing {\lpm}s for GeoAI. 
We begin by providing a brief overview of existing foundation models in Section \ref{sec:related}. 
Then in Section \ref{sec:exp}, we investigate the potential of existing {\lpm}s for GeoAI by systematically comparing
the performances of several popular foundation models with many state-of-the-art fully supervised task-specific machine learning or deep learning models on various tasks from different geospatial domains: 1) \textbf{Geospatial Semantics}: toponym recognition and location description recognition task; 2) \textbf{Health Geography}: US state-level and county-level dementia death count time series forecasting task; 3) \textbf{Urban Geography}: Point-of-interest (POI) based urban function classification task and street-level image-based noise intensity classification task; 4) \textbf{Remote Sensing}: remote sensing (RS) image scene classification task. 
The advantages and problems of {\lpm} on different geospatial tasks are discussed accordingly.
Next, in Section \ref{sec:multimodal}, we detail the challenges involved in developing {\lpm}s for GeoAI. 
Creating one single {\lpm} for all GeoAI data modalities can be a daunting task. To address this, we start this discussion by examining each data modality used in GeoAI tasks.
Then, we propose our vision for a novel multimodal {\lpm} framework for GeoAI that tackles the aforementioned challenges.
Finally, we highlight some potential risks and challenges that should be considered when developing
such general-purpose models for GeoAI in Section \ref{sec:risk} and conclude this paper in Section \ref{sec:conclude}. 

Our contributions can be summarized as follows:
\begin{itemize}
    \item To the best of our knowledge, this is the first work that systematically examines the effectiveness and problems of various existing cutting-edge foundation models on different geospatial tasks across multiple geoscience domains\footnote{This work is a significant extension of our previous 4-page vision paper published in ACM SIGPATIAL 2022 \cite{mai2022geoaifm} by adding five additional tasks in Health Geography, Urban Geography, and Remote Sensing domains.}. We establish various FM baselines on seven geospatial tasks for future Geospatial Artificial General Intelligence (GeoAGI) research.
    \item We discuss the challenges of developing a multimodal foundation model for GeoAI and provide a promising framework to achieve this goal.
    \item We discuss the risks and challenges that need to be taken into account during the development and evaluation process of the multimodal geo-foundation model.
\end{itemize}

\section{Related Work}  \label{sec:related}

\subsection{Language Foundation Model}

In less than a decade, computational natural language capabilities have been completely revolutionized \cite{peters2018elmo,kenton2019bert, raffel2020t5, brown2020gpt3} by large-scale language modeling ({\llm}s).  
Language modeling \cite{jurafsky2009intro} is the simple task of predicting the next token in a sequence given previous tokens\footnote{There is also a different variant which predicts masked spans in text \cite{kenton2019bert, raffel2020t5}.}, and it corresponds to a self-supervised objective in the sense that no human labeling is needed besides a natural text corpus. When applied to vast corpora such as documents of diverse topics from the internet, {\llm}s gain significant language understanding and generation capabilities.  
Various transfer-learning and scaling studies \cite{kaplan2020scaling, hernandez2021scaling, hoffmann2022optimal} have demonstrated an almost linear relationship between downstream task quality and the log sizes of self-supervised model and data. Combined with the ever-increasing availability of data and computing, language modeling has become a reliable approach for developing increasingly powerful models.

Representative examples of these {\llm}s are the OpenAI GPTs \cite{radford2018gpt,radford2019gpt2,brown2020gpt3,openai2023gpt4}. By pretraining from vast amounts of Web data, the GPT models gain knowledge of almost all domains on the Web, which can be leveraged to solve problems of diverse verticals \cite{brown2020gpt3}.
The interfaces to access such knowledge have become increasingly simple and intuitive -- ranging from supervised fine-tuning with labeled data \cite{radford2018gpt, radford2019gpt2}, to few-shot learning \cite{brown2020gpt3} and instructions \cite{ouyang2022instructgpt}, to conversation \cite{openai2022chatgpt} and multimodality  \cite{openai2023gpt4}. In this study, we provide a comprehensive analysis of the potentials and limitations of GPT and other LLMs when applied to different geospatial domains.

\subsection{Vision Foundation Model}
Computer vision has long been dominated by task-specific models: for example, YOLO \cite{yolo} for object detection, Detectron \cite{wu2019detectron2} for instance segmentation, and SRGAN \cite{srgan} for image super-resolution. The newest example is Meta AI's Segment Anything Model (SAM) \cite{kirillov2023sam}, which is designed for interactive object segmentation.
ResNet \cite{resnet} trained on ImageNet \cite{deng2009imagenet} has been used as the backbone feature extractor for many such tasks. It can be seen as the early form of a vision foundation model.

Inspired by the great success of language foundation models, the computer vision community builds large-scale vision foundation models that can be adapted to any vision task.
The most direct adoption of the idea from language models in computer vision is the image generation models.
Since the dominance of Generative Adversarial Networks \cite{goodfellow2020generative,karras2019style}, the quality of image generation models has seen a major breakthrough via the development of diffusion-based models \cite{ho2020ddpm}.
Imagen \cite{saharia2022imagen} builds on large transformer-based language models to understand text prompts and generates high-fidelity images using diffusion models.
DALL-E$\cdot$2 \cite{ramesh2022dalle2} trains a diffusion decoder to invert an image encoder from visual-language models such as CLIP. After pre-training, it is able to generate images of various styles and characteristics.
Stable Diffusion \cite{rombach2022stablediffusion} uses a Variational Autoencoder (VAE) \cite{Kingma2014vae} to convert raw images from pixel space to latent space where the diffusion processes are more manageable and stable. It has shown great flexibility in conditioning over text, pose, edge maps, semantic maps, and scene depths \cite{zhang2023controlnet}. 
GigaGAN \cite{kang2023gigagan}, on the other hand, is a recent attempt of scaling up GAN models.

Vision-Transformer (ViT) \cite{dosovitskiy2020vit} is a widely used architecture in vision foundation models. Large-scale ViT has been developed to scale up the model \cite{zhai2022scaling}. The Swin Transformer \cite{liu2021swin} model is designed to handle the unique challenges of adapting regular transformer models with various spatial resolutions in images. Other large-scale non-transformer models are also developed to reach the same level of performance: 
ConvNext \cite{liu2022convnext} is the ``modernized'' version of convolutional neural networks that has a large number of parameters and shows a similar level of performance as Swin Transformers.
MLP-mixer \cite{tolstikhin2021mixer} is an architecture that utilizes only multi-layer perceptrons on image data. It shows competitive scores on image classification datasets.

\subsection{Multimodal Foundation Model}
Developing artificial intelligence models that are capable of performing multimodal reasoning and understanding on complex data is a promising idea. Humans naturally perform multimodal reasoning in daily life \cite{pearson2019human} for example, when a person is thinking about the concept of `dog', they will not only think about the English word and its meaning but also a visual image and a sound associated with it. In the context of geospatial tasks, multimodal data are ubiquitous. In general, data from different modalities provide different `views' that complement each other and provide more information to facilitate a holistic understanding of the data.

Recently, much progress has been made in building large-scale multimodal foundation models for joint reasoning from various domains, in particular, vision and language.
CLIP \cite{radford2021clip,gabriel2021openclip} is one of the first widely-adopted vision-language joint training frameworks. It uses self-supervised contrastive learning to learn a joint embedding of visual and text features.
BLIP \cite{li2022blip} improves over CLIP by training on synthetically-generated captions from internet-collected images. It is designed to handle both visual-language understanding and generation tasks.
BEiT-3 \cite{wang2022beit3} is a general-purpose multimodal foundation model that achieves state-of-the-art performance on both vision and vision-language tasks. It combines features from multi-modality expert networks. 
Florence \cite{yuan2021florence} is a vision-language foundation model that learns universal visual-language representations for objects, scenes, images, videos, as well as captions. 
Similarly, KOSMOS-1 \cite{huang2023kosmos1} learns from web-scale multimodal data including text and image pairs. It can transfer knowledge from one modality to another. 
Flamingo \cite{Alayrac2022Flamingodeepmind} is a family of visual language models that can be adapted to novel tasks using only a few annotated examples, i.e., few-shot learning. It encodes images or videos as inputs along with textual tokens to jointly reason about vision tasks.
The newest version of the GPT model, the GPT-4 \cite{openai2023gpt4}, also claims to perform multimodal analysis including text, audio, images, and videos.


\section{Exploration of the Effectiveness of Existing {\lpm}s on Various Geospatial Domains} \label{sec:exp}

The first question we would like to ask is \textit{how the existing cutting-edge foundation models perform when compared with the state-of-the-art fully supervised task-specific models on various geospatial tasks.} 
Geography is a very broad discipline that includes various subdomains such as Geospatial Semantics \cite{kuhn2005geospatial,jones2008geographical,janowicz2012geospatial,hu2018geo,mai2021geoqa}, Health Geography \cite{kearns2002medical,rosenberg2014health,chang2021mobility}, Urban Geography, \cite{zhang2018measuring,cai2020traffic,zhu2020understanding,kang2021understanding,huang2023learning}, Remote Sensing \cite{mishra2012post,burke2021using,rolf2021generalizable,lee2021scalable,elmustafa2022understanding}, and so on. To address the aforementioned question, in the following, we conduct experiments using various {\lpm}s on different tasks in the four geospatial subdomains mentioned earlier. The advantages and weaknesses of existing {\lpm}s will be discussed in detail.


\vspace{-0.2cm}
\subsection{Geospatial Semantics}
\label{sec:exp_ner}

\begin{minipage}[c]{0.495\textwidth}
	\begin{lstlisting}[
	style=prompt-style, 
	basicstyle=\ttfamily\tiny,
	linewidth=\textwidth,
	breaklines=true,
	captionpos=b, 
	caption={Typonym recognition with {\llm}s, e.g., \gptthree. Yellow block: the text snippet to be annotated. Orange box: \gptthree~ outputs.
	8 few-shot samples are used in this prompt. We only show 1 here while skipping others with "..." to save space.
	},
	label={ls:prompt-tr},
	rulecolor=\color{black},
	frame=tb
	]
%*\colorbox{pinkannoback}{[Instruction]}*)...
%*\colorbox{blueannoback}{Paragraph:}*) Alabama State Troopers say a Greenville man has died of his injuries after being hit by a pickup truck on Interstate 65 in Lowndes County.
%*\colorbox{greenannoback}{Q:}*) Which words in this paragraph represent named places?
%*\colorbox{redannoback}{A:}*) Alabama; Greenville; Lowndes
...
--
%*\colorbox{blueannoback}{Paragraph:}*) %*\colorbox{yellowannoback}{The Town of Washington is to what Williamsburg is to Virginia.}*)
%*\colorbox{greenannoback}{Q:}*) Which words in this paragraph represent named places?
%*\colorbox{redannoback}{A:}*)%*\colorbox{orangeannoback}{Washington; Williamsburg; Virginia}*)
	\end{lstlisting}
\end{minipage}
\begin{minipage}[c]{0.482\textwidth}
	\begin{lstlisting}[
	style=prompt-style, 
	basicstyle=\ttfamily\tiny,
	linewidth=\textwidth,
	breaklines=true,
	captionpos=b, 
	caption={Location description recognition with {\llm}s, e.g., \gptthree. Yellow block: the input text snippet. Orange box: \gptthree~ outputs.
	11 few-shot samples are used while 1 is shown.
	},
	label={ls:prompt-ldr},
	frame=tb
	]
%*\colorbox{pinkannoback}{[Instruction]}*)...
%*\colorbox{blueannoback}{Paragraph:}*) Papa stranded in home. Water rising above waist. HELP 8111 Woodlyn Rd, 77028 #houstonflood
%*\colorbox{greenannoback}{Q:}*) Which words in this paragraph represent location descriptions?
%*\colorbox{redannoback}{A:}*) 8111 Woodlyn Rd, 77028
...
--
%*\colorbox{blueannoback}{Paragraph:}*) %*\colorbox{yellowannoback}{HurricaneHarvey Help Need AT 7506 Jackrabbit Rd, Houston, TX 77095.}*)
%*\colorbox{greenannoback}{Q:}*) Which words in this paragraph represent location descriptions?
%*\colorbox{redannoback}{A:}*)%*\colorbox{orangeannoback}{7506 Jackrabbit Rd, Houston, TX 77095}*)
	\end{lstlisting}
\end{minipage}

As a starting point for our discussion, we first 
demonstrate empirically
the promise of leveraging {\llm}s for solving geospatial semantics tasks.
We hope that our results not only 
demonstrate the effectiveness of such general-purpose, few-shot learners in the geospatial semantics domain, but also challenges the current paradigm of training task-specific models as a common practice in GeoAI research. 

We compare the performance of 4 pre-trained GPT-2 \cite{radford2019gpt2} models of varying sizes provided by Huggingface as well as the most recent \gptthree~ \cite{brown2020gpt3} (i.e., text-davinci-002), \instructgpt~ \cite{ouyang2022instructgpt} (i.e., text-davinci-003), and \chatgpt~ \cite{openai2022chatgpt} (i.e., gpt-3.5-turbo) models developed by OpenAI 
with multiple \emph{supervised, task-specific} baselines on two representative geospatial semantics tasks: (1) toponym recognition \cite{gritta2018camcoder,wang2019EUPEG}, and (2) location description recognition \cite{hu2020people}. 

\subsubsection{Toponym Recognition} \label{sec:exp_topo}
Toponym recognition is a subtask of named entity recognition (NER), with the goal of identifying named places from a given text snippet. 
We use the Hu2014 \cite{hu2014improving} and Ju2016 \cite{ju2016things} benchmark datasets as two representative datasets for this task.
We adapt 7 pre-trained GPT models to perform toponym recognition tasks by using appropriate prompts containing few-shot training examples. 
In the prompt, we provide several training samples as few-shot learning samples in the form of natural language instructions. 
One example of such a prompt is illustrated in Listing \ref{ls:prompt-tr}, while the full prompts can be found in List \ref{ls:prompt-tr-all} in Appendix \ref{sec:app_prompt}.
It is worth noting that \chatgpt~, as a foundation model,  is optimized for chatbot purposes and expects conversational inputs rather than a single big prompt. 
In order to conduct a controlled experiment, we first use the same prompt shown in Listing \ref{ls:prompt-tr} to instruct all 7 pre-trained GPT models to perform toponym recognition. We also convert the few-shot examples into a list of conversations and use them as the inputs for \chatgpt~ which is denoted as \chatgptc~ while the \chatgpt~ using the original prompt is indicated as \chatgptr. 

Table \ref{tab:ner} compares all 8 GPT models with 15 baselines on two datasets -- Hu2014 \cite{hu2014improving} and Ju2016 \cite{ju2016things}. The same test sets have been used to evaluate the performances of all models. Please refer to Hu et al. \cite{hu2014improving} and Ju et al. \cite{ju2016things} for detailed descriptions of both datasets. 
Those 15 baselines are classified into three groups as shown in Table \ref{tab:ner}: (A) general NER (named entity recognition) models; (B) no neural network (NN) based geoparsers; (C) fully supervised task-specific NN-based geoparsers. 
All models in Group C are trained in a supervised manner on the same separated training datasets.
With the exception of the smallest \gpttwo~ model, all other {\llm}s consistently outperform the fully-supervised baselines on the Hu2014 dataset, 
even though they only require a small set of natural language instructions without any additional training.
\gptthree~ in particular demonstrated an 8.7\% performance improvement over the previous SOTA (TopoCluster \cite{delozier2016topocluster}). Interestingly, new GPT models such as \instructgpt~ and \chatgpt~ do not show higher performances on the Hu2014 dataset. While \instructgpt~ shows a smaller performance drop which is acceptable, two \chatgpt~ models show more significant performance decreases. One reasonable hypothesis is that \chatgpt~ is further optimized based on \instructgpt~ for chatbot applications that may not be ``flexible'' enough to be adapted to new tasks such as toponym recognition.  

Based on previous studies \cite{wang2019EUPEG,wang2020neurotpr}, the Ju2016 dataset is a very difficult task. 
On this dataset, we found that
\gpttwoxl~ outperforms the previous SOTA (NeuroTPR \cite{wang2020neurotpr}) by over 2.5\% while using only \emph{8 few-shot examples in the prompt}.
In contrast, a task-specific model, such as NeuroTPR, requires supervised training on 599 labeled tweets and labeled sentences generated from 3000 Wikipedia articles. 
\gptthree~ and \instructgpt~ does not show performance improvement on the Ju2016 dataset over \gpttwoxl. Similar to the finding on the Hu2014 dataset, \chatgpt~ shows a significant performance decrease on the Ju2016 dataset. 
In accordance with existing empirical findings \cite{radford2019gpt2,brown2020gpt3}, 
we also found that the performance of these {\llm}s tended to scale with the number of learnable parameters.

\subsubsection{Location Description Recognition} \label{sec:exp_loc_dec}
The location description recognition task is slightly more challenging -- given a text snippet (e.g., tweets), the goal is to recognize more fine-grained location descriptions such as home addresses, highways, roads, and administration regions.
HaveyTweet2017 \cite{hu2020people} is used as one representative benchmark dataset for this task. 
The same set of pre-trained GPT models and 15 baselines are used for this task. Listing \ref{ls:prompt-ldr} shows one example prompt used in this task and the full prompt can be seen in Listing \ref{ls:prompt-ldr-all} in Appendix \ref{sec:app_prompt}.

Table \ref{tab:ner} summarizes the evaluation results of different models on the HaveyTweet2017 dataset. 
The same test set of HaveyTweet2017 is used to evaluate all GPT models as well as 15 baseline models. 
On the HaveyTweet2017 dataset, \gptthree~ achieves the best recall score across all methods. 
However, all {\llm}s have rather low precision (and therefore low F1-scores).
This is because
{\llm}s implicitly convert the location description recognition problem into a natural language generation problem (see List \ref{ls:prompt-ldr}), 
meaning that they are not guaranteed to generate tokens that appear in the input text.
Based on the experimental results in Table \ref{tab:ner}, 
we can clearly see that by using just \textit{a small number of few-shot samples, {\llm}s can outperform the fully-supervised, task-specific models on well-defined geospatial semantics tasks}.
This showcases the potential of {\llm}s to dramatically reduce the need for customized architectures or large labeled datasets for geospatial tasks. 
However, how to develop appropriate prompts to instruct {\llm}s for a given geospatial semantics task require further investigation.

\vspace{-0.2cm}
\begin{table}[t!]
\caption{
Evaluation results of various GPT models and baselines on 
two geospatial semantics  tasks: 
 (1) toponym recognition (Hu2014 \cite{hu2014improving} and Ju2016 \cite{ju2016things}) and (2) location description recognition (HaveyTweet2017 \cite{hu2020people}) . 
We classify all models into four groups:
(A) General NER;
(B) No Neural Network (NN) based geoparsers;
(C) Fully-supervised NN-based geoparsers;
(D) Few-show learning with {\llm}s. 
"(\#Param)" indicates the number of learnable parameters of {\llm}s. 
"(nar. loc.)" and "(bor. loc.)" indicate narrow location models and broad location models defined in \cite{wang2020neurotpr}. 
The  results of all baselines (i.e., models in Group A, B, and C) are obtained from \cite{wang2019EUPEG} and \cite{wang2020neurotpr} except "0.675$^{\dag}$", 
which is obtained by rerunning the official code of \cite{wang2020neurotpr}.
The evaluation results of different GPT models (Group D) are done by using pre-trained \gpttwo/\gptthree/\instructgpt/\chatgpt~ models with appropriate prompts. 
The results of four \gpttwo~ models are obtained by using Huggingface pre-trained \gpttwo models with various model sizes.
The last four models are obtained by using various OpenAI's GPT models -- text-davinci-002, text-davinci-003, and gpt-3.5-turbo -- which are denoted as \gptthree, \instructgpt, and \chatgpt~ respectively. Since \chatgpt~ expects conversational inputs rather than a single big prompt, we experiment with two versions of \chatgpt. \chatgptr~ indicates we use the same prompt as other GPT models while \chatgptc~ indicates we convert the few-shot examples in the prompt into a list of conversations.
$^*$Due to OpenAI API limitations, we evaluate \gptthree, \instructgpt, and \chatgpt~ on randomly sampled 544 Ju2016 examples (10\%  of the dataset).
}
\label{tab:ner}
 \vspace{-0.35cm}
\centering
\setlength{\tabcolsep}{1.5pt}
{ 
\begin{tabular}{l|l|c|c|c|ccc}
\toprule
\multirow{3}{*}{}                                         & \multirow{3}{*}{Model}                           & \multirow{3}{*}{\#Param} & \multicolumn{2}{c}{Toponym Recognition} & \multicolumn{3}{|c}{Location Description Recognition} \\ \cline{4-8}
                                                          &                                                  &                          & Hu2014             & Ju2016             & \multicolumn{3}{c}{HaveyTweet2017}                   \\ \cline{4-8}
                                                          &                                                  &                          & Accuracy $\downarrow$           & Accuracy $\downarrow$           & Precision $\downarrow$        & Recall $\downarrow$          & F-Score $\downarrow$         \\ \hline
\multirow{8}{*}{(A)}                          & Stanford NER (nar. loc.) \cite{finkel2005stanfordner}                  & -                        & 0.787              & 0.010              & \textbf{0.828}   & 0.399           & 0.539           \\
                                                          & Stanford NER (bro. loc.) \cite{finkel2005stanfordner}                   & -                        & -                  & 0.012              & 0.729            & 0.44            & 0.548           \\
                                                          & Retrained Stanford NER \cite{finkel2005stanfordner}                          & -                        & -                  & 0.078              & 0.604            & 0.410           & 0.489           \\
                                                          & Caseless Stanford NER (nar. loc.)  \cite{finkel2005stanfordner}        & -                        & -                  & 0.460              & 0.803            & 0.320           & 0.458           \\
                                                          & Caseless Stanford NER (bro. loc.) \cite{finkel2005stanfordner}          & -                        & -                  & 0.514              & 0.721            & 0.336           & 0.460           \\
                                                          & spaCy NER (nar. loc.)  \cite{honnibal2017spacy}                    & -                        & 0.681              & 0.000              & 0.575            & 0.024           & 0.046           \\
                                                          & spaCy NER (bro. loc.)  \cite{honnibal2017spacy}                     & -                        & -                  & 0.006              & 0.461            & 0.304           & 0.366           \\
                                                          & DBpedia Spotlight\cite{mendes2011dbpedia}                                         & -                        & 0.688              & 0.447              & -                & -               & -               \\ \hline
\multirow{3}{*}{(B)}                 & Edinburgh  \cite{alex2015edinburgh}                                      & -                        & 0.656              & 0.000              & -                & -               & -               \\
                                                          & CLAVIN \cite{wang2019EUPEG}                                          & -                        & 0.650              & 0.000              & -                & -               & -               \\
                                                          & TopoCluster \cite{delozier2016topocluster}                                     & -                        & 0.794              & 0.158              & -                & -               & -               \\ \hline
\multirow{4}{*}{(C)} & CamCoder \cite{gritta2018camcoder}                                        & -                        & 0.637              & 0.004              & -                & -               & -               \\
                                                          & Basic BiLSTM+CRF  \cite{lample2016bilstm}           & -                        & -                  & 0.595              & 0.703            & 0.600           & 0.649           \\
                                                          & DM NLP (top. rec.) \cite{wang2019dm_nlp} & -                        & -                  & 0.723              & 0.729            & 0.680           & 0.703           \\
                                                          & NeuroTPR \cite{wang2020neurotpr}                                  & -                        & 0.675$^{\dag}$              & 0.821              & 0.787            & 0.678           & \textbf{0.728}  \\ \hline
\multirow{8}{*}{(D)}                          & \gpttwo~ \cite{radford2019gpt2}                                            & 117M                     & 0.556              & 0.650              & 0.540            & 0.413           & 0.468           \\
                                                          & \gpttwom~ \cite{radford2019gpt2}                                     & 345M                     & 0.806              & 0.802              & 0.529            & 0.503           & 0.515           \\
                                                          & \gpttwol~ \cite{radford2019gpt2}                                      & 774M                     & 0.813              & 0.779              & 0.598            & 0.458           & 0.518           \\
                                                          & \gpttwoxl~ \cite{radford2019gpt2}                                         & 1558M                    & 0.869              & \textbf{0.846}     & 0.492            & 0.470           & 0.481           \\
                                                          & \gptthree~ \cite{brown2020gpt3}                             & 175B                     & \textbf{0.881}     & 0.811$^*$              & 0.603            & \textbf{0.724}  & 0.658           \\
                                                          & \instructgpt~ \cite{ouyang2022instructgpt}                                & 175B                     & 0.863              & 0.817$^*$              & 0.567            & 0.688           & 0.622           \\
                                                          & \chatgptr~ \cite{openai2022chatgpt}                            & 176B                     & 0.800              & 0.696$^*$              & 0.516            & 0.654           & 0.577           \\
                                                          & \chatgptc~ \cite{openai2022chatgpt}                           & 176B                     & 0.806              & 0.656$^*$              & 0.548            & 0.665           & 0.601 \\ \bottomrule         
\end{tabular}
}
\end{table}

\subsection{Health Geography} \label{sec:exp_alz}

The next set of experiments focuses on an important health geography problem --  dementia death counts time series forecasting for a given geographic region such as cities, counties, states, etc. With a growing share of older adults in the population, it is estimated that more than 7 million US adults aged 65 or older were living with dementia in 2020, and the number could increase to over 9 million by 2030 and nearly 12 million by 2040 \cite{zissimopoulos2018impact}. Alzheimer’s disease, the most common type of dementia, has been reported to be one of the top leading causes of death in the US, with 1 in 3 seniors dying with Alzheimer's or another dementia by 2019 \cite{alzheimer2022alzheimer}. Notably, there are substantial and longstanding geographical disparities in mortality due to dementia \cite{alzheimer2021changing,akushevich2021geographic}. Subnational planning and prioritizing dementia prevention strategies require local mortality data. Prediction of dementia deaths at the sub-national level will assist in informing future tailored health policies to eliminate geographical disparities in dementia and to achieve national health goals.

In this work, we conduct time series forecasting on the number of deaths due to dementia in two geographic region levels – state level and county level. The dementia data are obtained from the US Centers for Disease Control and Prevention Wide-ranging Online Data for Epidemiologic Research (CDC WONDER\footnote{\url{https://wonder.cdc.gov/ucd-icd10.html}}), which is a publicly available dataset. Dementia deaths are classified according to the International Classification of Diseases, Tenth Revision (ICD-10), including unspecified dementia (F03), Alzheimer’s disease (G30), vascular dementia (F01), and other degenerative diseases of nervous system, not elsewhere classified (G31) \cite{kramarow2019dementia}.

\subsubsection{US State-Level Dementia Time Series Forecasting} \label{sec:exp_alz_state}
We collect annual time series of dementia death counts for all 51 US states between 1999 and 2020. The time series from 1999 to 2019 are used as training data, and the numbers in 2020 are used as ground truth labels. 
The same set of pre-trained GPT models used in Section \ref{sec:exp_ner} are utilized in this task. 
Different from the geospatial semantics experiments, we utilize all GPT models in a zero-shot setting since we think the historical time series data is enough for a \llm~ to perform the forecasting. Listing \ref{ls:prompt-alz-state} shows one example prompt we use in this experiment by using California as an example. 

With only 51 time series, each consisting of 22 data points, many sequential deep learning models such as RNNs (recurrent neural networks) and Transformers \cite{vaswani2017attention} are at risk of overfitting on this dataset. So we pick the state-of-the-art machine learning-based time series forecasting model -- ARIMA (Autoregressive integrated moving average) as the fully supervised task-specific baseline model. 
We train individual ARIMA models on each state's time series using data from 1999 to 2019, and perform forecasting on data in 2020.
Hyperparameter tuning is performed on all ARIMA hyperparameters to obtain the best results.
Additionally, we use 
persistence model \cite{notton2018forecasting,paulescu2021nowcasting} as a reference. A persistence model assumes that the future value of a time series remains the same between the current time and the forecast time. In our case, we use the dementia death count of each state in 2019 as the prediction for the value in 2020. 

Table \ref{tab:alz_state} presents a comparison of model performances among different GPT models and two baselines. Interestingly, all \gpttwo~ models perform poorly on all evaluation metrics. Their performances are even worse than the simple persistence model. This suggests that \gpttwo~ may struggle with zero-shot time series forecasting. On the other hand, \gptthree, \instructgpt, and two \chatgpt~ models demonstrate reasonable performances. Of particular interest is that \instructgpt~ outperforms the best ARIMA model on all evaluation metrics even though \instructgpt~ is not finetuned on this specific task.
We propose two hypothetical reasons for the strong performance of \instructgpt~ in the time series forecasting task: 
1) After training on a large-scale text corpus, \instructgpt~ may have developed the intelligence necessary to perform zero-shot time series forecasting, which is fundamentally an autoregressive problem.
2) It is possible that \instructgpt~ and \gptthree~ may be exposed to US state-level dementia time series data during their training on the large-scale text corpus.

While we cannot determine which of these reasons is the primary factor behind \instructgpt's success, 
these results are very encouraging. 
Similar to the results in Table \ref{tab:ner}, two \chatgpt~ models underperform \instructgpt. 
More experiment analysis can be seen in the county-level experiments.

\begin{table}[t!]
\caption{
Evaluation results of various GPT models and baselines on the
US state-level dementia time series forecasting task. 
We classify all models into four groups:
(A) Simple persistent model;
(B) Fully supervised machine learning models such as ARIMA;
(C) Zero-shot learning with {\llm}s. 
"(\#Param)" indicates the number of learnable parameters of {\llm}s. 
The denotations of different GPT models are the same as Table \ref{tab:ner}.
Four evaluation metrics are used: MSE (mean square error), MAE (mean absolute error), MAPE (mean absolute percentage error), and R$^2$.
$\uparrow$ and $\downarrow$ indicate the direction of better models for each metric.
For all GPT models, we encode time series information between 1999 and 2019 in the prompt and let it generate data in 2020.
}
\label{tab:alz_state}
 \vspace{-0.35cm}
\centering
{ 
\begin{tabular}{l|l|c|c|c|c|c}
\toprule
& Model                       & \#Param & MSE $\downarrow$                  & MAE $\downarrow$              & MAPE $\downarrow$           & R$^2$ $\uparrow$            \\ \hline
(A)  Simple                       & Persistence \cite{notton2018forecasting,paulescu2021nowcasting}                 & -       & 985,179        & 630         & 0.096          & 0.971          \\ \hline
(B) Supervised ML                 & ARIMA \cite{jenkins2011time}                       & -       & 562,768          & 462          & 0.067          & 0.984          \\ \hline
\multirow{8}{*}{(C) Zero shot LM} & \gpttwo~ \cite{radford2019gpt2}                        & 117M    & 44,635,055      & 4,898        & 0.955          & -0.271         \\
& \gpttwom~ \cite{radford2019gpt2}                 & 345M    & 42,315,630       & 4,616        & 0.745          & -0.209         \\
& \gpttwol~ \cite{radford2019gpt2}                  & 774M    & 39,039,733       & 4,250        & 0.779          & -0.132         \\
& \gpttwoxl~ \cite{radford2019gpt2}                     & 1558M   & 35,355,840       & 3,912        & 0.709          & -0.026         \\
& \gptthree~ \cite{brown2020gpt3}            & 175B    & 587,263          & 474          & 0.070          & 0.983          \\
& \instructgpt~ \cite{ouyang2022instructgpt}            & 175B    & \textbf{387,413} & \textbf{365} & \textbf{0.055} & \textbf{0.989} \\
& \chatgptr~ \cite{openai2022chatgpt}               & 176B    & 1,143,675        & 623          & 0.121          & 0.967          \\
& \chatgptc~ \cite{openai2022chatgpt}  & 176B    & 4,224,811        & 1,131        & 0.240          & 0.890    \\ \bottomrule     
\end{tabular}
}
\vspace{-0.3cm}
\end{table}


\begin{minipage}[c]{0.48\textwidth}
	\begin{lstlisting}[
	style=prompt-style, 
	basicstyle=\ttfamily\tiny,
	linewidth=\textwidth,
	breaklines=true,
	captionpos=b, 
	caption={US state-level Alzimier time series forecasting with LLMs by zero-shot learning. Yellow block: the historical time series data of one US state. Orange box: the outputs of \instructgpt. Here, we use California as an example and the correct answer is 29400.
	},
	label={ls:prompt-alz-state},
	rulecolor=\color{black},
	frame=tb
	]
%*\colorbox{pinkannoback}{[Instruction]}*)This is a set of time series forecasting problems.
The `Paragraph` is a time series of the numbers of deaths from alzheimer's disease for one of US state from 1999 to 2019.
The goal is to predict the number of deaths from alzheimer's disease at this state in 2020. Please give a single number as the prediction.
--
--
%*\colorbox{blueannoback}{Paragraph:}*) At California, From 1999 to 2019, the numbers of deaths from alzheimer's disease are %*\colorbox{yellowannoback}{6761 in 1999, 6760 in 2000, 7474}*) %*\colorbox{yellowannoback}{ in 2001, 8366 in 2002, 9760 in 2003, 9806 in 2004, 11497 in 2005,}*) %*\colorbox{yellowannoback}{  13520 in 2006, 13730 in 2007, 16395 in 2008, 16290 in 2009, 18000}*) %*\colorbox{yellowannoback}{in 2010, 19924 in 2011,20814 in 2012, 22061 in 2013, 22412 in 2014,}*) %*\colorbox{yellowannoback}{23606 in 2015, 24060 in 2016, 25017 in 2017, 25218 in 2018, and}*) %*\colorbox{yellowannoback}{25810 in 2019.}*)
%*\colorbox{greenannoback}{Q:}*) Please forecast the number in 2020 at California? 
%*\colorbox{redannoback}{A:}*)%*\colorbox{orangeannoback}{26670}*)
	\end{lstlisting}
\end{minipage}
\begin{minipage}[c]{0.015\textwidth}
\end{minipage}
\begin{minipage}[c]{0.50\textwidth}
	\begin{lstlisting}[
	style=prompt-style, 
	basicstyle=\ttfamily\tiny,
	linewidth=\textwidth,
	breaklines=true,
	captionpos=b, 
	caption={US county-level Alzimier time series forecasting with LLMs by zero-shot learning. Yellow block: the historical time series data of one US county. Orange box: the outputs of \instructgpt. Here, we use Santa Barbara County, CA as an example and the correct answer is 373.
	},
	label={ls:prompt-alz-county},
	rulecolor=\color{black},
	frame=tb
	]
%*\colorbox{pinkannoback}{[Instruction]}*)This is a set of time series forecasting problems.
The `Paragraph` is a time series of the numbers of deaths from alzheimer's disease for one of US counties from 1999 to 2019.
The goal is to predict the number of deaths from alzheimer's disease at this county in 2020. Please give a single number as the prediction.
--
--
%*\colorbox{blueannoback}{Paragraph:}*) At Santa Barbara County, CA, from 1999 to 2019, the numbers of deaths from alzheimer's disease are %*\colorbox{yellowannoback}{126 in 1999, 114 in 2000, 124 in 2001, 127 in 2002, 156 in 2003,}*) %*\colorbox{yellowannoback}{ 154 in 2004, 175 in2005, 172 in 2006, 171 in 2007, 248 in 2008, 204}*) %*\colorbox{yellowannoback}{in 2009, 241 in 2010, 260 in 2011, 297 in 2012, 283 in 2013, 308 in}*) %*\colorbox{yellowannoback}{2014, 358 in 2015, 365 in 2016, 334 in 2017, 363 in 2018,}*) %*\colorbox{yellowannoback}{and 328 in 2019.}*)
%*\colorbox{greenannoback}{Q:}*) Please forecast the number in 2020 at Santa Barbara County, CA? 
%*\colorbox{redannoback}{A:}*)%*\colorbox{orangeannoback}{345}*)
	\end{lstlisting}
\end{minipage}

\subsubsection{US County-Level Dementia Time Series Forecasting} \label{sec:exp_alz_county}
In terms of county-level data, we utilized the dementia death count time series of all US counties with available data, resulting in a total of 2447 US counties selected for analysis. We only considered counties with dementia annual death records spanning more than four years between 1999 and 2020.
Similarly to Section \ref{sec:exp_alz_state}, 
we utilize all available data up to the given year for training ARIMA models and generating GPT prompts, and then make predictions for the following year.
We employ the same set of GPT models and baselines as in the state-level experiment to conduct the county-level experiment.
Listing \ref{ls:prompt-alz-county} shows one example prompt we use in this experiment by using Santa Barbara County, CA as an example. 

\begin{table}[t!]
\caption{
Evaluation results of various GPT models and baselines on the
US county-level dementia time series forecasting task. 
We use same model set and evaluation metrics as Table \ref{tab:alz_state}.
}
\label{tab:alz_county}
\centering
{ 
\begin{tabular}{l|l|c|c|c|c|c}
\toprule
& Model                       & \#Param & MSE $\downarrow$                  & MAE $\downarrow$              & MAPE $\downarrow$           & R$^2$ $\uparrow$            \\ \hline
(A)  Simple                       & Persistence \cite{notton2018forecasting,paulescu2021nowcasting}                 & -       & 1,648        & 16.9          & 0.189          & 0.979          \\ \hline
(B) Supervised ML                 & ARIMA \cite{jenkins2011time}                       & -       & 1,133      & 15.1          & 0.193          & 0.986          \\ \hline
\multirow{8}{*}{(C) Zero shot LLMs} & \gpttwo~ \cite{radford2019gpt2}                        & 117M    & 77,529       & 92.0          & 0.587          & -0.018         \\
& \gpttwom~ \cite{radford2019gpt2}                 & 345M    & 226,259     & 108.1         & 0.611          & -2.824         \\
& \gpttwol~ \cite{radford2019gpt2}                  & 774M    & 211,881      & 94.3          & 0.581          & -1.706         \\
& \gpttwoxl~ \cite{radford2019gpt2}                     & 1558M   & 162,778      & 99.8          & 0.627          & -1.082         \\
& \gptthree~ \cite{brown2020gpt3}            & 175B    & 1,105        & 14.5          & 0.180          & 0.986          \\
& \instructgpt~ \cite{ouyang2022instructgpt}            & 175B    & \textbf{831} & \textbf{13.3} & \textbf{0.179} & \textbf{0.989} \\
& \chatgptr~ \cite{openai2022chatgpt}               & 176B    & 4,115        & 23.2          & 0.217          & 0.955          \\
& \chatgptc~ \cite{openai2022chatgpt} & 176B    & 3,402        & 20.7          & 0.231          & 0.944             \\ \bottomrule     
\end{tabular}
}
\end{table}


Table \ref{tab:alz_county} compares the results of different models. Similar findings can be seen from these results. All \gpttwo~ models perform poorly. However, both \gptthree~ and \instructgpt~ outperform the best ARIMA models on all evaluation metrics, while two \chatgpt~ models underperform them. Among the two \chatgpt~ models, \chatgptc~ are slightly better than \chatgptr~ on all metrics except MAPE.

\begin{figure*}
	\centering \tiny
	\vspace*{-0.2cm}
	\begin{subfigure}[b]{0.33\textwidth}  
		\centering 
		\includegraphics[width=\textwidth]{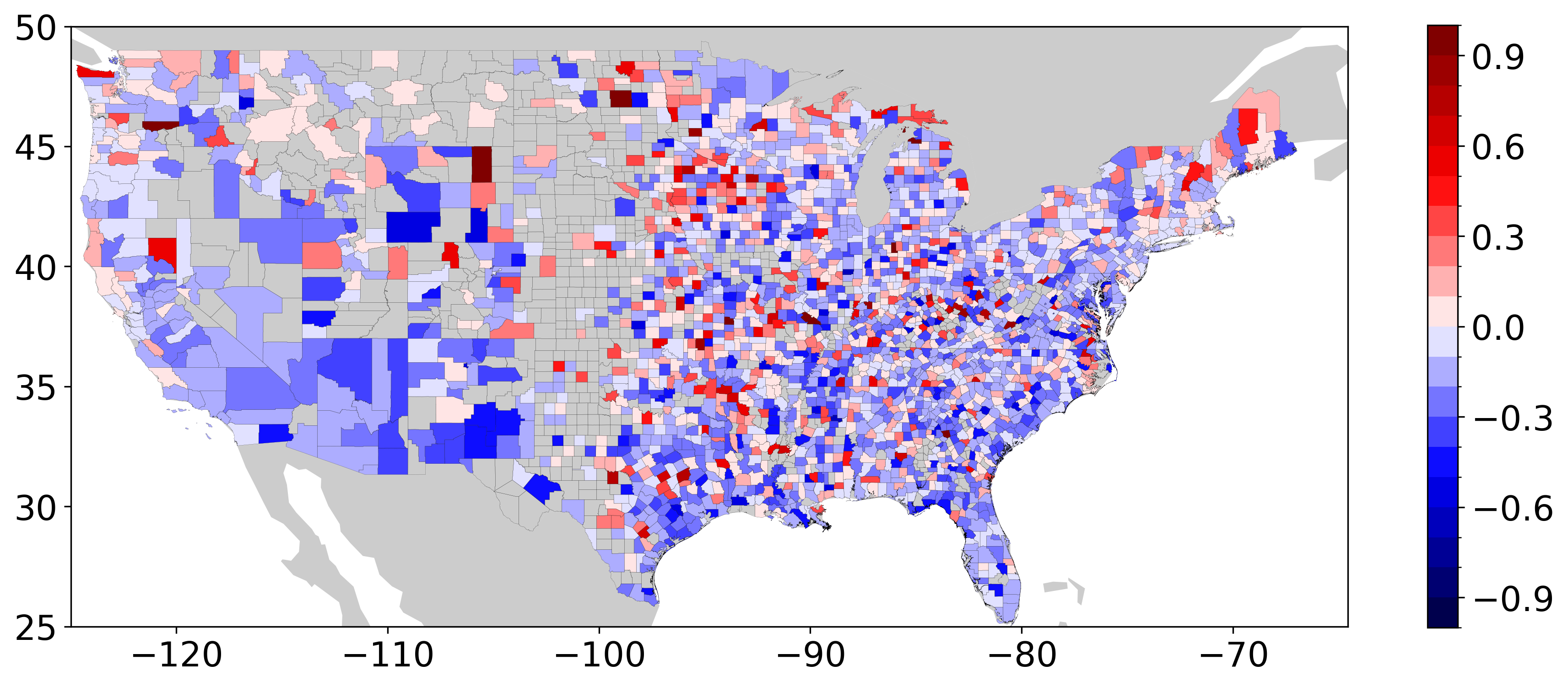}\vspace*{-0.2cm}
		\caption[]%
		{{ 
		Persistence
		}}    
		\label{fig:alz_county_persist}
	\end{subfigure}
        \begin{subfigure}[b]{0.33\textwidth}  
		\centering 
		\includegraphics[width=\textwidth]{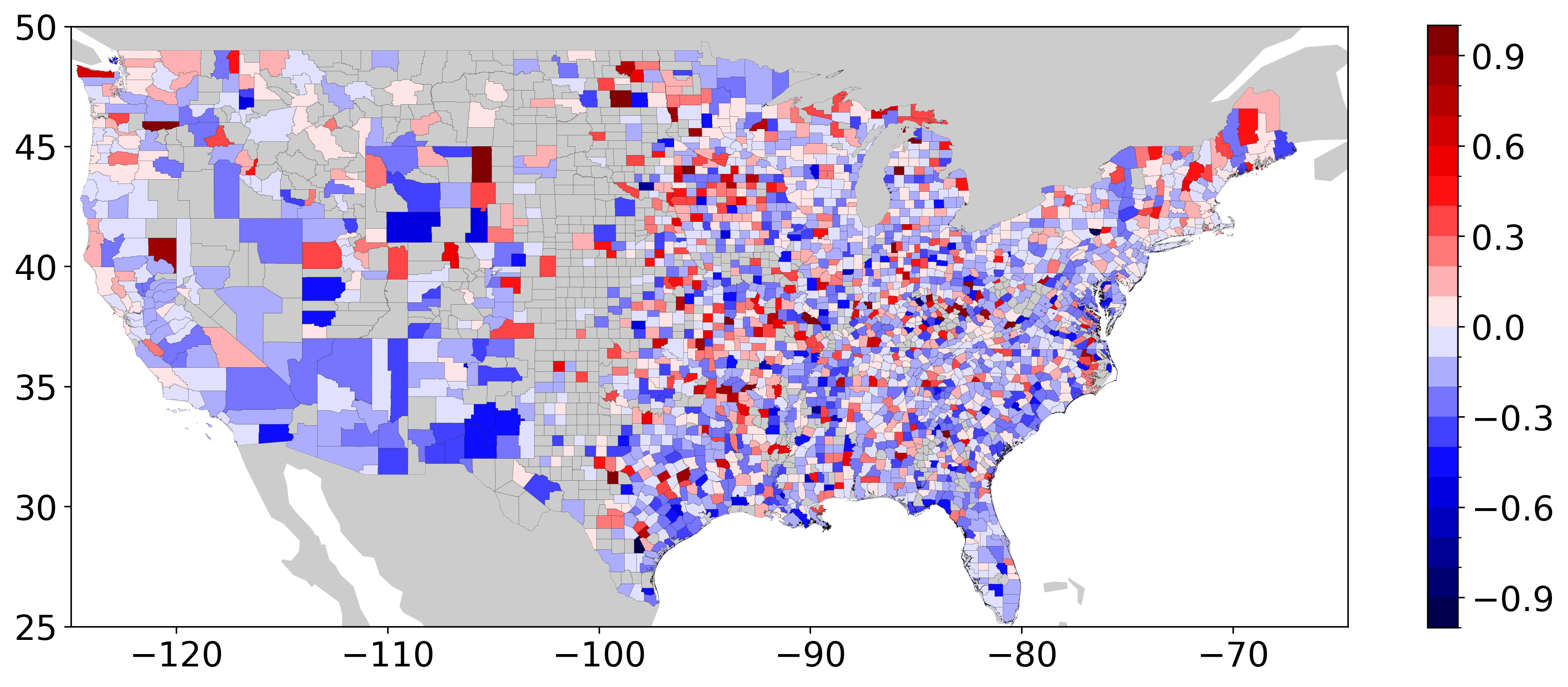}\vspace*{-0.2cm}
		\caption[]%
		{{ 
		ARIMA
		}}    
		\label{fig:alz_county_arima}
	\end{subfigure}
        \begin{subfigure}[b]{0.33\textwidth}  
		\centering 
		\includegraphics[width=\textwidth]{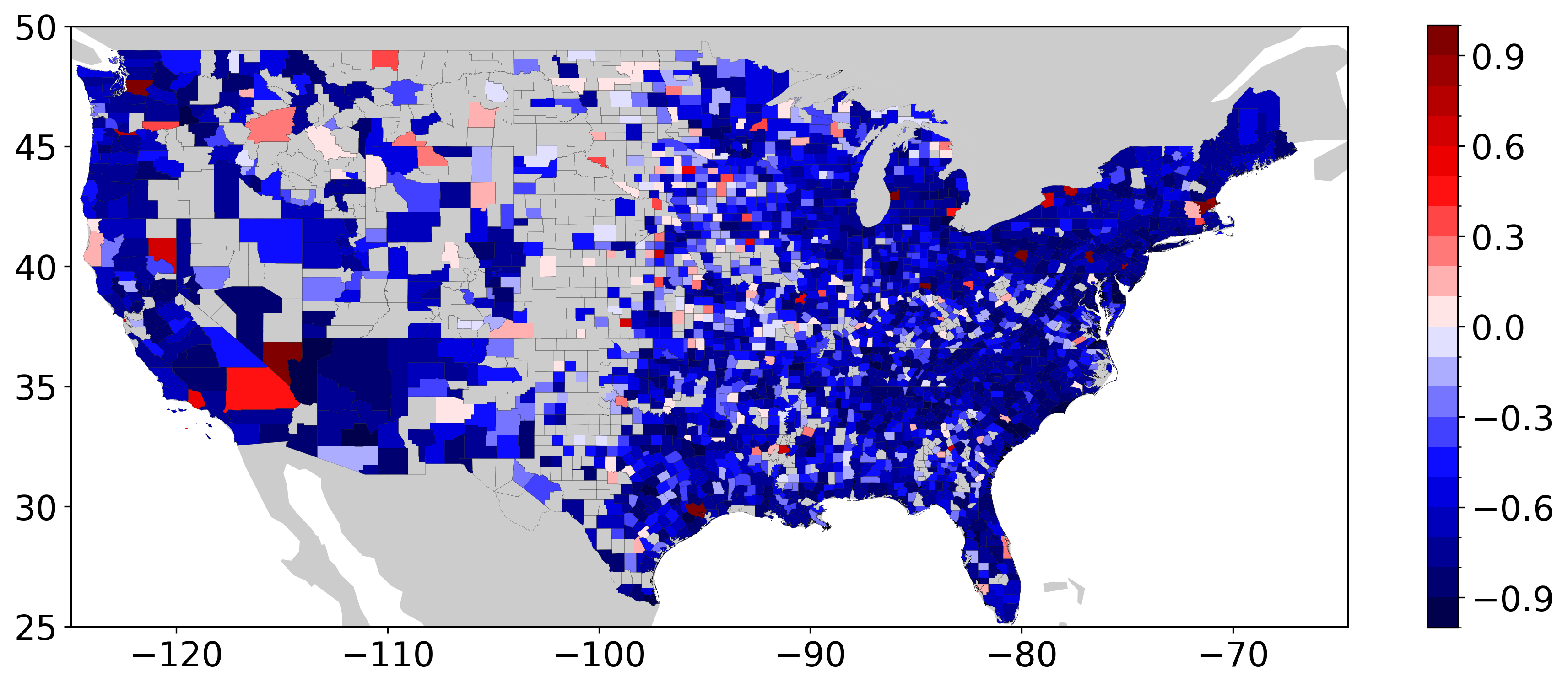}\vspace*{-0.2cm}
		\caption[]%
		{{ 
		\gpttwo
		}}    
		\label{fig:alz_county_gpt2}
	\end{subfigure}
        \begin{subfigure}[b]{0.33\textwidth}  
		\centering 
		\includegraphics[width=\textwidth]{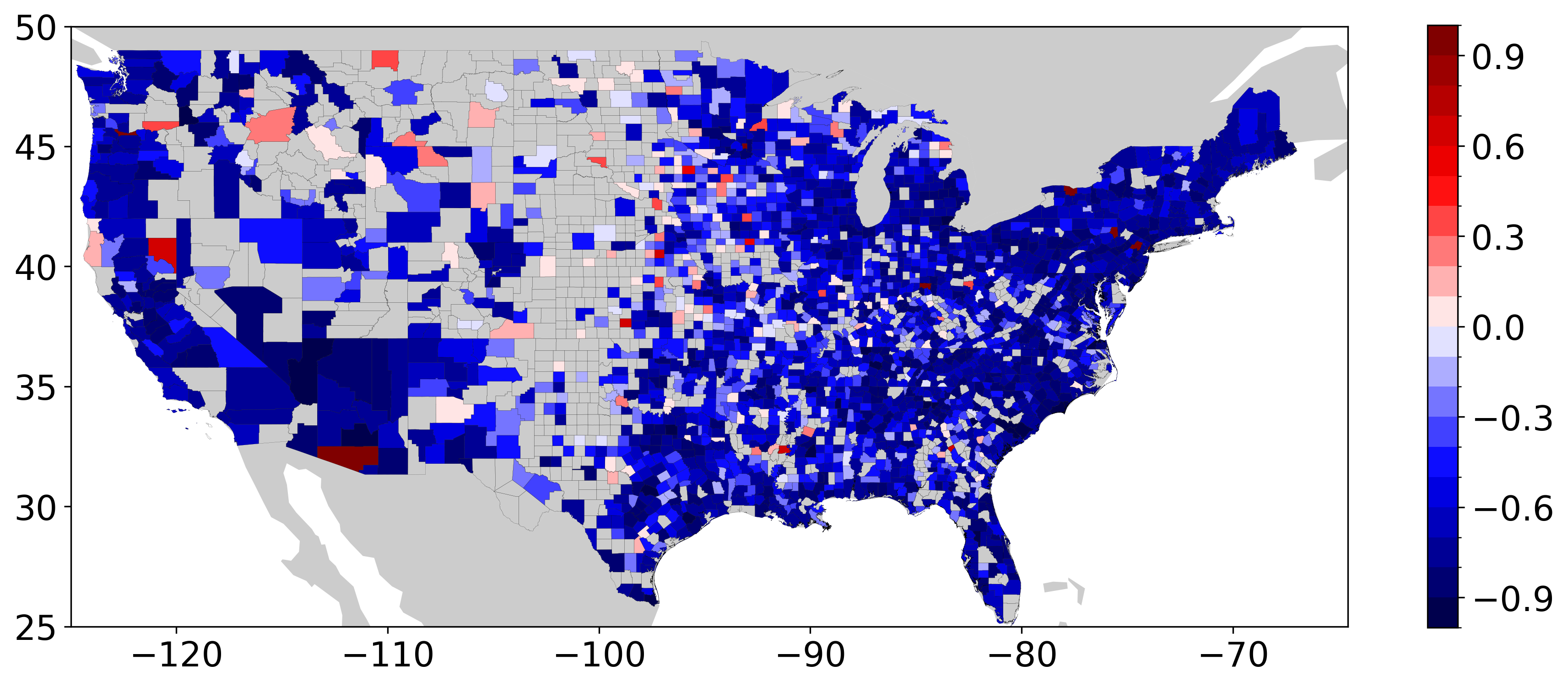}\vspace*{-0.2cm}
		\caption[]%
		{{
		\gpttwom
		}}    
		\label{fig:alz_county_gpt2-medium}
	\end{subfigure}
        \begin{subfigure}[b]{0.33\textwidth}  
		\centering 
		\includegraphics[width=\textwidth]{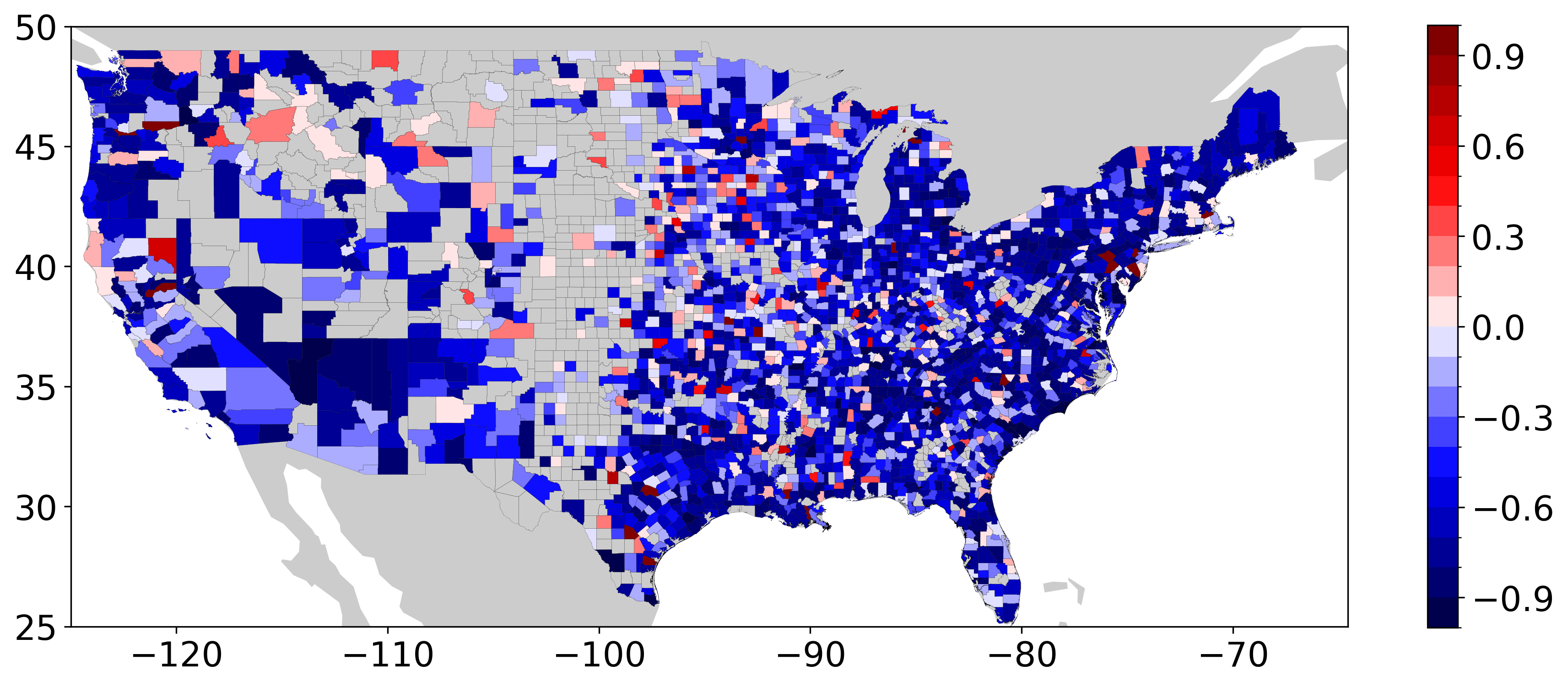}\vspace*{-0.2cm}
		\caption[]%
		{{ 
		\gpttwol
		}}    
		\label{fig:alz_county_gpt2-large}
	\end{subfigure}
        \begin{subfigure}[b]{0.33\textwidth}  
		\centering 
		\includegraphics[width=\textwidth]{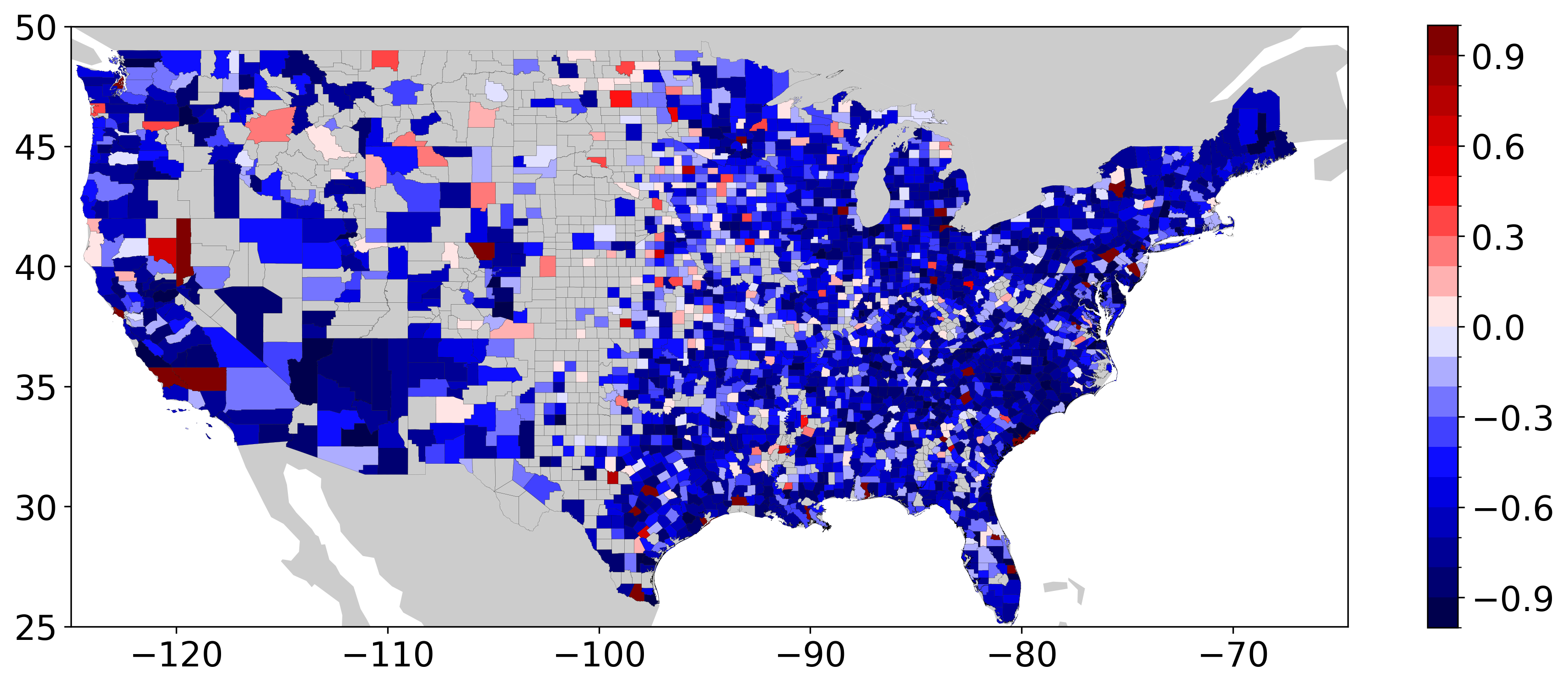}\vspace*{-0.2cm}
		\caption[]%
		{{
            \gpttwoxl
		}}    
		\label{fig:alz_county_gpt2-xl}
	\end{subfigure}
        \begin{subfigure}[b]{0.33\textwidth}  
		\centering 
		\includegraphics[width=\textwidth]{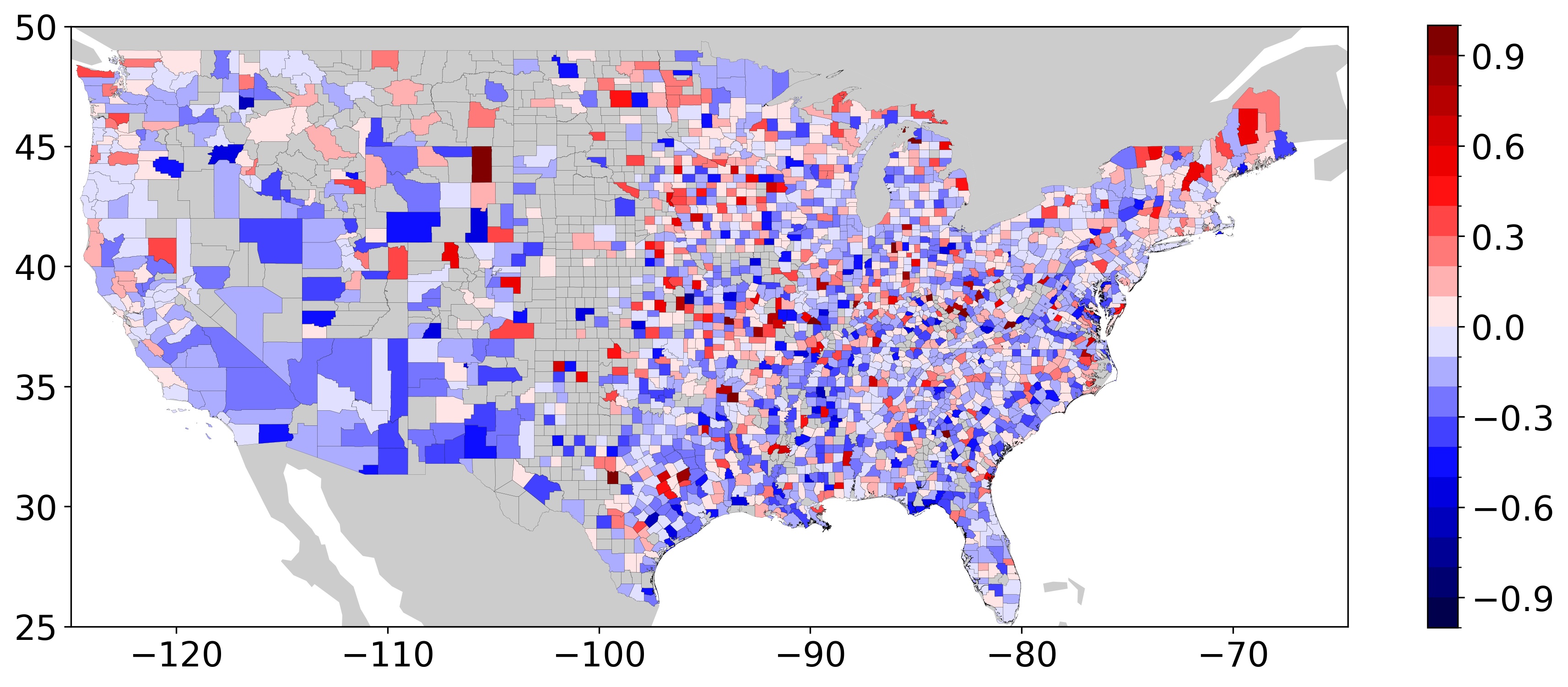}\vspace*{-0.2cm}
		\caption[]%
		{{ 
		\gptthree
		}}    
		\label{fig:alz_county_text-davinci-002}
	\end{subfigure}
        \begin{subfigure}[b]{0.33\textwidth}  
		\centering 
		\includegraphics[width=\textwidth]{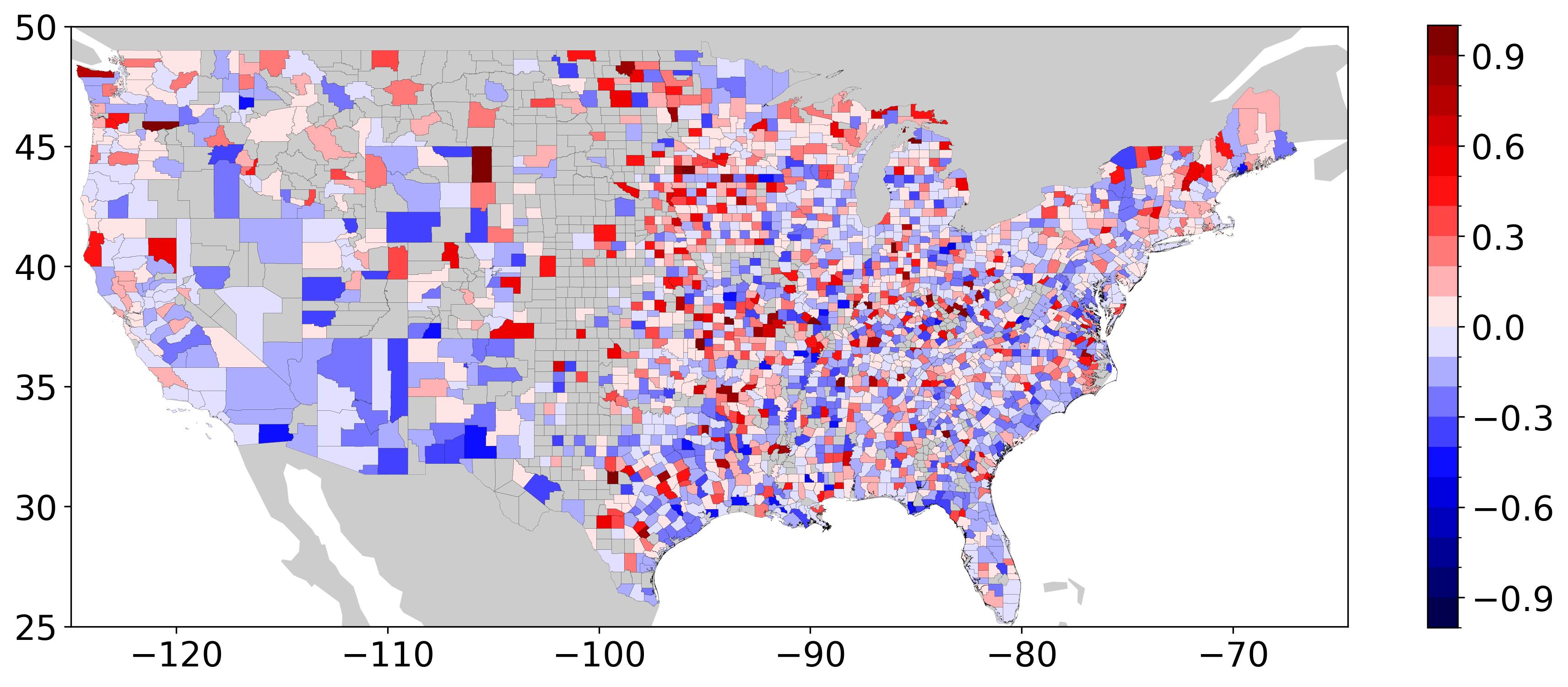}\vspace*{-0.2cm}
		\caption[]%
		{{\instructgpt
		}}    
		\label{fig:alz_county_text-davinci-003}
	\end{subfigure}
        \begin{subfigure}[b]{0.33\textwidth}  
		\centering 
		\includegraphics[width=\textwidth]{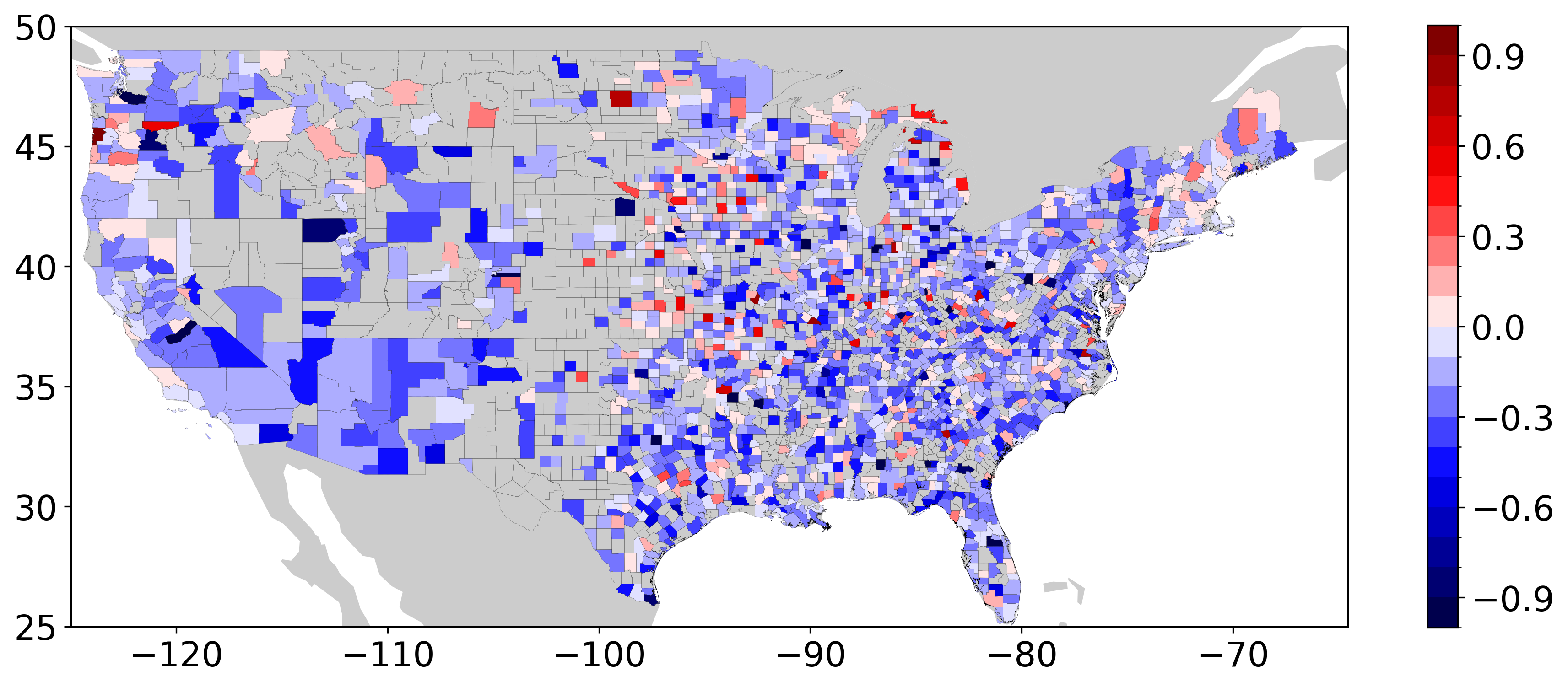}\vspace*{-0.2cm}
		\caption[]%
		{{\chatgptr
		}}    
		\label{fig:alz_county_gpt-3.5-turbo}
	\end{subfigure}
        \begin{subfigure}[b]{0.33\textwidth}  
		\centering 
		\includegraphics[width=\textwidth]{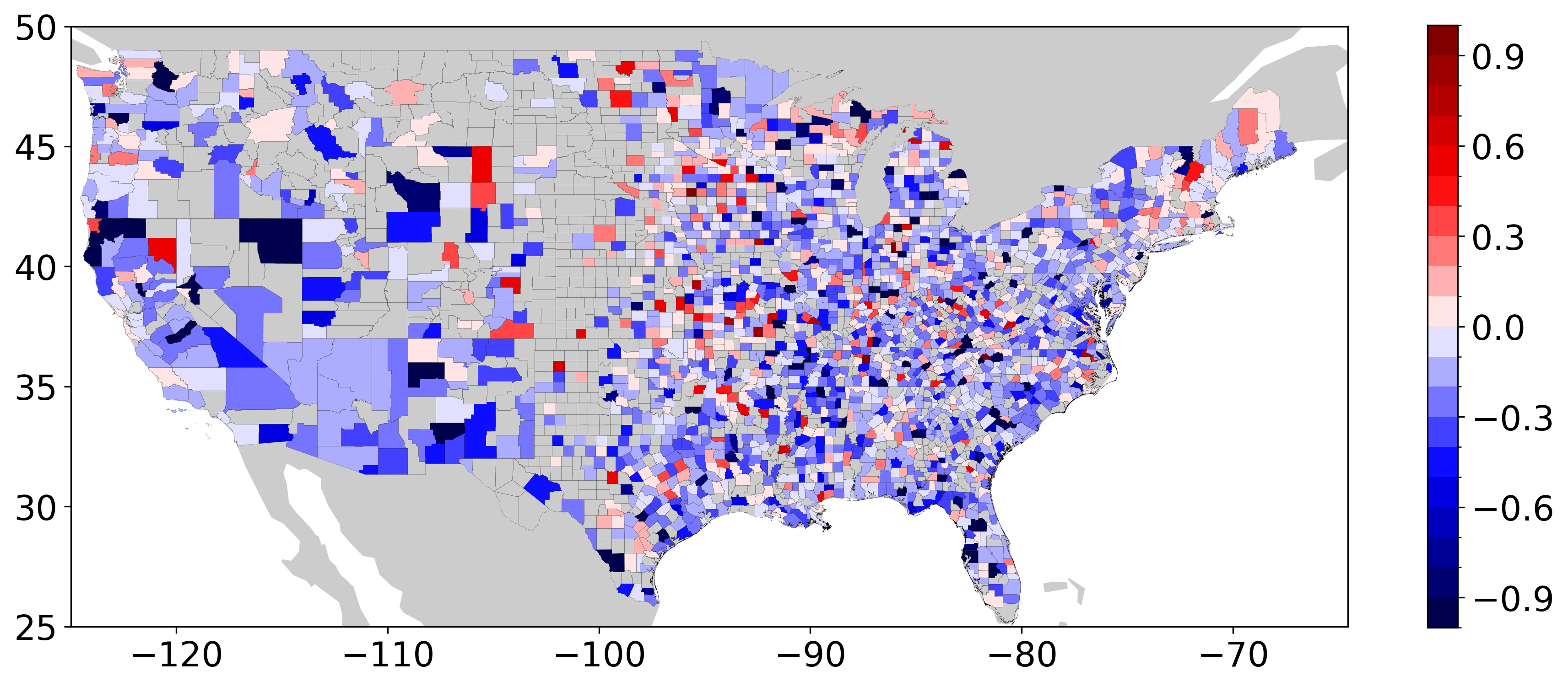}\vspace*{-0.2cm}
		\caption[]%
		{{\chatgptc
		}}    
		\label{fig:alz_county_gpt-3.5-turbo_conversation}
	\end{subfigure}

	\caption{Prediction error maps of each baseline and GPT model on US county-level dementia death count time series forecasting task. The color on each US count indicates the percentage error $PE = (Prediction - Label)/Label$ of each model prediction on this county. Those counties in gray color indicate their dementia data during 1999 and 2020 are not available. 
 }
	\label{fig:alz_county_all}
    \vspace*{-0.15cm}
\end{figure*}

To further understand the geographical distributions of prediction errors for each model, we visualize the prediction errors of each model on each US county in Figure \ref{fig:alz_county_all}. In the figure, the red color represents overestimations of the corresponding model while the blue colors indicate underestimations. Moreover, the intensity of the color indicates the magnitude of the prediction error, with darker colors representing larger errors. We can see that Persistence, ARIMA, \gptthree, and \instructgpt~ generally demonstrate better forecasting performance. However, the prediction percentage errors are not uniformly distributed across different US counties. As persistence uses the previous year's data as the prediction, Figure \ref{fig:alz_county_persist} indicates that the growth rates of dementia death counts are uneven for different counties. The southwest of the U.S. shows a recent increase in dementia death counts which leads the persistence model to underestimate the true data. 
The current maps of prediction errors show that the distributions of errors of \gptthree~ and \instructgpt~ are not uniform across the US counties, and it is unclear whether the uneven distribution is due to the geographic bias encoded in the models or the spatial heterogeneity of the growth rate of dementia death counts. Further analysis is needed to determine the cause of these uneven distributions.

One obvious observation from Figure \ref{fig:alz_county_all} is that all \gpttwo~ models turn to significantly underestimate the dementia data. To understand the cause of this behavior and the superiority of \gptthree~ and \instructgpt, we showcase the generated answers of different GPT models for four US counties in Table \ref{tab:alz_county_examples}. From Table \ref{tab:alz_county_examples}, it is evident that \gpttwo~ in many times will repeat the information provided in our prompt rather than generating novel predictions. For example, in the Clarke County, GA and Santa Barbara County, CA cases, all three \gpttwo~ models (i.e., \gpttwom, \gpttwol, and \gpttwoxl) predict the same numbers as the data in 1999. 
This suggests that these models rely heavily on the prompt information instead of learning from the time series data, which could explain their inferior performance compared to other models such as \gptthree~ and \instructgpt.
In the other two counties' cases, the predictions of the \gpttwo~ models vary significantly. In most cases, both \instructgpt~ and \chatgptr~ generate a single number as the prediction, indicating that they understand the task they are expected to perform. The only exception is the Santa Barbara County case, where \chatgptr~ generates a short sentence containing a reasonable prediction. 
However, based on our evaluation, the predictions of \chatgptr~ are not as good as those of \gptthree. Interestingly, when using \chatgpt~ in a conversational context, i.e., \chatgptc, \chatgpt~ usually returns a very long sentence. In the New York County case, \chatgptc~ is unable to give a prediction, suggesting that \chatgpt~ is useful in a chatbot context but may not be the best choice for other tasks such as time series forecasting.

\begin{table}[t!]
\caption{
Generated answers by various GPTs models for four US counties in US county-level dementia time series forecasting problem.
``County'' and ``Label'' columns indicate the US county and its true dementia death count in 2020.
``Generated Answer'' indicates the generated answer for different GPT models in which the numbers in bold are the predictions we extract from these answers. 
}
\label{tab:alz_county_examples}
\centering
\setlength{\tabcolsep}{1.5pt}
{ 
\scriptsize 
\begin{tabular}{l|l|l|p{10.5cm}}
\toprule
County                                    & Label                & Model                       & Generated Answer                                                   \\ \hline
\multirow{8}{*}{Clarke County, GA}        & \multirow{8}{*}{79}  & \gpttwo~                       & The number of deaths from alzheimer's disease is \textbf{21} in 1999, 37 in 2000, 32 in 2001, 25 in 2002, 29 in 2003, 44 in 2004, 38 in 2005, 41 in 2006, 52 in 2007, 46 in 2008                                                                                                                                                                                                                                                                       \\ \cline{3-4}
                                          &                      & \gpttwom~                & The number of deaths from alzheimer's disease in Clarke County, GA is \textbf{21} in 2020.                                                                                                                                                                                                                                                                                                                                                             \\ \cline{3-4}
                                          &                      & \gpttwol~                  & The number of deaths from alzheimer's disease in 2020 is: \textbf{21} in 2020, 37 in 2021, 32 in 2022, 25 in 2023, 29 in 2024, 44 in 2025, 38 in 2026, 42 in 2027, 42                                                                                                                                                                                                                                                                                  \\ \cline{3-4}
                                          &                      & \gpttwoxl~                     & The number in 2020 is: \textbf{21}.                                                                                                                                                                                                                                                                                                                                                                                                                    \\ \cline{3-4}
                                          &                      & \gptthree~            & The number of deaths from Alzheimer's disease in 2020 at Clarke County, GA is forecast to be \textbf{81}.                                                                                                                                                                                                                                                                                                                                              \\ \cline{3-4}
                                          &                      & \instructgpt~            & \textbf{83}                                                                                                                                                                                                                                                                                                                                                                                                                                            \\ \cline{3-4}
                                          &                      & \chatgptr~              & \textbf{78}                                                                                                                                                                                                                                                                                                                                                                                                                                            \\ \cline{3-4}
                                          &                      & \chatgptc~ & Based on the available historical data, the forecast for the number of deaths from Alzheimer's disease in Clarke County, GA in 2020 would be \textbf{81}.                                                                                                                                                                                                                                                                                              \\ \hline 
\multirow{8}{*}{New York County, NY}      & \multirow{8}{*}{835} & \gpttwo~                       & The number of deaths from alzheimer's disease is \textbf{91} in 1999, 121 in 2000, 118 in 2001, 155 in 2002, 214 in 2003, 243 in 2004, 219 in 2005, 294 in 2006, 296 in 2007, 364 in 2008                                                                                                                                                                                                                                                              \\ \cline{3-4}
                                          &                      & \gpttwom~                & The number of deaths from alzheimer's disease in New York County, NY is \textbf{9,965}.                                                                                                                                                                                                                                                                                                                                                                \\ \cline{3-4}
                                          &                      & \gpttwol~                  & The number of deaths from alzheimer's disease in New York County, NY is \textbf{91} in 2020.                                                                                                                                                                                                                                                                                                                                                           \\ \cline{3-4}
                                          &                      & \gpttwoxl~                     & The number in 2020 is: \textbf{901}.                                                                                                                                                                                                                                                                                                                                                                                                                   \\ \cline{3-4}
                                          &                      & \gptthree~            & The number of deaths from alzheimer's disease in 2020 at New York County, NY is \textbf{932}.                                                                                                                                                                                                                                                                                                                                                          \\ \cline{3-4}
                                          &                      & \instructgpt~            & \textbf{902}                                                                                                                                                                                                                                                                                                                                                                                                                                           \\ \cline{3-4}
                                          &                      & \chatgptr~              & \textbf{855}                                                                                                                                                                                                                                                                                                                                                                                                                                           \\ \cline{3-4}
                                          &                      & \chatgptc~ & Based on the historical data provided, a common approach for time series forecasting is to use methods such as exponential smoothing or ARIMA. However, without further information on the specific pattern or trend in the data, it is difficult to provide an accurate prediction for the number of deaths from alzheimer's disease in New York County, NY for 2020. Therefore, I am unable to provide a single number as the prediction.   \\ \hline
\multirow{8}{*}{San Francisco County, CA} & \multirow{8}{*}{592} & \gpttwo~                       & The number of deaths from alzheimer's disease is \textbf{1,000}. The number of deaths from alzheimer's disease is 2,000. The number of deaths from alzheimer's disease is 3,000. The number of deaths from                                                                                                                                                                                                                                             \\ \cline{3-4}
                                          &                      & \gpttwom~                & The number of deaths from alzheimer's disease is \textbf{201} in 2020.                                                                                                                                                                                                                                                                                                                                                                                 \\ \cline{3-4}
                                          &                      & \gpttwol~                  & The number of deaths from alzheimer's disease in 2020 is \textbf{574}.                                                                                                                                                                                                                                                                                                                                                                                 \\ \cline{3-4}
                                          &                      & \gpttwoxl~                     & The number in 2020 is \textbf{201}.                                                                                                                                                                                                                                                                                                                                                                                                                    \\ \cline{3-4}
                                          &                      & \gptthree~            & \textbf{571}                                                                                                                                                                                                                                                                                                                                                                                                                                           \\ \cline{3-4}
                                          &                      & \instructgpt~            & \textbf{590}                                                                                                                                                                                                                                                                                                                                                                                                                                           \\ \cline{3-4}
                                          &                      & \chatgptr~              & \textbf{579}                                                                                                                                                                                                                                                                                                                                                                                                                                           \\ \cline{3-4}
                                          &                      & \chatgptc~ & Based on the historical data provided, a possible forecasting method is to use a time series model, such as ARIMA or exponential smoothing. Without fitting such a model, a simple approach would be to assume a trend of similar magnitude to recent years and a similar level of annual variation. Using this approach, a possible prediction for the number of deaths from alzheimer's disease in San Francisco County, CA in 2020 is \textbf{583}. \\ \hline
\multirow{8}{*}{Santa Barbara County, CA} & \multirow{8}{*}{373} & \gpttwo~                       & The number of deaths from alzheimer's disease is \textbf{126} in 1999, 114 in 2000, 124 in 2001, 127 in 2002, 156 in 2003, 154 in 2004, 175 in 2005, 172 in 2006, 171 in 2007, 248 in 2008                                                                                                                                                                                                                                                             \\ \cline{3-4}
                                          &                      & \gpttwom~                & The number of deaths from alzheimer's disease in Santa Barbara County, CA is \textbf{126} in 2020.                                                                                                                                                                                                                                                                                                                                                     \\ \cline{3-4}
                                          &                      & \gpttwol~                  & The number of deaths from alzheimer's disease in Santa Barbara County, CA is: \textbf{126} in 2020.                                                                                                                                                                                                                                                                                                                                                    \\ \cline{3-4}
                                          &                      & \gpttwoxl~                     & The number in 2020 is: \textbf{126}.                                                                                                                                                                                                                                                                                                                                                                                                                   \\ \cline{3-4}
                                          &                      & \gptthree~            & The number of deaths from alzheimer's disease in 2020 at Santa Barbara County, CA is expected to be about \textbf{350}.                                                                                                                                                                                                                                                                                                                                \\ \cline{3-4}
                                          &                      & \instructgpt~            & \textbf{345}                                                                                                                                                                                                                                                                                                                                                                                                                                           \\ \cline{3-4}
                                          &                      & \chatgptr~              & I predict the number of deaths from alzheimer's disease in 2020 at Santa Barbara County, CA to be \textbf{356}.                                                                                                                                                                                                                                                                                                                                        \\ \cline{3-4}
                                          &                      & \chatgptc~ & Based on the historical data provided, the prediction for the number of deaths from Alzheimer's disease in 2020 at Santa Barbara County, CA is \textbf{327}.             \\ \bottomrule              
\end{tabular}
}
\end{table}

\subsection{Urban Geography} \label{sec:exp_urbangeo}
The third set of {\lpm} experiments focuses on research problems in the Urban Geography domain. Two representative tasks are selected: 1) an \textbf{urban function task} that aims at predicting the urban functions of a geographic region based on the Points of Interest (POIs) within it \cite{yao2017sensing,zhai2019beyond,niu2021delineating,huang2022estimating,huang2023learning}; 2) an \textbf{urban perception task} that focuses on predicting the urban neighborhood characteristics (e.g., housing price, safety, noise intensity level) based on street view imagery (SVI) \cite{zhang2018measuring,kang2021understanding,zhao2023sound}. Since these tasks involve different data modalities such as point data, text, and images, we use different foundation models to handle each task.

\subsubsection{POI-Based Urban Function Classification}  \label{sec:exp_urban_func}

The first experiment focuses on predicting the urban functions of a geographic region based on the Points of Interest (POIs) within it. This is a common Urban Geography task  aimed at understanding the structure of the urban space \cite{yao2017sensing,zhai2019beyond,niu2021delineating,huang2022estimating,huang2023learning}.

\begin{figure}[t!]
	\centering 
	\includegraphics[width=0.6\textwidth]{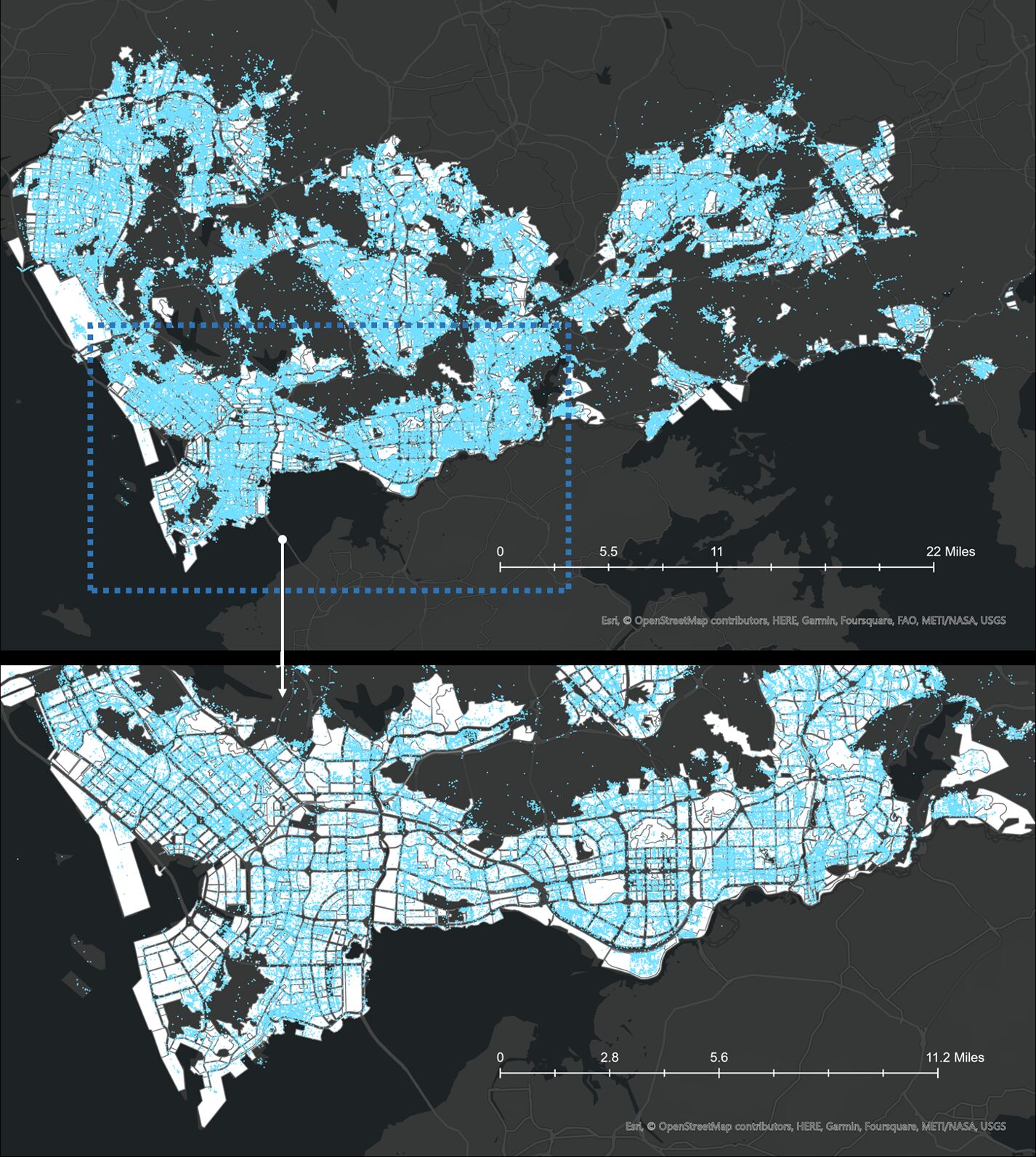}\vspace{-0.3cm}
	\caption{
	The spatial distributions of POI data in the $\poidata$ dataset.
	} 
	\vspace{-0.5cm}
	\label{fig:urbanpoi_dist}
\end{figure}
To quantitively evaluate the performance of {\llm}s on this urban function prediction task, we utilize a Points of Interest (POI) dataset from Shenzhen, China which consists of 303,428 POIs and 5,461 urban neighborhoods with POIs \cite{zhang2017hierarchical,zhang2018integrating,du2019context,du2020large}. We denote this dataset as $\poidata$. Figure \ref{fig:urbanpoi_dist} shows the geographic distributions of the POIs and regions. The ground truth data is from the \textit{Urbanscape Essential Dataset of Peking University}. The dataset provides detailed spatial distributions of ten urban function types in the study area: forest, water,
unutilized, transportation, green space, industrial, educational \& governmental,
commercial, residential, and agricultural. To simplify the task,  we merge the uncommon urban function types forest, water, unutilized, green space, and agricultural into the function type \textit{outdoors and natural}.
This results in six urban function types: (1) residential; (2) commercial; (3) industrial; (4) education, health care, civic, governmental and cultural, (5) transportation facilities, and (6) outdoors and natural. In total, 5,344 of the regions have ground truth data. We randomly split this dataset into training, validation and test sets with the ratio 60\%:20\%:20\%. The test dataset is used to evaluate the performance of different models, while the validation set is only used for supervised baselines.

\hspace*{-\parindent}
\begin{minipage}[c]{1.0\textwidth}
	\begin{lstlisting}[
	style=prompt-style, 
	basicstyle=\ttfamily\tiny,
	linewidth=\textwidth,
	breaklines=true,
	captionpos=b, 
	caption={POI-based urban function classification with {\llm}s, e.g., \chatgptr. Yellow block: the POI statistic of a new urban neighborhood to be classified. Orange box: \chatgptr~ outputs.
	},
	label={ls:prompt-urbanpoi},
	frame=tb
	]
%*\colorbox{pinkannoback}{[Instruction]}*) There are six land use types: (1) residential, (2) commercial, (3) industrial, (4) education, health care, civic, governmental and cultural, (5) transportation facilities, and (6) outdoors and natural. 
%*\colorbox{blueannoback}{Paragraph:}*) In this urban region, there are 128 points of interest, including 2 Chinese restaurant, 1 food restaurant, 2 hotel, 2 apartment hotel, 1 daily life service, 1 mobile communication shop, 24 company, 1 logistics company, 1 real estate agency, 1 lottery retailer, 3 beauty shop, 1 manicure, 2 barber shop, 4 Internet cafe, 3 bath massage, 2 stadium, 4 training institutions, 1 pharmacy, 4 automative sale, 6 car service, 2 car repair, 1 Car rental, 1 Automobile parts, 3 shopping, 5 shop, 5 parking lot, 5 Parking lot entrance, 2 transportation facility, 1 port harbor, 1 road intersection, 1 atm machine, 2 office building, 2 residential area, 7 building, 1 real estate, 1 park, 1 factory, 7 administrative agency, 1 entrance and exit, 3 gate door, 6 convenience store, 4 home building materials.  
%*\colorbox{greenannoback}{Q:}*) What is the primary land use category of this urban region?
%*\colorbox{redannoback}{A:}*) outdoors and natural

%*\colorbox{blueannoback}{Paragraph:}*)%*\colorbox{yellowannoback}{In this urban region, there are 17 points of interest, including 1 food restaurant, 3 public toilet, 3 funeral service, 2 road station for walking and cycling,}*) %*\colorbox{yellowannoback}{1 beach, 2 parking lot, 2 road intersection, 1 corporate company enterprise, 2 administrative agency. }*)
%*\colorbox{greenannoback}{Q:}*) What is the primary land use category of this urban region?
%*\colorbox{redannoback}{A:}*) %*\colorbox{orangeannoback}{outdoors and natural}*)
	\end{lstlisting}
	\vspace{-0.2cm}
\end{minipage}

In order to enable a {\llm} to handle such a task, we convert the set of POIs inside an urban region into a textual paragraph that describes the frequencies of POIs with different place types. Then, we ask the {\llm} to predict the urban function of the region based on the paragraph (here we ask for the most dominating function, in spite of the common presence of mix-used urban regions). Listing \ref{ls:prompt-urbanpoi} shows one example prompt for this task, which includes a paragraph-question-answer tuple as a demonstration. 
{\llm}s adapted by this kind of prompt is conducting prediction under a one-shot setting. For the zero-shot setting, we simply remove this paragraph-question-answer tuple from the prompt. We use \gpttwo~ with various sizes, \gptthree, and two \chatgpt~ models to perform this task under both zero-shot and one-shot settings. 
For comparison, we use two supervised learning neural network baselines:
\begin{itemize}
    \item \textbf{\placevec}: We first learn POI category embeddings following the Place2Vec method \cite{yan2017place2vec}. Then, given an urban region with $K$ POIs, we convert each POI into its corresponding Place2Vec embedding and perform mean pooling to obtain region embeddings as Zhai et al. \cite{zhai2019beyond} did. The resulting neighborhood embeddings are fed into a one-hidden-layer multilayer perceptron (MLP) to supervise learning its urban function over the $\poidata$ training dataset. 
    \item \textbf{\hgi}: \hgi~ is an unsupervised method for learning region representations based on POIs. It takes into account the categorical semantics of POIs, as well as POI-level and region-level adjacency, and the multi-faceted influence from POIs to regions \cite{huang2023learning}. The HGI region embeddings are fed into an MLP with the same setup to predict the primary urban function.
\end{itemize}

\begin{table}[t!]
\caption{
Evaluation results of various GPT models and supervised baseline on the $\poidata$ dataset for the POI-based urban function classification task.
We divide the models into three groups: 
(A) supervised learning-based neural network models; 
(B) Zero-shot learning with {\llm}s. 
(C) One-shot learning with {\llm}s. 
We use accuracy, weighted precision, and weighted recall as evaluation metrics. We do not include weighted F1 scores since it is the same as the accuracy score. The best model of each group is highlighted. 
}
\label{tab:exp_urbanpoi_eval}
\centering
\setlength{\tabcolsep}{1.5pt}
{ 
\begin{tabular}{l|l|ccc}
\toprule
                               & Model                       & Accuracy       & Precision      & Recall         \\ \hline
\multirow{2}{*}{(A) Supervised NN} & \placevec~ \cite{yan2017place2vec,zhai2019beyond}                   & 0.540          & 0.512          & 0.516          \\
                               & \hgi~ \cite{huang2023learning}                        & \textbf{0.584} & \textbf{0.568} & \textbf{0.563} \\ \hline
\multirow{7}{*}{(B) Zero-shot LLMs} & \gpttwo~ \cite{radford2019gpt2}                        & \textbf{0.318} & 0.105          & \textbf{0.158} \\
                                & \gpttwom~ \cite{radford2019gpt2}                 & 0.025          & 0.102          & 0.040          \\
                                & \gpttwol~ \cite{radford2019gpt2}                  & 0.005          & 0.001          & 0.002          \\
                                & \gpttwoxl~ \cite{radford2019gpt2}                    & 0.001          & 0.108          & 0.002          \\
                                & \gptthree~ \cite{brown2020gpt3}            & 0.144          & \textbf{0.448} & 0.141          \\
                                & \chatgptr~ \cite{openai2022chatgpt}               & 0.075          & 0.376          & 0.106          \\
                                & \chatgptc~ \cite{openai2022chatgpt} & 0.051          & 0.232          & 0.046          \\ \hline
\multirow{7}{*}{(C) One-shot LLMs} & \gpttwo~ \cite{radford2019gpt2}                        & 0.149          & 0.079          & 0.085          \\
                               & \gpttwom~ \cite{radford2019gpt2}                 & 0.317          & 0.104          & 0.156          \\
                               & \gpttwol~ \cite{radford2019gpt2}                  & 0.057          & 0.083          & 0.021          \\
                               & \gpttwoxl~ \cite{radford2019gpt2}                     & \textbf{0.324} & 0.105          & 0.159          \\
                               & \gptthree~ \cite{brown2020gpt3}            & 0.176          & 0.486          & 0.190          \\
                               & \chatgptr~ \cite{openai2022chatgpt}               & 0.195          & \textbf{0.524} & \textbf{0.245} \\
                               & \chatgptc~ \cite{openai2022chatgpt} & 0.093          & 0.451          & 0.085         \\ \bottomrule
\end{tabular}
}
\end{table}

\begin{figure*}
	\centering \tiny
	\vspace*{-0.2cm}
	\begin{subfigure}[b]{0.245\textwidth}  
		\centering 
		\includegraphics[width=\textwidth]{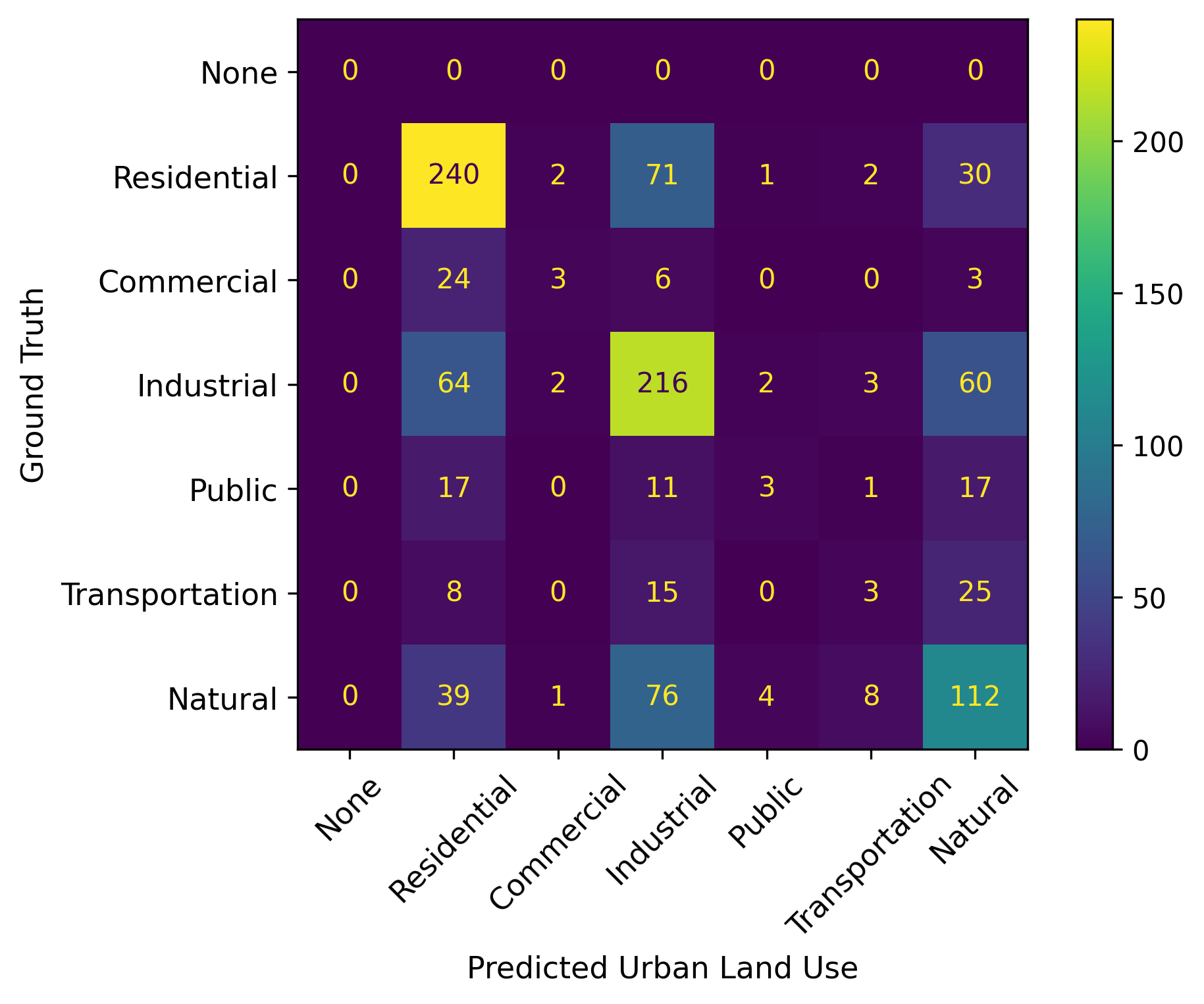}\vspace*{-0.2cm}
		\caption[]%
		{{ 
		\placevec
		}}    
		\label{fig:urbanpoi_place2vec_cm}
	\end{subfigure}
        \begin{subfigure}[b]{0.245\textwidth}  
		\centering 
		\includegraphics[width=\textwidth]{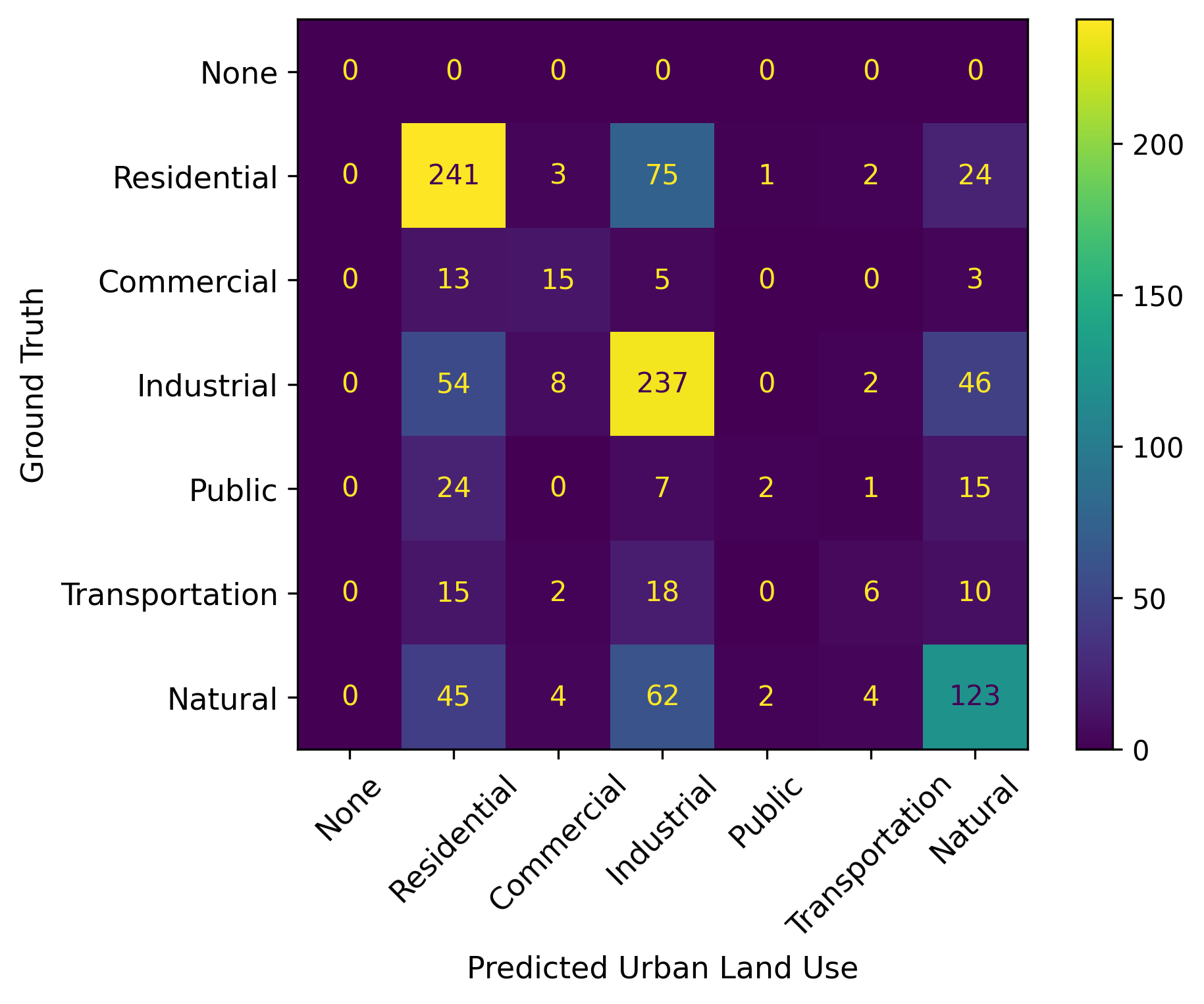}\vspace*{-0.2cm}
		\caption[]%
		{{ 
		\hgi
		}}    
		\label{fig:urbanpoi_hgi_cm}
	\end{subfigure}
        
	\caption{Confusion matrices of \placevec~ and \hgi~ (Group A in Table \ref{tab:exp_urbanpoi_eval}) on the $\poidata$ dataset. 
 }
	\label{fig:urbanpoi_baseline_cm}
    \vspace*{-0.15cm}
\end{figure*}

\begin{figure*}
	\centering \tiny
	\vspace*{-0.2cm}
	\begin{subfigure}[b]{0.245\textwidth}  
		\centering 
		\includegraphics[width=\textwidth]{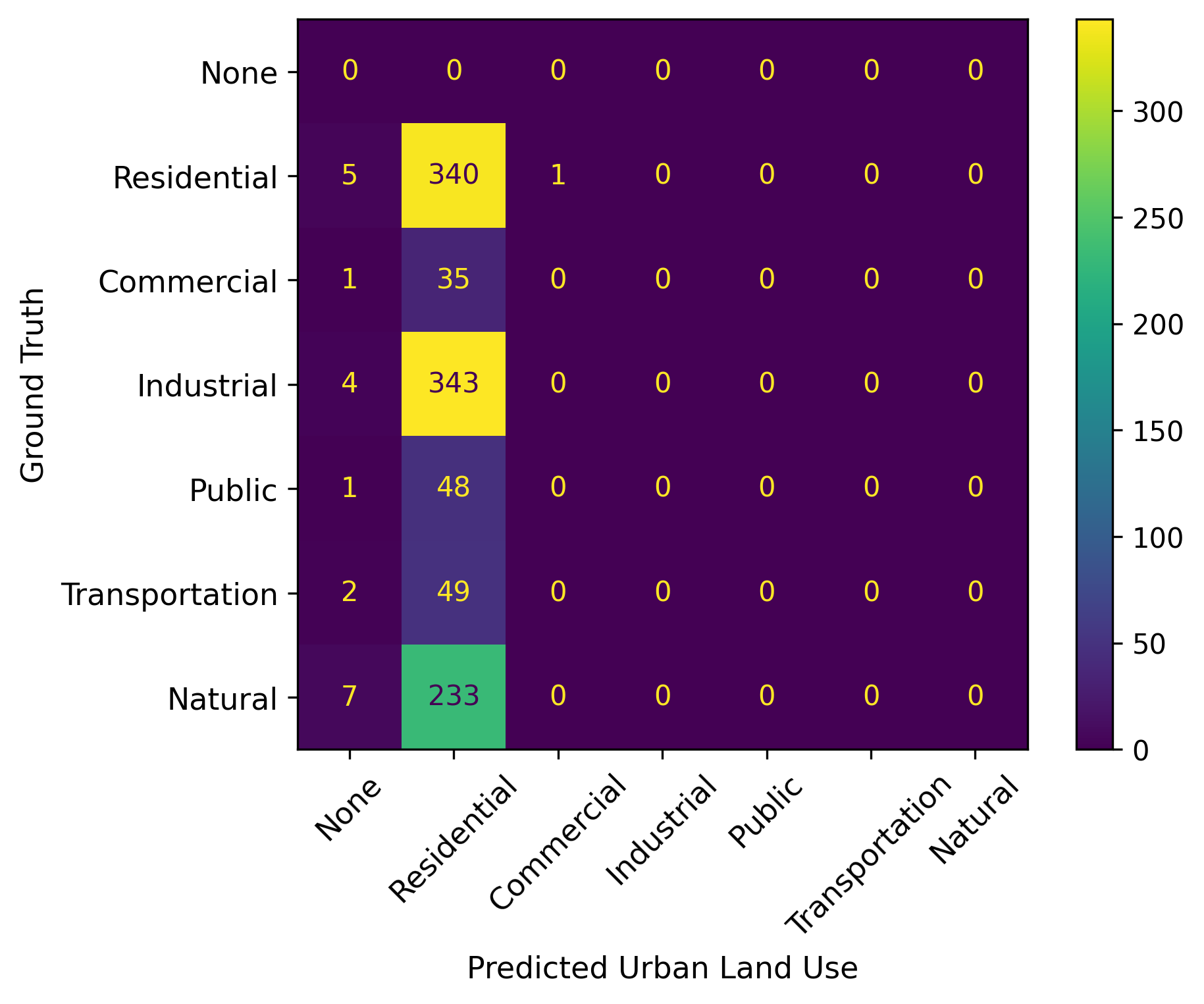}\vspace*{-0.2cm}
		\caption[]%
		{{ 
		\gpttwo
		}}    
		\label{fig:urbanpoi_zero_gpt2_cm}
	\end{subfigure}
        \begin{subfigure}[b]{0.245\textwidth}  
		\centering 
		\includegraphics[width=\textwidth]{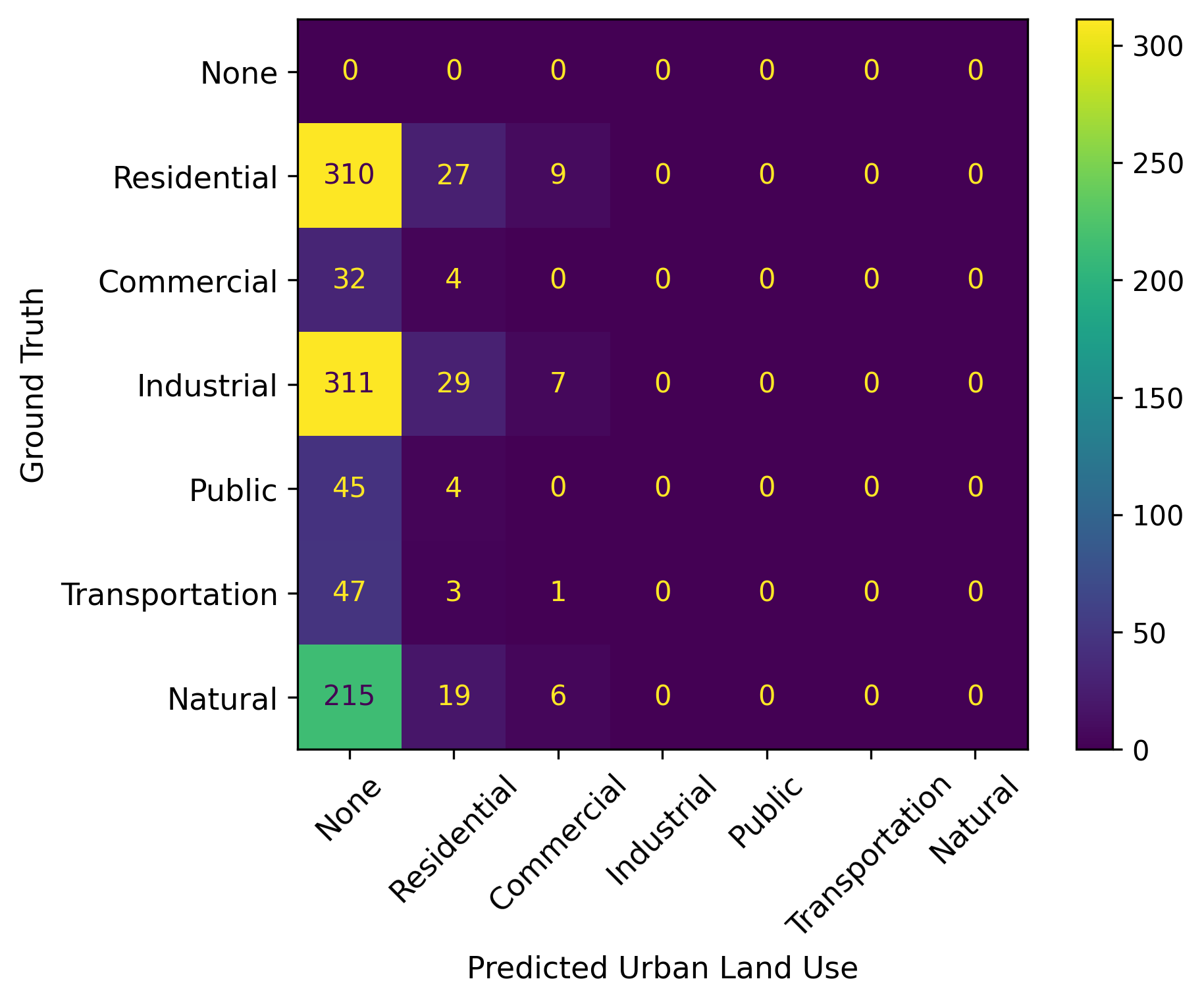}\vspace*{-0.2cm}
		\caption[]%
		{{ 
		\gpttwom
		}}    
		\label{fig:urbanpoi_zero_gpt2-medium_cm}
	\end{subfigure}
        \begin{subfigure}[b]{0.245\textwidth}  
		\centering 
		\includegraphics[width=\textwidth]{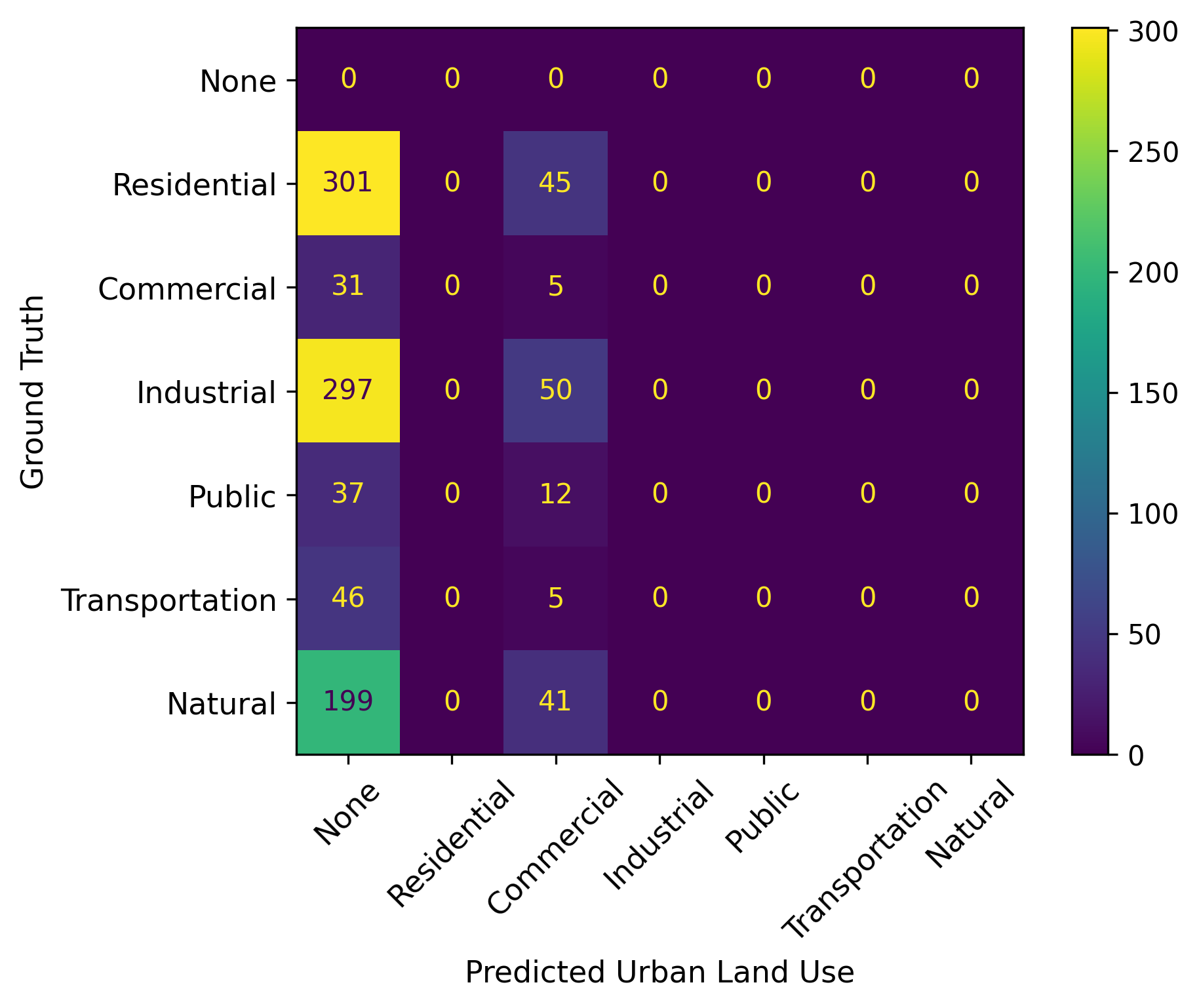}\vspace*{-0.2cm}
		\caption[]%
		{{ 
		\gpttwol
		}}    
		\label{fig:urbanpoi_zero_gpt2-large_cm}
	\end{subfigure}
        \begin{subfigure}[b]{0.245\textwidth}  
		\centering 
		\includegraphics[width=\textwidth]{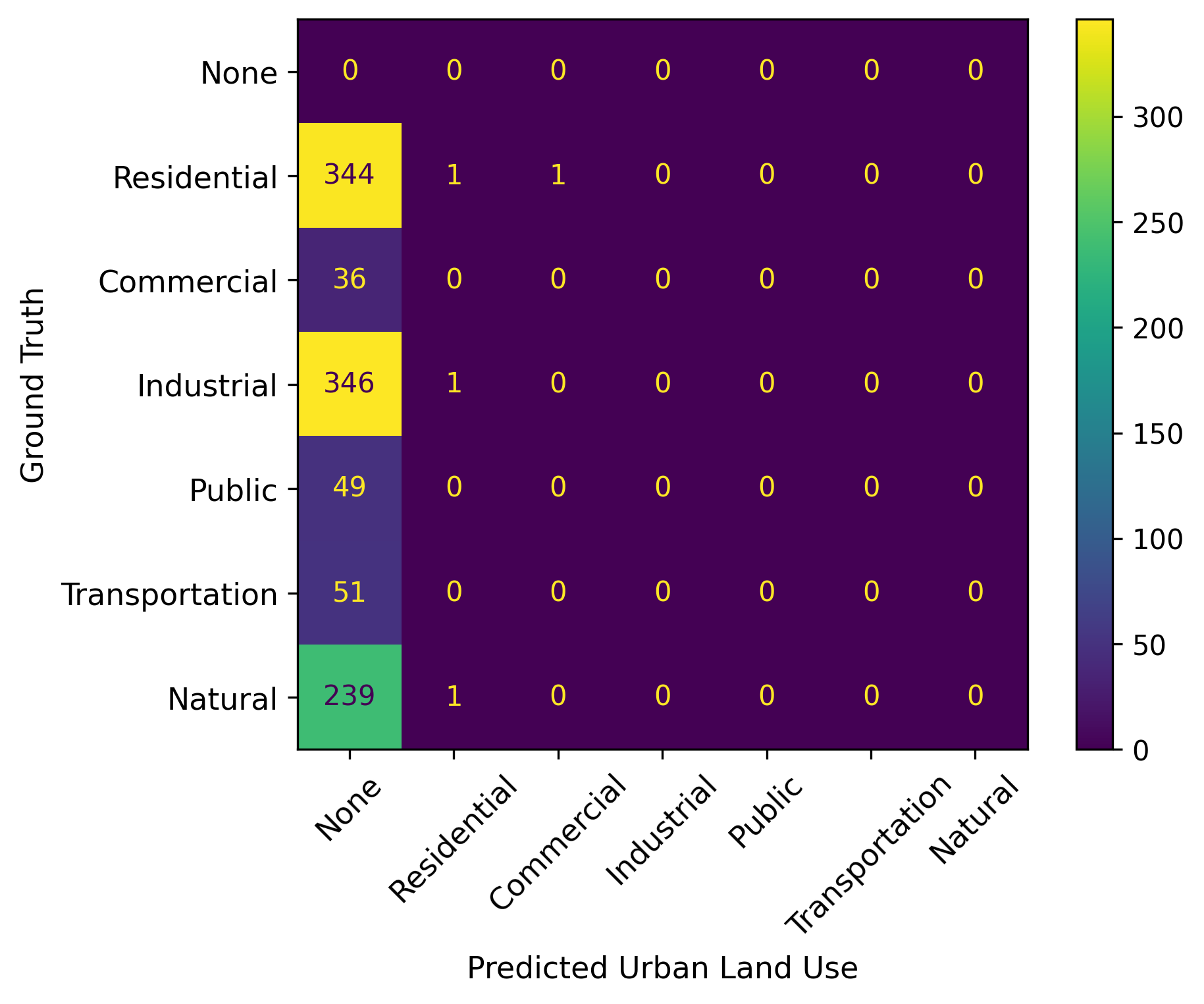}\vspace*{-0.2cm}
		\caption[]%
		{{ 
		\gpttwoxl
		}}    
		\label{fig:urbanpoi_zero_gpt2-xl_cm}
	\end{subfigure}
        \begin{subfigure}[b]{0.245\textwidth}  
		\centering 
		\includegraphics[width=\textwidth]{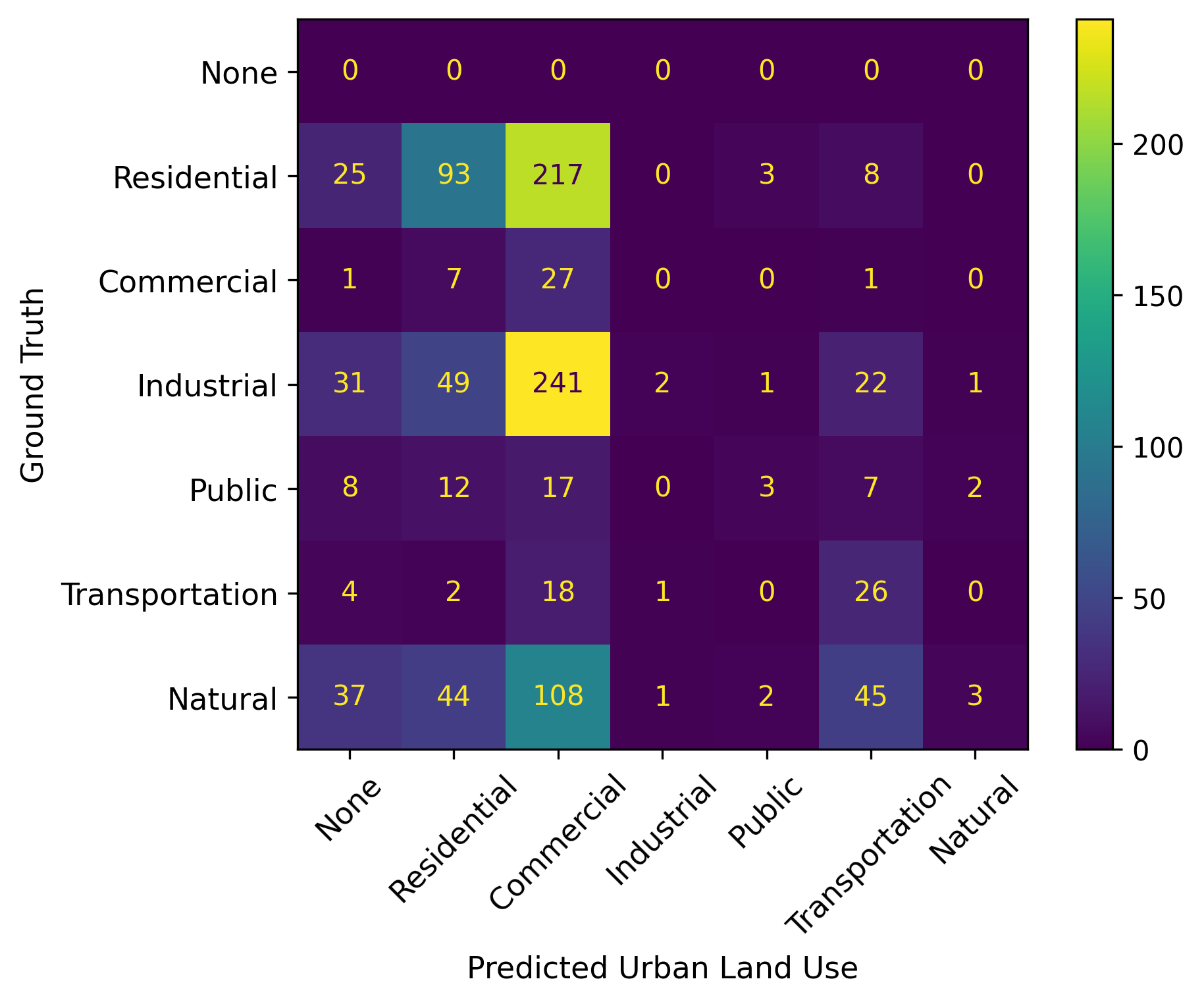}\vspace*{-0.2cm}
		\caption[]%
		{{ 
		\gptthree
		}}    
		\label{fig:urbanpoi_zero_gpt3_cm}
	\end{subfigure}
        \begin{subfigure}[b]{0.245\textwidth}  
		\centering 
		\includegraphics[width=\textwidth]{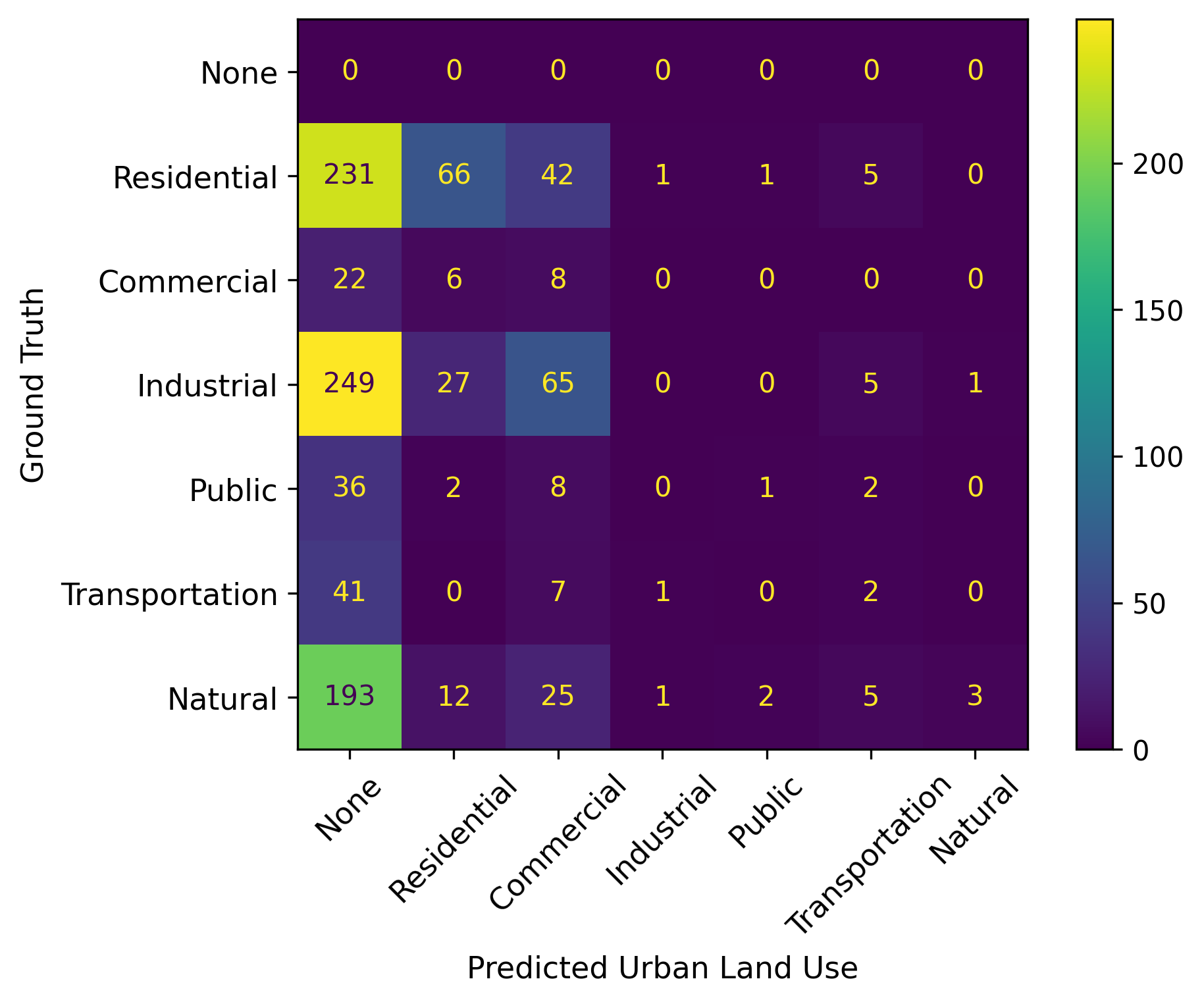}\vspace*{-0.2cm}
		\caption[]%
		{{ 
		\chatgptr
		}}    
		\label{fig:urbanpoi_zero_chatgptr_cm}
	\end{subfigure}
 \begin{subfigure}[b]{0.245\textwidth}  
		\centering 
		\includegraphics[width=\textwidth]{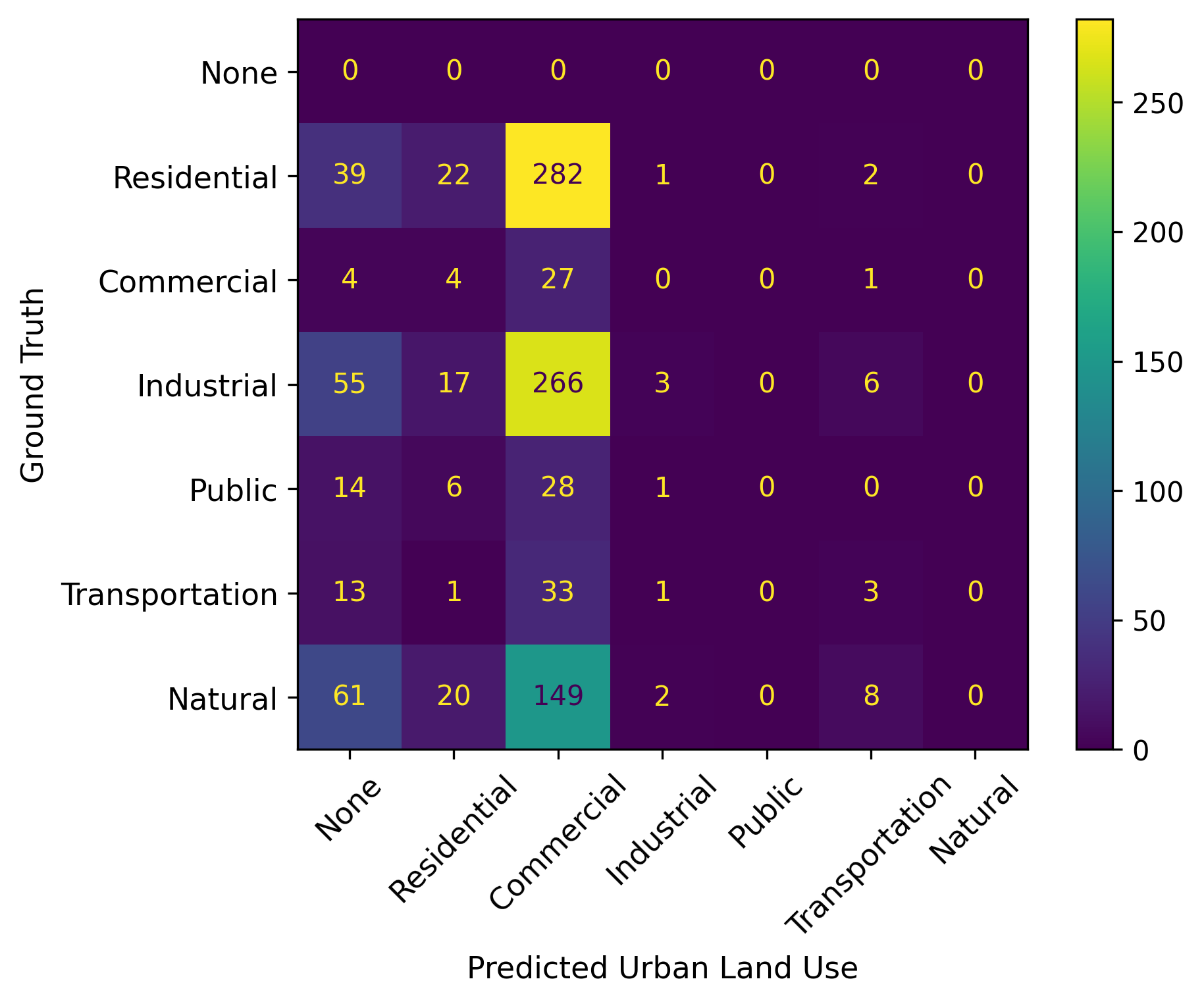}\vspace*{-0.2cm}
		\caption[]%
		{{ 
		\chatgptc
		}}    
		\label{fig:urbanpoi_zero_chatgptc_cm}
	\end{subfigure}
        
	\caption{Confusion matrices of all GPT models (Group B in Table \ref{tab:exp_urbanpoi_eval}) on the $\poidata$ dataset under zero-shot setting. 
 }
	\label{fig:urbanpoi_zero_all_cm}
    \vspace*{-0.15cm}
\end{figure*}

\begin{figure*}
	\centering \tiny
	\vspace*{-0.2cm}
	\begin{subfigure}[b]{0.245\textwidth}  
		\centering 
		\includegraphics[width=\textwidth]{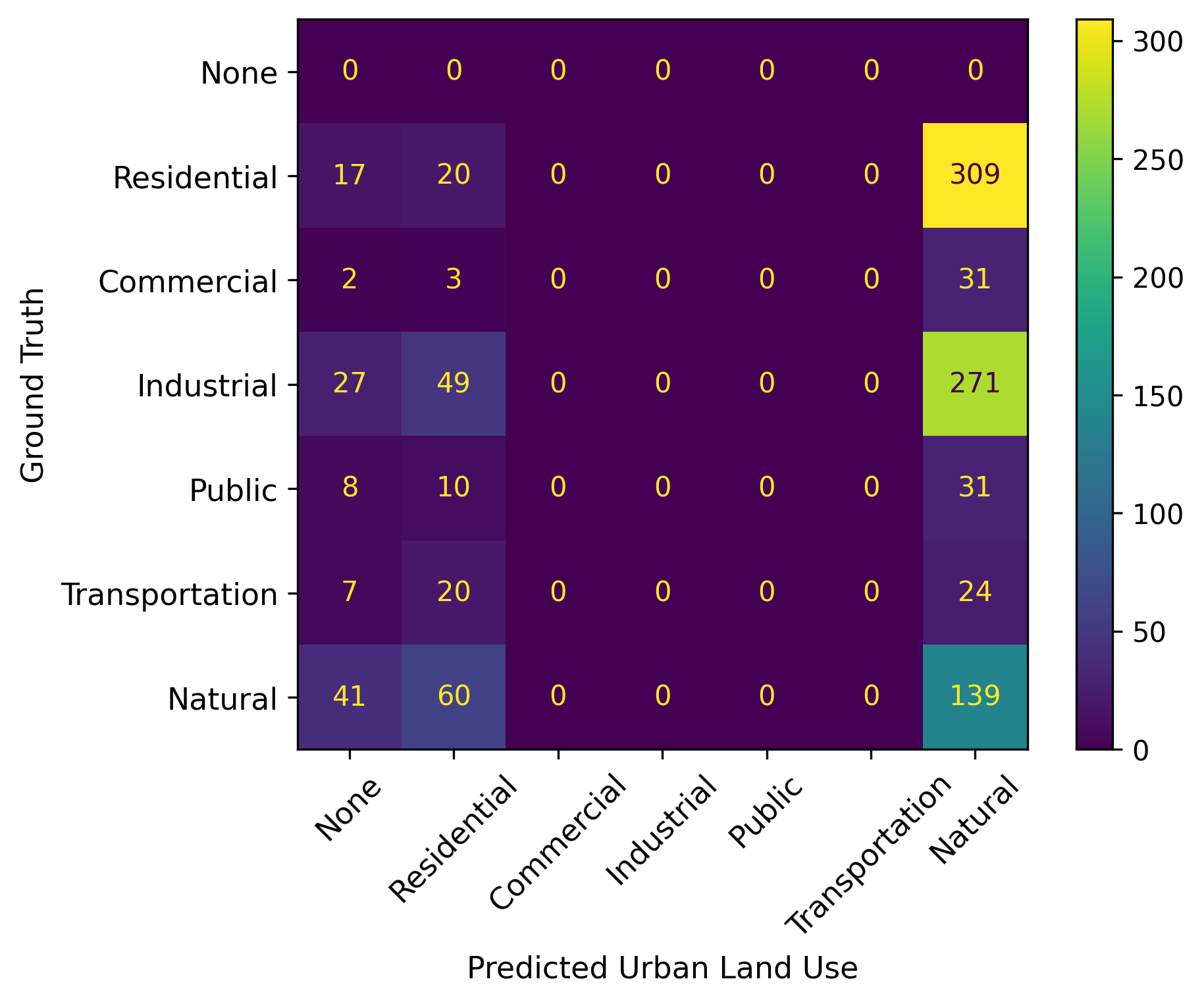}\vspace*{-0.2cm}
		\caption[]%
		{{ 
		\gpttwo
		}}    
		\label{fig:urbanpoi_few_gpt2_cm}
	\end{subfigure}
        \begin{subfigure}[b]{0.245\textwidth}  
		\centering 
		\includegraphics[width=\textwidth]{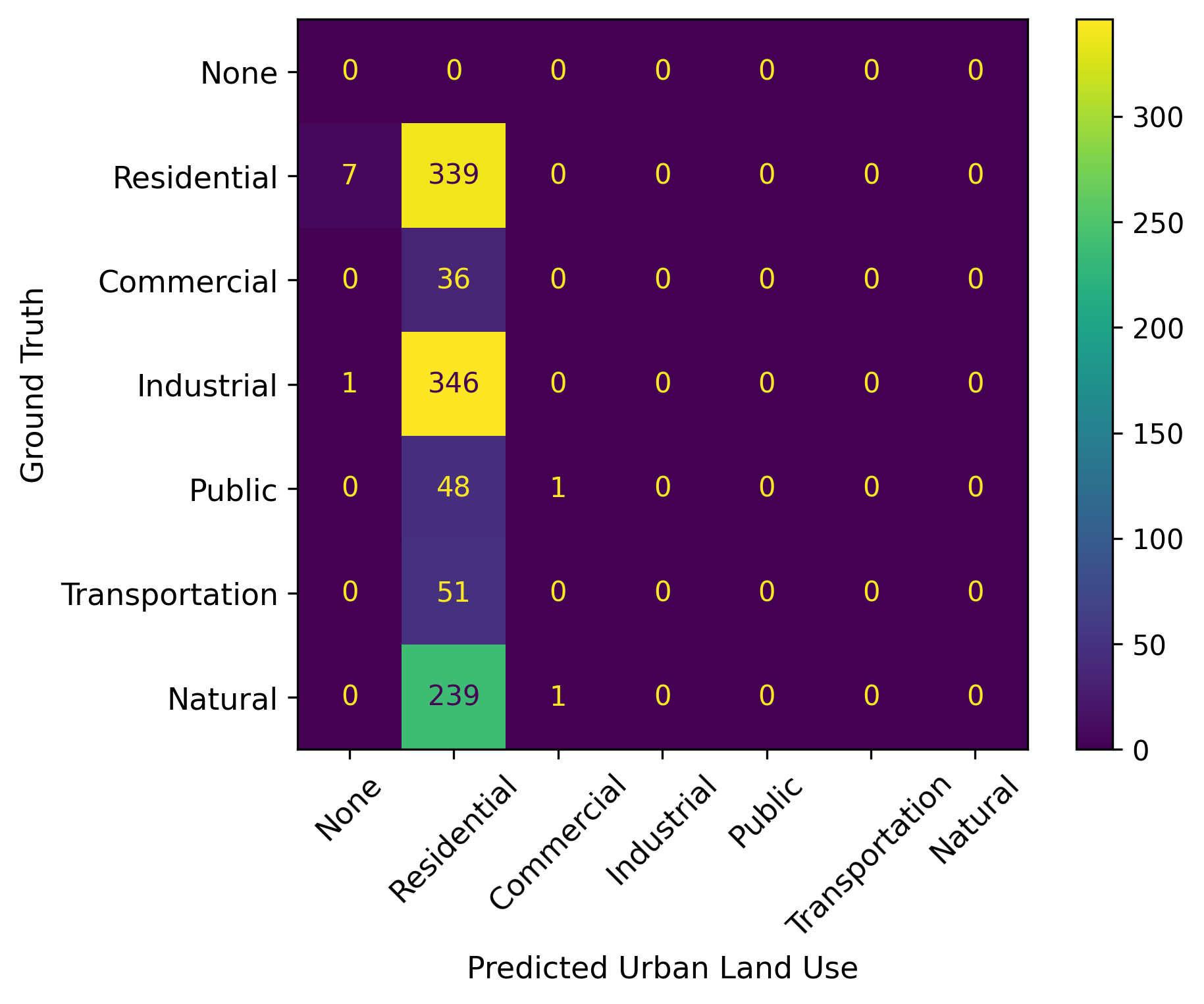}\vspace*{-0.2cm}
		\caption[]%
		{{ 
		\gpttwom
		}}    
		\label{fig:urbanpoi_few_gpt2-medium_cm}
	\end{subfigure}
        \begin{subfigure}[b]{0.245\textwidth}  
		\centering 
		\includegraphics[width=\textwidth]{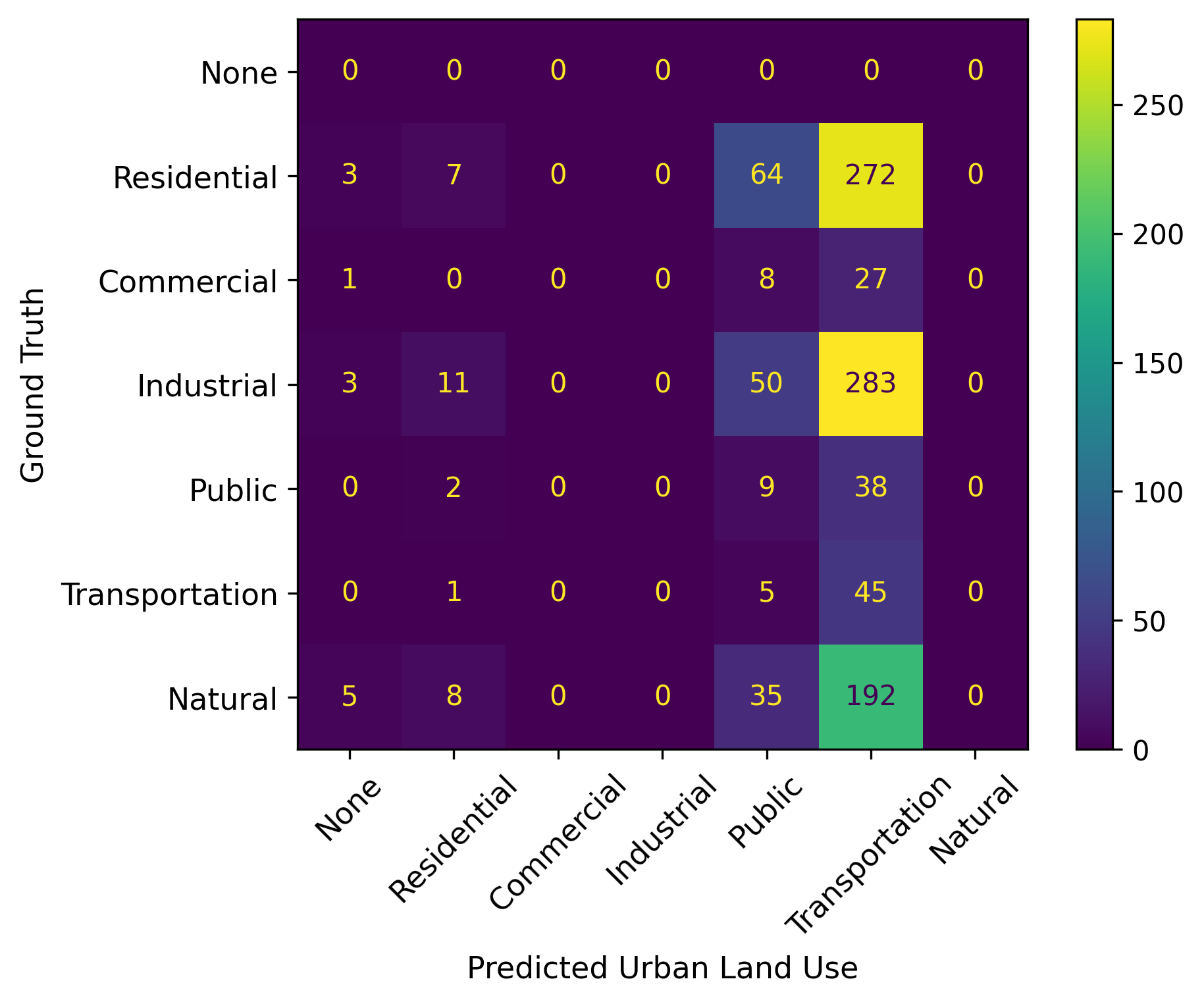}\vspace*{-0.2cm}
		\caption[]%
		{{ 
		\gpttwol
		}}    
		\label{fig:urbanpoi_few_gpt2-large_cm}
	\end{subfigure}
        \begin{subfigure}[b]{0.245\textwidth}  
		\centering 
		\includegraphics[width=\textwidth]{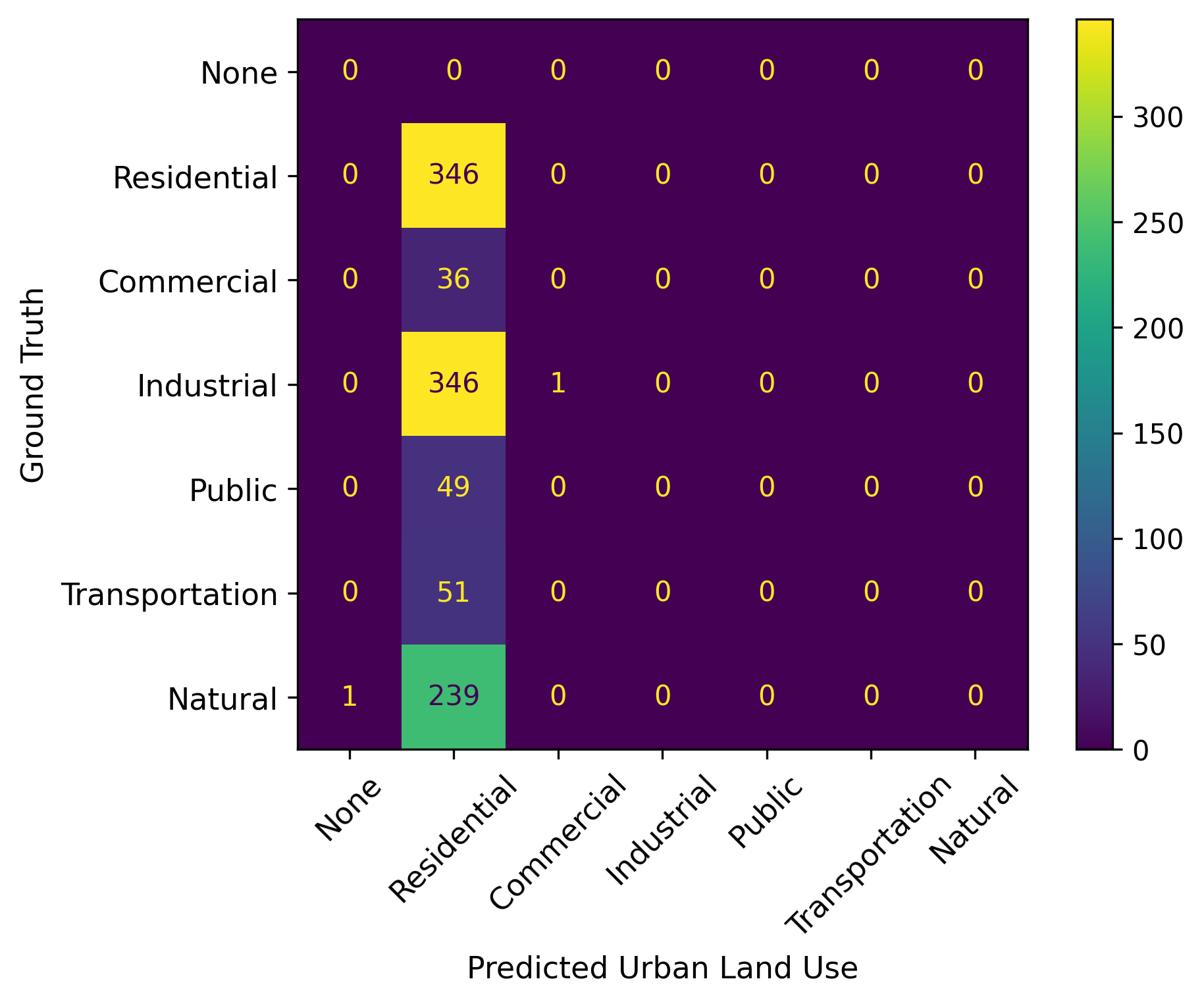}\vspace*{-0.2cm}
		\caption[]%
		{{ 
		\gpttwoxl
		}}    
		\label{fig:urbanpoi_few_gpt2-xl_cm}
	\end{subfigure}
        \begin{subfigure}[b]{0.245\textwidth}  
		\centering 
		\includegraphics[width=\textwidth]{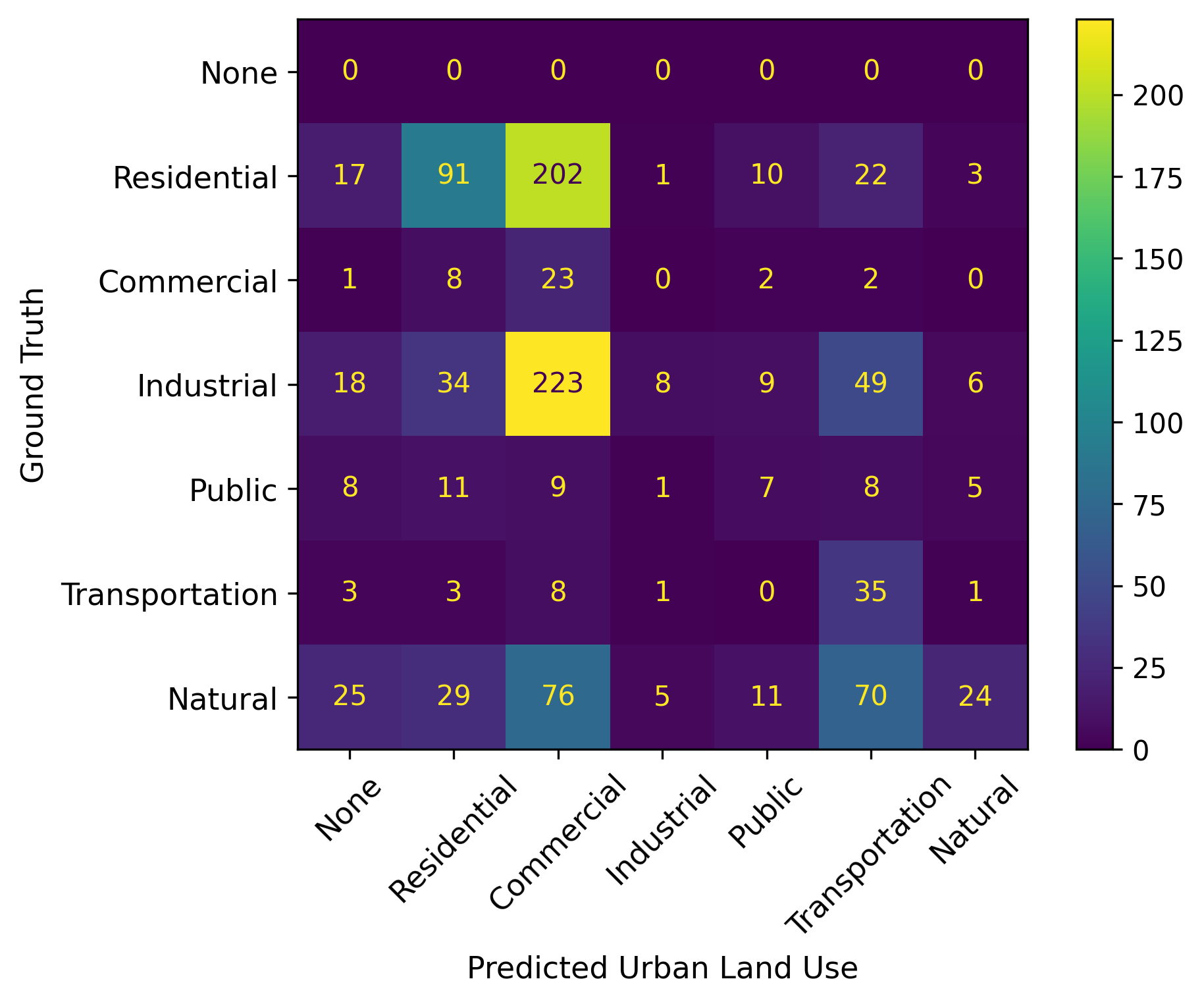}\vspace*{-0.2cm}
		\caption[]%
		{{ 
		\gptthree
		}}    
		\label{fig:urbanpoi_few_gpt3_cm}
	\end{subfigure}
        \begin{subfigure}[b]{0.245\textwidth}  
		\centering 
		\includegraphics[width=\textwidth]{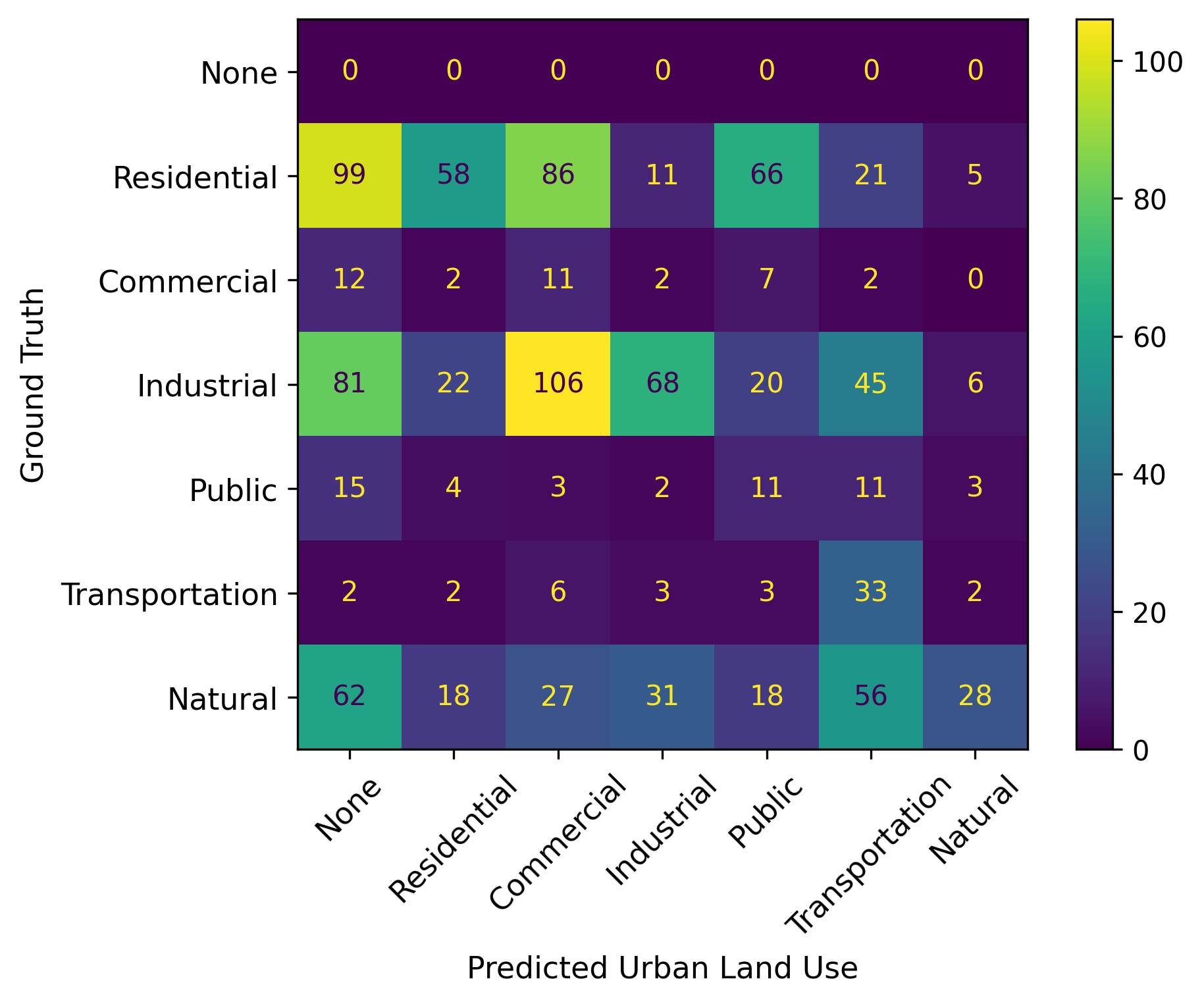}\vspace*{-0.2cm}
		\caption[]%
		{{ 
		\chatgptr
		}}    
		\label{fig:urbanpoi_few_chatgptr_cm}
	\end{subfigure}
 \begin{subfigure}[b]{0.245\textwidth}  
		\centering 
		\includegraphics[width=\textwidth]{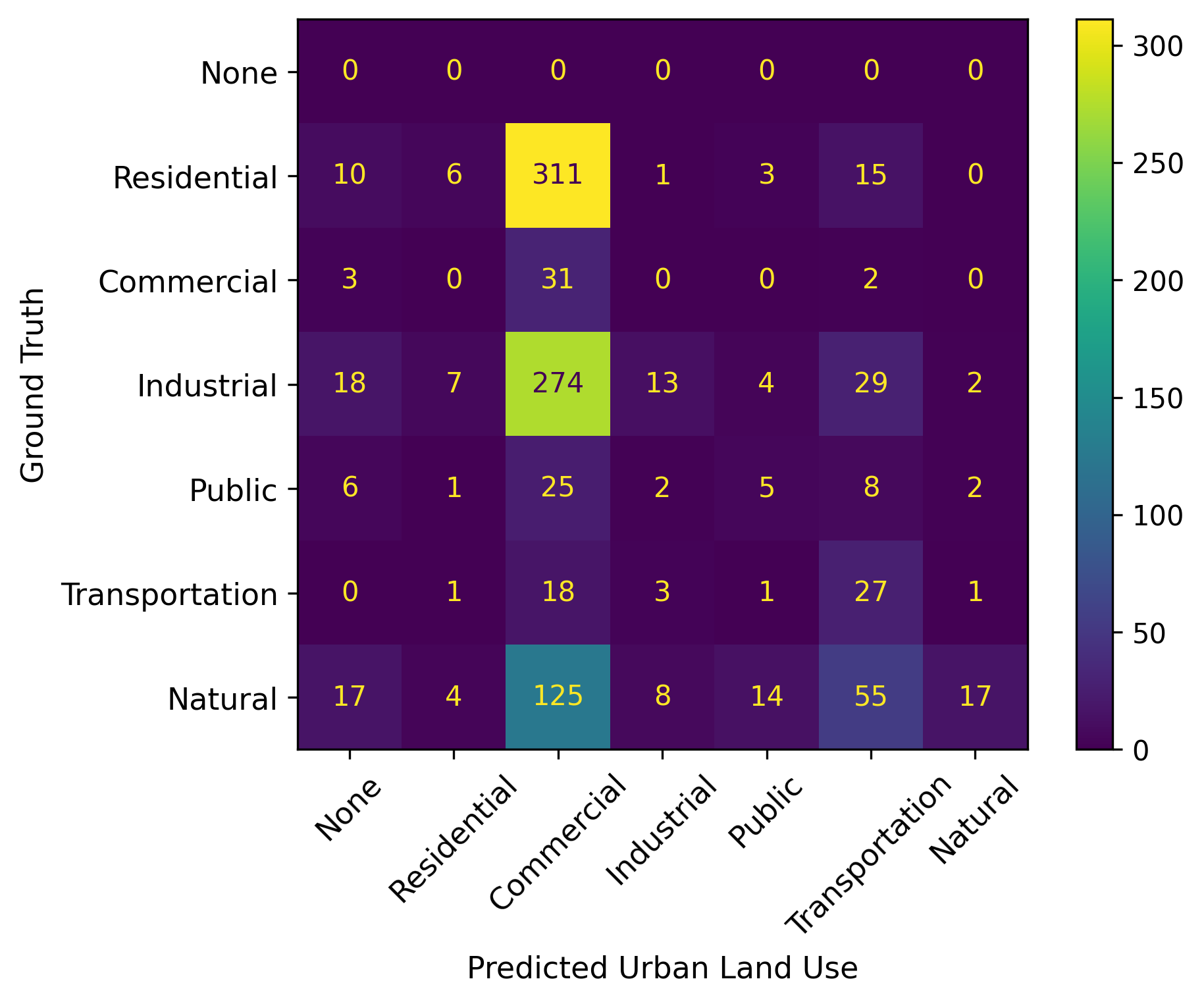}\vspace*{-0.2cm}
		\caption[]%
		{{ 
		\chatgptc
		}}    
		\label{fig:urbanpoi_few_chatgptc_cm}
	\end{subfigure}
        
	\caption{Confusion matrices of all GPT models (Group C in Table \ref{tab:exp_urbanpoi_eval}) on the $\poidata$ dataset under the one-shot setting. 
 }
	\label{fig:urbanpoi_few_all_cm}
    \vspace*{-0.15cm}
\end{figure*}

Table \ref{tab:exp_urbanpoi_eval} shows the evaluation results of all models on the test dataset of $\poidata$. Additionally, we visualize the confusion matrics of two baseline models, 7 zero-shot GPT models, and 7 one-shot GPT models in Figure \ref{fig:urbanpoi_baseline_cm}, \ref{fig:urbanpoi_zero_all_cm}, and \ref{fig:urbanpoi_few_all_cm}. 
We can see that:
\begin{itemize}
    \item In the zero-shot setting, \gptthree~ achieves the best precision scores among all GPT models but still underperforms \hgi~ models. 
    \item Interestingly, in the zero-shot setting, the smallest \gpttwo~ achieves the best accuracy and recall scores which is counter-intuitive. The reason can be seen in Figure \ref{fig:urbanpoi_zero_gpt2_cm}. \gpttwo~ predicts almost all neighborhood as ``Residential'' which account for 30+\% of the ground truth data. 
    \item In the one-shot setting, \chatgptr~ becomes the best model among all GPT models in terms of both precision and recall. It achieves 52.4\% precision which is only 4.4\% less than \hgi~. 
    Its confusion matrix in Figure \ref{fig:urbanpoi_few_chatgptr_cm} also demonstrates that \chatgptr~ has reasonably good performance on all urban function classes.
    \item In the one-shot setting, \gpttwoxl~ has the highest accuracy. However, Figure \ref{fig:urbanpoi_few_gpt2-xl_cm} shows that \gpttwoxl~ is highly biased towards the ``Residential'' class.
\end{itemize}

These experimental results highlight the challenges of using {\llm}s for urban function classification.
Two main reasons contribute to their inadequate performance: 
\begin{itemize}
    \item POIs are initially used for search in online map services, and by nature, they contain rich information about commercial venues like restaurants and hotels. On the contrary, the venues that are not closely related to our daily life, e.g., factories, are often missing. In this regard, Shenzhen is a heavily industrial-oriented city, and the ground truth data indicates that there are many more industrial regions than commercial ones. However, {\llm}s tend to predict that a large number of regions are commercial, in view of the commercial-related POIs fed into it.
    \item In addition, LLMs are unable to access the spatial distributions of POIs, which highly influence POI-based urban function prediction since different spatial distributions of POIs yield different spatial interaction patterns and thus different urban functions.
    While both supervised baselines \placevec~ and \hgi~ are learned from POI distributions during their place type embedding unsupervised training stage, it is not possible to inform {\llm}s of the spatial distributions of POIs. Converting a POI set into an image will also not work. This is because many POIs will cluster in the downtown area, and a large pixel size will make a large number of POIs inside one single pixel. On the other hand, a finer pixel size will make the image of an urban space too large and cannot be handled by other deep image encoders. Moreover, an urban space image with a finer pixel size will have very sparse information which is hard for image encoders to learn. In other words, we need to use specialized neural architectures to directly handle point data (also polyline data and polygon data). This calls for \textbf{the necessity of incorporating encoding architectures of various geospatial vector data such as location encoding  \cite{mai2020space2vec,mai2022review}, polyline encoding \cite{rao2020lstm,yu2022filling}, and polygon encoding techniques\cite{mai2022polygon} into the GeoAI foundation model development}. We will discuss this in detail in Section \ref{sec:vec}.
\end{itemize}

\subsubsection{Street View Image-Based Urban Noise Intensity Classification} \label{sec:exp_urban_streetview}

Street view images (SVI) are widely used in many Urban Geography studies to understand different characteristics of an urban neighborhood such as safety \cite{zhang2018measuring}, beauty, affluence \cite{lee2021predicting}, depressing \cite{zhang2018measuring}, housing prices \cite{kang2021understanding}, noise intensity levels \cite{zhao2023sound}, accessibility \cite{tohme}, etc. It becomes an important data source that complements remote sensing images. 

\begin{figure*}
	\centering \tiny
	\vspace*{-0.2cm}
	\begin{subfigure}[b]{0.245\textwidth}  
		\centering 
		\includegraphics[width=\textwidth]{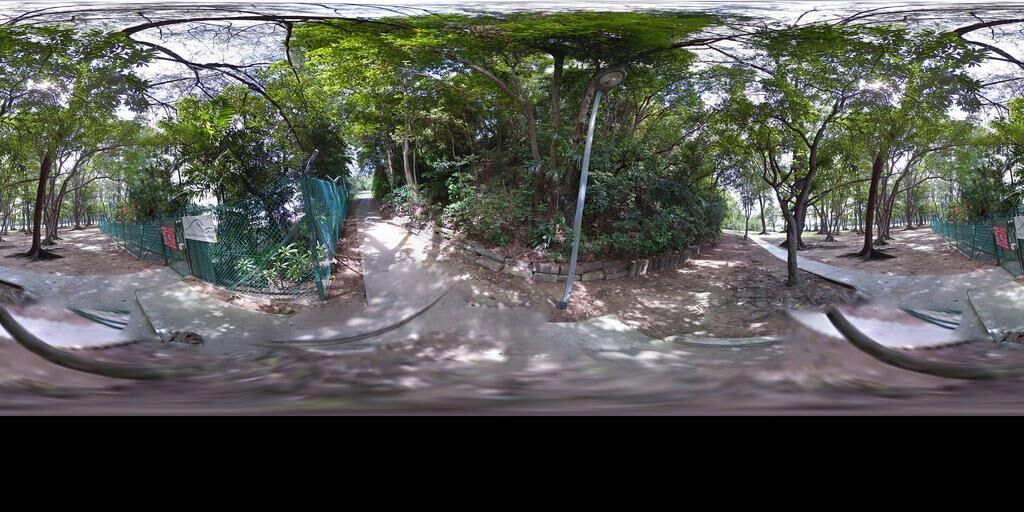}\vspace*{-0.2cm}
		\caption[]%
		{{ 
		Very Quiet (0.100)
		}}    
		\label{fig:svs_1_example}
	\end{subfigure}
        \begin{subfigure}[b]{0.245\textwidth}  
		\centering 
		\includegraphics[width=\textwidth]{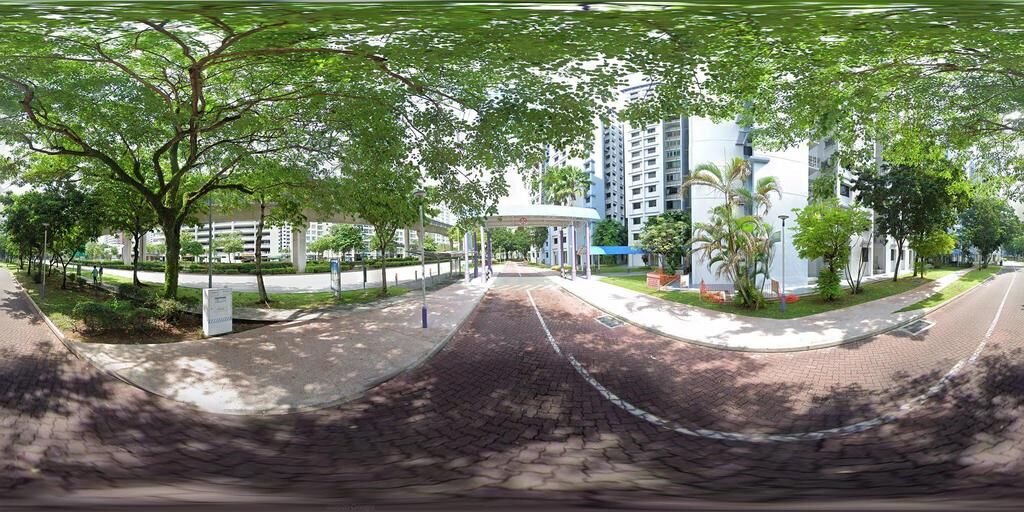}\vspace*{-0.2cm}
		\caption[]%
		{{ 
		Quiet (0.350)
		}}    
		\label{fig:svs_2_example}
	\end{subfigure}
        \begin{subfigure}[b]{0.245\textwidth}  
		\centering 
		\includegraphics[width=\textwidth]{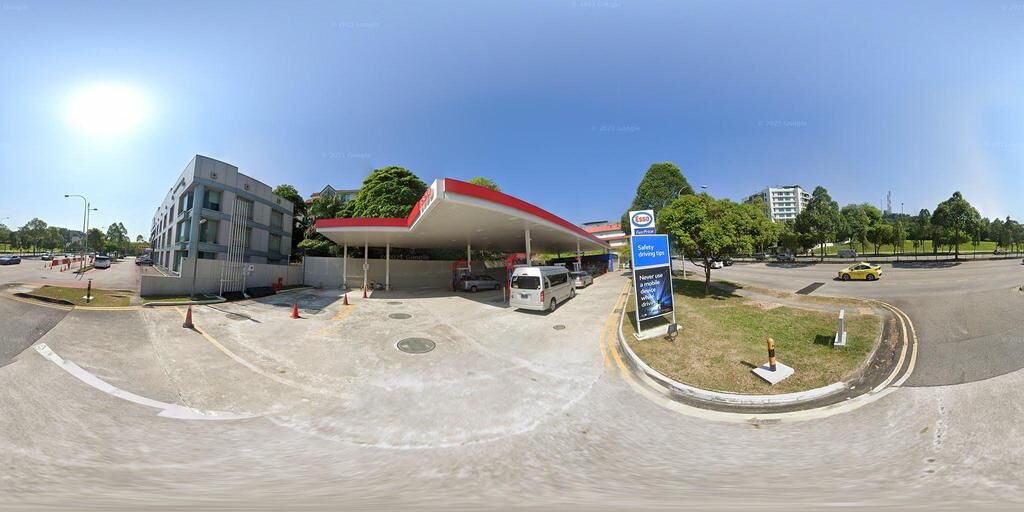}\vspace*{-0.2cm}
		\caption[]%
		{{ 
		Noisy (0.600)
		}}    
		\label{fig:svs_3_example}
	\end{subfigure}
        \begin{subfigure}[b]{0.245\textwidth}  
		\centering 
		\includegraphics[width=\textwidth]{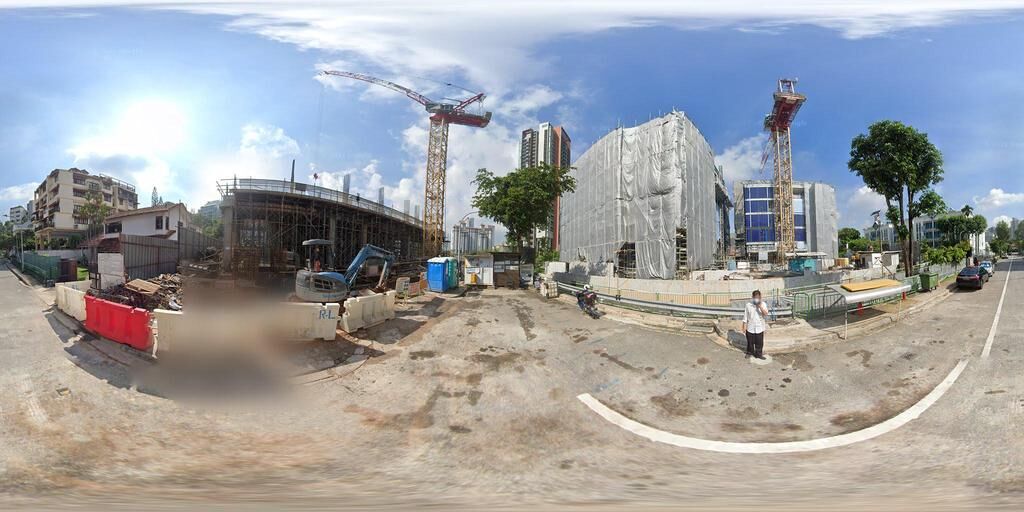}\vspace*{-0.2cm}
		\caption[]%
		{{ 
		Very Noisy (0.965)
		}}    
		\label{fig:svs_4_example}
	\end{subfigure}
	\caption{Some street view image examples in $\svs$ dataset. The image caption indicates the noise intensity class this image belongs to and the numbers in parenthesis indicate the original noise intensity scores from Zhao et al. \cite{zhao2023sound}. 
 }
	\label{fig:svs_example}
    \vspace*{-0.15cm}
\end{figure*}

In this work, we use a recently developed street view image noise intensity dataset developed by Zhao et al. \cite{zhao2023sound} as a representative urban perception task. This dataset consists of 579 street-view images collected from 
Singapore. The noise intensity score (between 0 and 1) were collected based on a human survey. Please refer to their Github\footnote{\url{https://github.com/ualsg/Visual-soundscapes}} for a detailed description of this dataset. Since the sound intensity score is not a commonly agreed metric but an indicator defined by Zhao et al. \cite{zhao2023sound},  it would be challenging for visual foundation models trained on general web data such as OpenCLIP \cite{gabriel2021openclip} and BLIP \cite{li2022blip} to directly predict such a score. Therefore, we discretize the original noise intensity score of each street view image into four classes: very quiet (0 - 0.25), quiet (0.25 - 0.50), noisy (0.50 - 0.75), and very noisy (0.75 - 1.00). We denote this dataset as $\svs$. Figure \ref{fig:svs_example} illustrates some street view image examples from each noise intensity class. We randomly split $\svs$ into 50\% training and 50\% testing sets, where the testing dataset is used to evaluate different CNN and foundation models.



Since all GPT models (except GPT-4) used in previous experiments are pure language models that cannot handle data modalities such as images. So for the street view image-based noise intensity prediction task, we select the latest high-performance open visual-language foundation models (VLFM) including \openclip~ \cite{gabriel2021openclip}, \blip~ \cite{li2022blip}, and \flamingo~ \cite{awadalla2023OpenFlamingo}. Although, there exist more powerful visual-language foundation models such as DeepMind's Flamingo-9B \cite{Alayrac2022Flamingodeepmind}, KOSMOS-1 \cite{huang2023kosmos1}, and GPT-4 \cite{openai2023gpt4}, they are not openly accessible, nor do they provide API access yet\footnote{Note that the GPT-4 API still does not support visual question answering at the time we submit this paper.}.
We describe the setting of each VLFM as follows: 
\begin{itemize}
    \item \textbf{\openclipl}: We use an \openclip~ \cite{gabriel2021openclip} ViT L/14 model pre-trained with the LAION-2B English subset of LAION-5B\footnote{\url{https://laion.ai/blog/laion-5b/}} as a small-sized \openclip~ model. We download the pre-trained model from Huggingface\footnote{\url{https://huggingface.co/laion/CLIP-ViT-L-14-laion2B-s32B-b82K}}.
    \item \textbf{\openclipb}: We use the \openclip~ \cite{gabriel2021openclip} ViT-bigG/14 model trained with the LAION-2B English subset of LAION-5B as a larger-sized \openclip~ model. The pre-trained model is from Huggingface\footnote{\url{https://huggingface.co/laion/CLIP-ViT-bigG-14-laion2B-39B-b160k}}.
    \item \textbf{\blip}: We use the pre-trained \blip-2 model \cite{li2023blip} provided by Huggingface\footnote{\url{https://huggingface.co/Salesforce/blip2-flan-t5-xl}} which consists of a CLIP-like image encoder, a Querying Transformer (Q-Former), and a large language model (Flan T5-xl). 
    \item \textbf{\flamingo}: We use the pre-trained OpenFlamingo-9B model \cite{awadalla2023OpenFlamingo} provided by Huggingface\footnote{\url{https://huggingface.co/openflamingo/OpenFlamingo-9B}} which consists of an image encoder (CLIP ViT-L/14 \cite{gabriel2021openclip}) and a large language model (LLaMA-7B \cite{touvron2023llama}).
\end{itemize}

All VLFMs are evaluated on the testing set of $\svs$ in a zero-shot setting. Since different VLFMs require different image input formats and expect different styles of text prompts, we describe the zero-shot pipeline for each VLFM below:
\begin{itemize}
    \item \textbf{\openclipl} and \textbf{\openclipb}: We first encode four noise intensity class names into four text embeddings by using a text template of the form ``\textit{a city area with the noise intensity of [NOISE\_INTENSITY\_CLASS]}''. 
    Then given a street view image, we use \openclip~ ViT image encoder to encode them into an image embedding. The cosine similarity between this image embedding and all four class text embeddings are computed and the class with the highest similarity will be picked as the prediction. 
    \item \textbf{\blip}: Given a street view image, we use a prompt of the form ``\textit{What is the noise intensity of this area, is it 1. very quiet, 2. quiet, 3. noisy, or 4. very noisy?}'' to instruct the language encoder of \blip~ to predict its noise intensity class. 
    \item \textbf{\flamingo}: We use a prompt of the form ``\textit{There are four noise intensity levels: 1. very quiet, 2. quiet, 3. noisy, or 4. very noisy. <image>The noise intensity of this area is}'' to instruct \flamingo~ to predict the noise intensity of the given image. Here ``<image>'' denotes an image token and CLIP ViT-L/14 is used as the encoder.
\end{itemize}

We select four convolutional neural network models (CNNs) as the alternative baselines to compare against these VLFMs: \alexnet~ \cite{krizhevsky2017alexnet}, \rsnet18 \cite{he2016resnet}, \rsnet50 \cite{he2016resnet}, and \dsnet161 \cite{huang2017densenet}.
The weights of all CNNs models are first initialized by the Place365 pre-trained weights \cite{zhou2017places365}, and only their final softmax layers are finetuned with full supervision on the $\svs$ training dataset. We choose this linear probing method instead of fully finetuning the whole CNN architecture due to the very limited training data size.

\begin{table}[t!]
\caption{
Evaluation results of various vision-language foundation models and baselines on the urban street view image-based noise intensity classification dataset, \svs~ \cite{zhao2023sound}. 
We classify models into two groups:
(A) Supervised finetuned convolutional neural networks (CNNs); 
(B) Zero-shot learning with visual-language foundation models (VLFMs).
We use accuracy and weighted F1 scores as evaluation metrics. The best scores for each group are highlighted.
}
\label{tab:exp_streetview}
\centering
{ 
\begin{tabular}{l|l|c|cc}
\toprule
                                     & Model           & \#Param & Accuracy       & F1             \\ \hline
\multirow{4}{*}{(A) Supervised Finetuned CNNs} & \alexnet~ \cite{krizhevsky2017alexnet}        & 58M     & 0.452          & 0.405          \\ 
                                     & \rsnet18 \cite{he2016resnet}        & 11M     & 0.493          & \textbf{0.442} \\
                                     & \rsnet50 \cite{he2016resnet}       & 24M     & \textbf{0.500} & 0.436          \\
                                     & \dsnet161 \cite{huang2017densenet}     & 27M     & 0.486          & 0.382          \\ \hline
\multirow{4}{*}{(B) Zero-shot FMs}           & \openclipl~ \cite{radford2021learning,gabriel2021openclip,schuhmann2022laionb}     &    427M   & 0.128          & 0.089          \\
                                     & \openclipb~ \cite{radford2021learning,gabriel2021openclip,schuhmann2022laionb}  &     2.5B    & 0.169          & 0.178          \\
                                     & \blip~ \cite{li2022blip,li2023blip}           &    3.9B  & \textbf{0.452} & \textbf{0.405} \\
                                     & \flamingo~ \cite{awadalla2023OpenFlamingo} & 8.3B      & 0.262          & 0.127      \\ \bottomrule   
\end{tabular}
}
\end{table}

\begin{figure*}
	\centering \tiny
	\vspace*{-0.2cm}
	\begin{subfigure}[b]{0.245\textwidth}  
		\centering 
		\includegraphics[width=\textwidth]{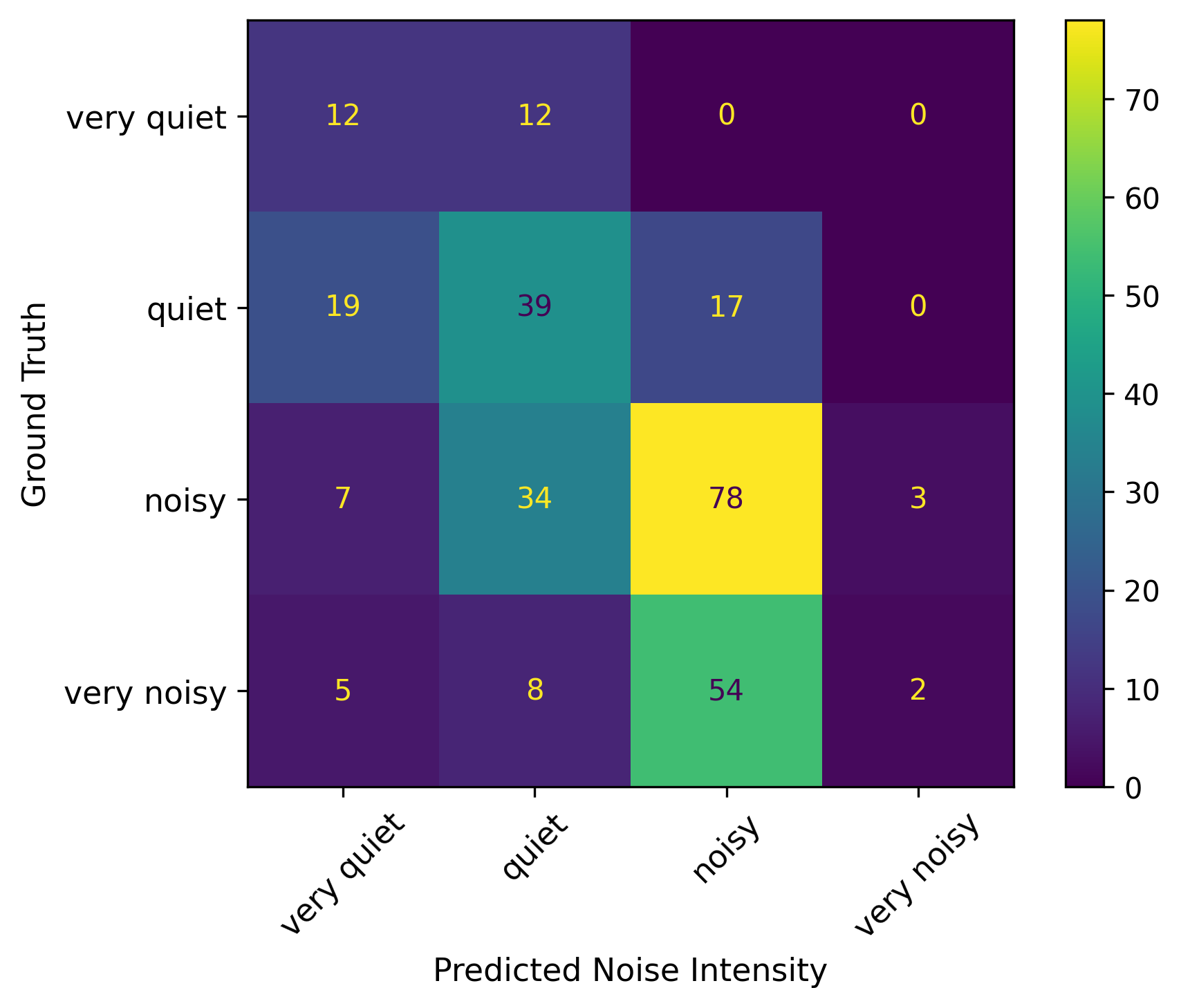}\vspace*{-0.2cm}
		\caption[]%
		{{ 
		\alexnet
		}}    
		\label{fig:svs_alexnet_cm}
	\end{subfigure}
        \begin{subfigure}[b]{0.245\textwidth}  
		\centering 
		\includegraphics[width=\textwidth]{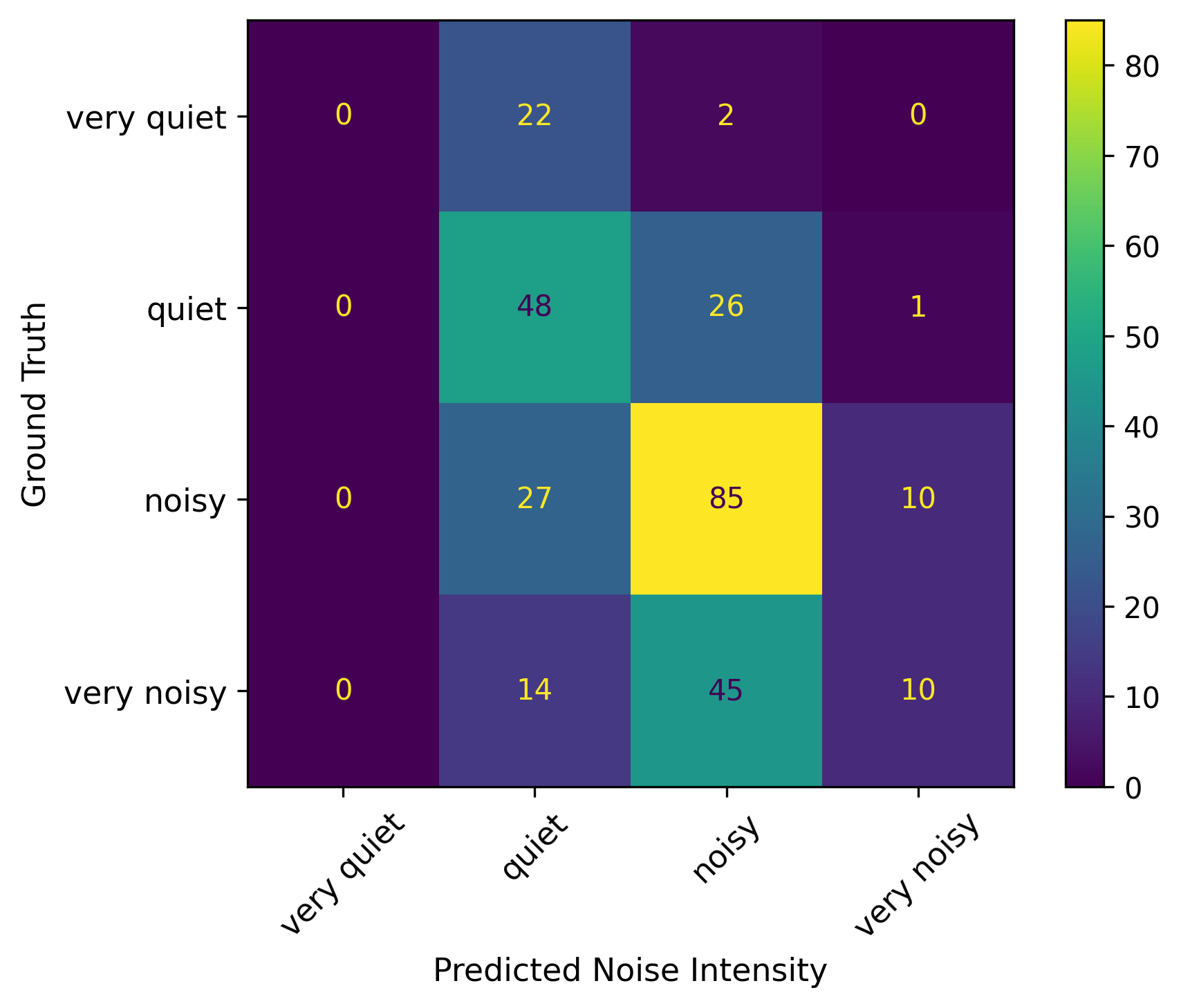}\vspace*{-0.2cm}
		\caption[]%
		{{ 
		\rsnet18
		}}    
		\label{fig:svs_resnet18_cm}
	\end{subfigure}
        \begin{subfigure}[b]{0.245\textwidth}  
		\centering 
		\includegraphics[width=\textwidth]{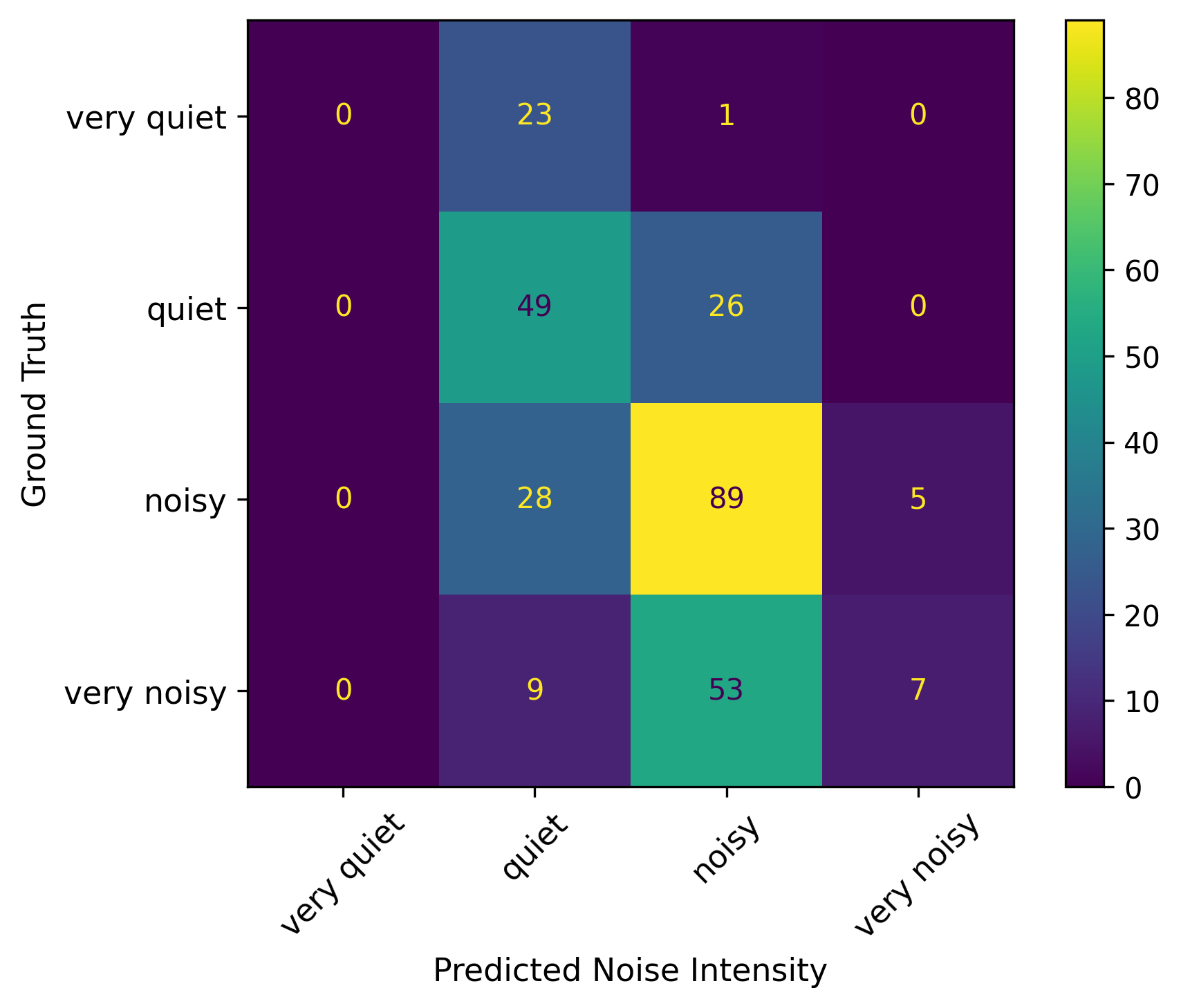}\vspace*{-0.2cm}
		\caption[]%
		{{ 
		\rsnet50
		}}    
		\label{fig:svs_resnet50_cm}
	\end{subfigure}
        \begin{subfigure}[b]{0.245\textwidth}  
		\centering 
		\includegraphics[width=\textwidth]{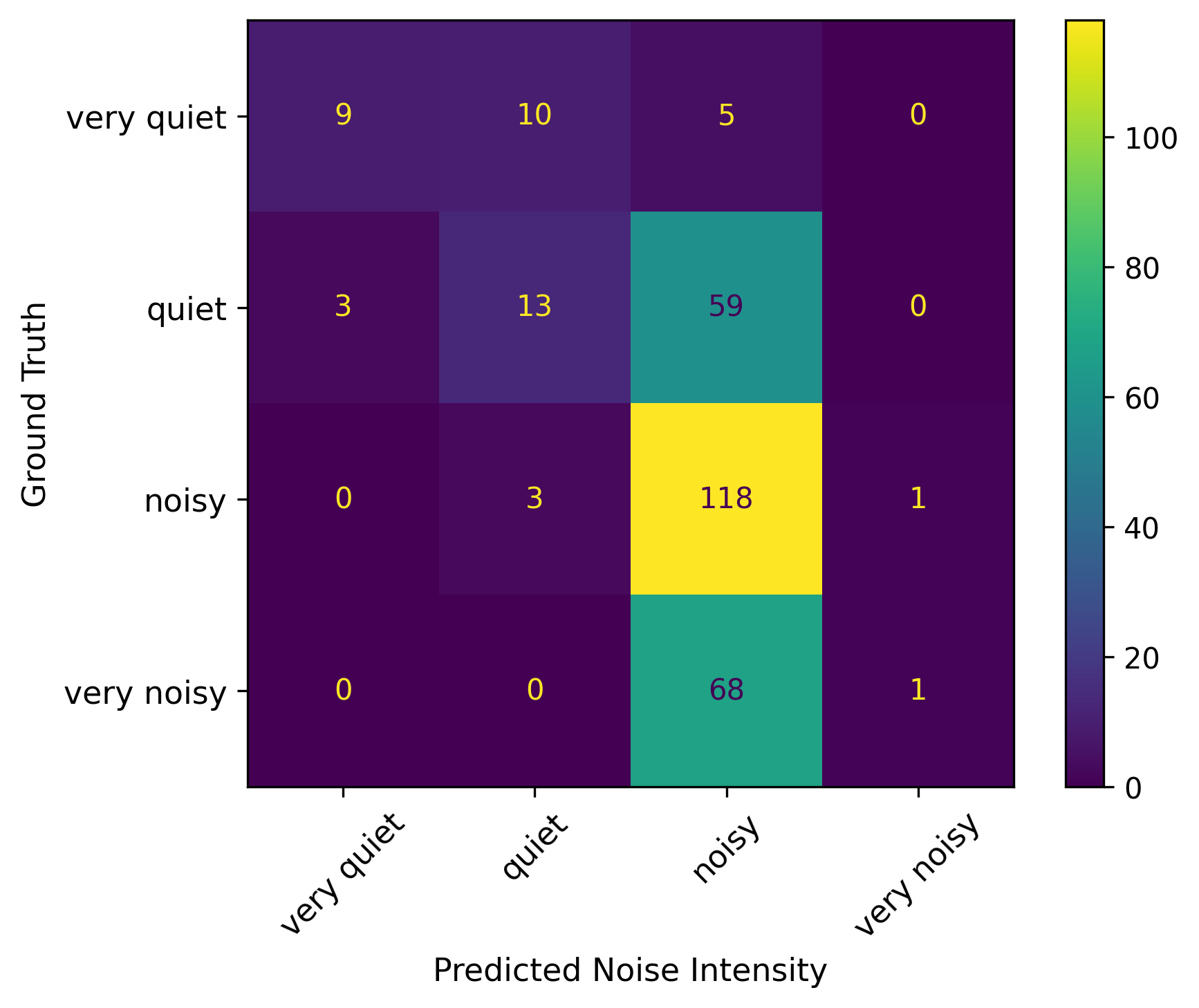}\vspace*{-0.2cm}
		\caption[]%
		{{ 
		\dsnet161
		}}    
		\label{fig:svs_densenet161_cm}
	\end{subfigure}
        \begin{subfigure}[b]{0.245\textwidth}  
		\centering 
		\includegraphics[width=\textwidth]{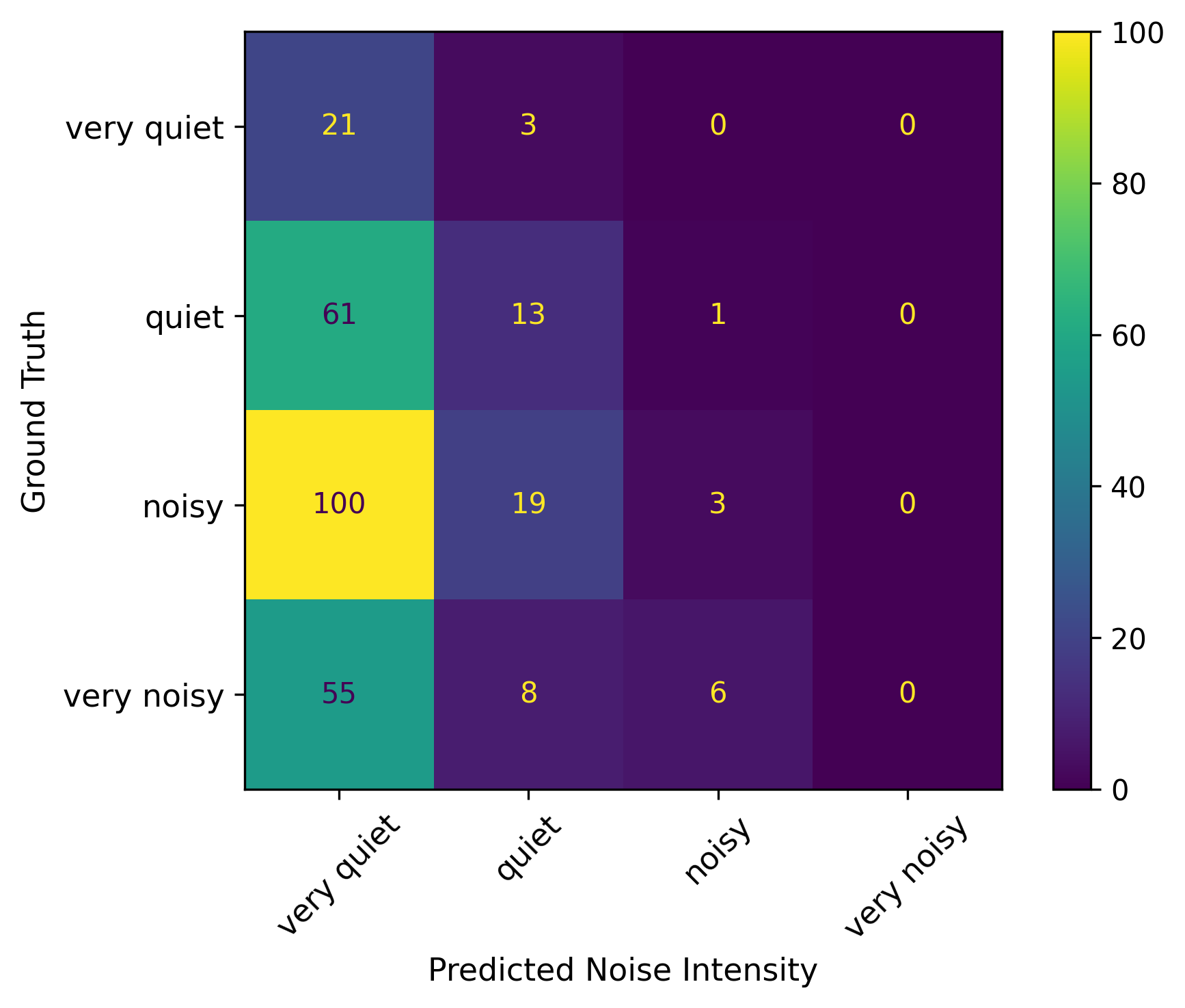}\vspace*{-0.2cm}
		\caption[]%
		{{ 
		\openclipl
		}}    
		\label{fig:svs_openclip_l_cm}
	\end{subfigure}
        \begin{subfigure}[b]{0.245\textwidth}  
		\centering 
		\includegraphics[width=\textwidth]{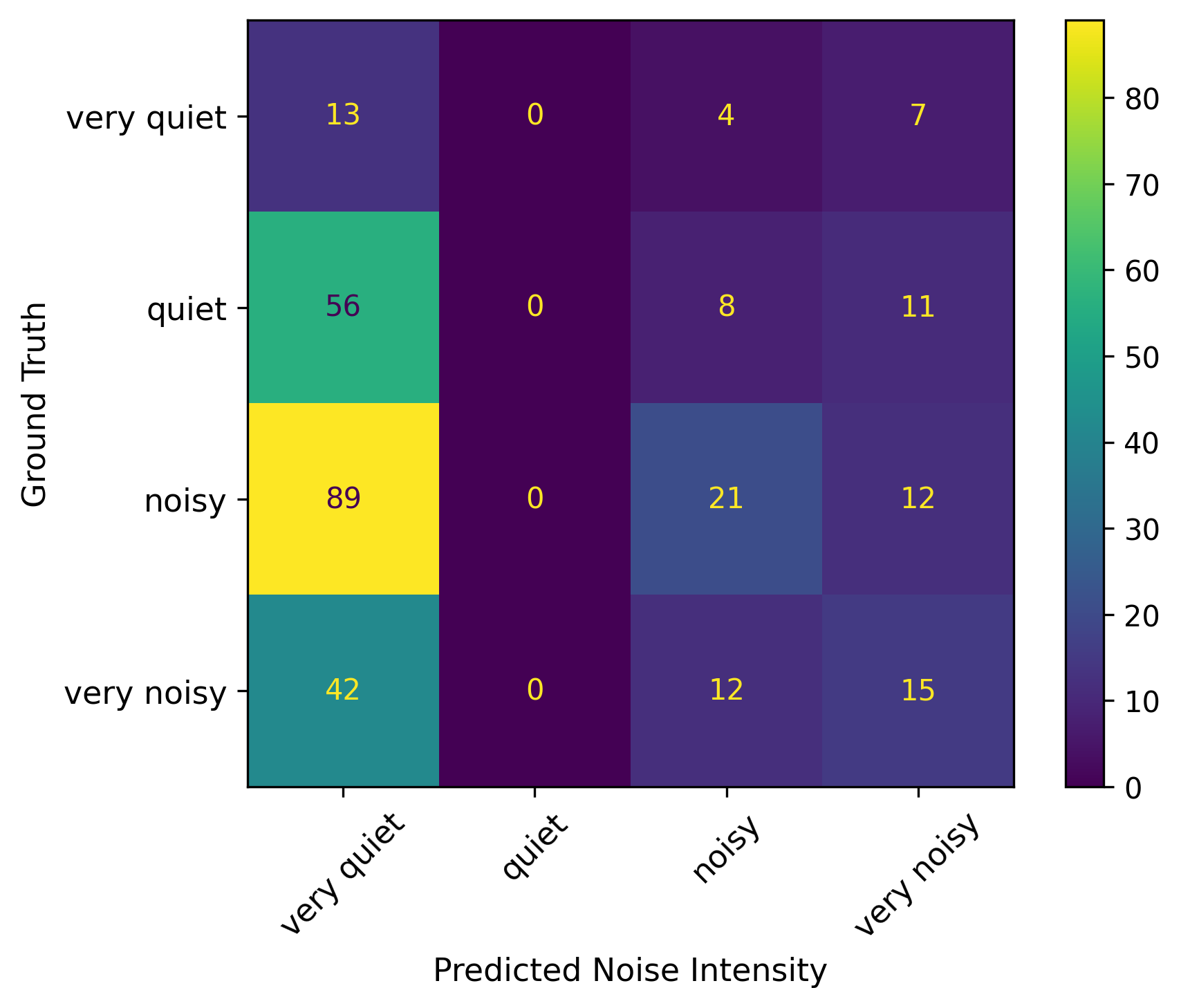}\vspace*{-0.2cm}
		\caption[]%
		{{ 
		openclipb
		}}    
		\label{fig:svs_openclip_b_cm}
	\end{subfigure}
        \begin{subfigure}[b]{0.245\textwidth}  
		\centering 
		\includegraphics[width=\textwidth]{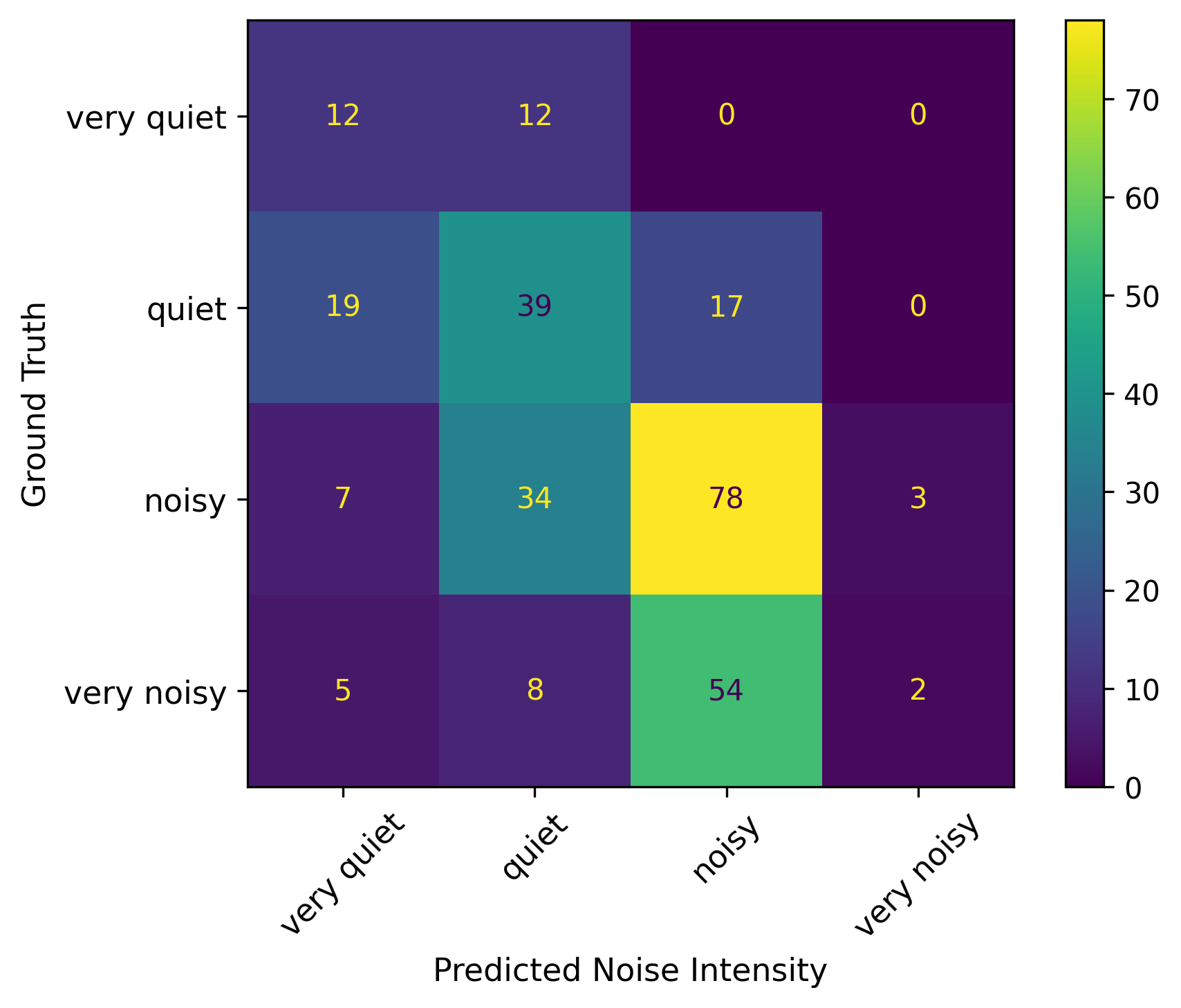}\vspace*{-0.2cm}
		\caption[]%
		{{ 
		\blip
		}}    
		\label{fig:svs_blip_cm}
	\end{subfigure}
        \begin{subfigure}[b]{0.245\textwidth}  
		\centering 
		\includegraphics[width=\textwidth]{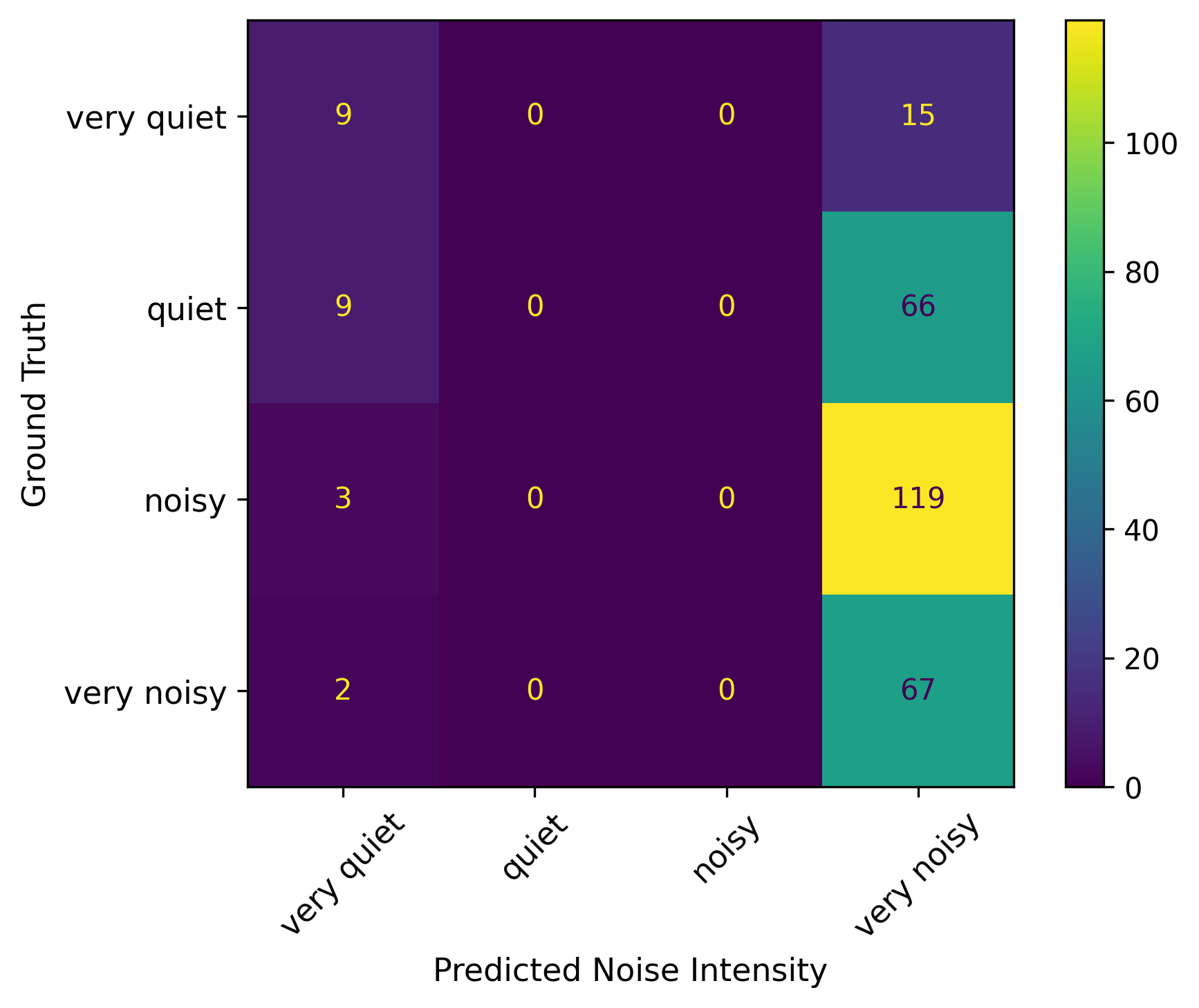}\vspace*{-0.2cm}
		\caption[]%
		{{ 
		\flamingo
		}}    
		\label{fig:svs_openflamingo_cm}
	\end{subfigure}
	\caption{Confusion matrices of all baselines and visual-language FMs on $\svs$ dataset. 
 }
	\label{fig:svs_all_cm}
    \vspace*{-0.15cm}
\end{figure*}

Table \ref{tab:exp_streetview} compares the performances of different finetuned CNN models with four zero-shot VLFMs. The results show that \blip~ achieves the best accuracy and weighted F1 score among all VLFMs in the zero-shot learning setting.  The performance of \blip~ is comparable to those of \alexnet~ but is still slightly worse than the best model, \rsnet18 and \rsnet50. To further understand the classification accuracy of different models on each noise intensity class, we visualize the confusion matrices of all models in Figure \ref{fig:svs_all_cm}. We can see that the predictions of \openclipl, \openclipb, and \flamingo~ are highly biased. \openclipl~ and \openclipb~ tend to classify most street view images as `very quiet' while \flamingo~ classifies most images as `very noisy'. On the other hand, only \blip~ shows balanced and reasonable predictions on all four noise intensity classes, similar to those fine-tuned CNN models. 

These results are very encouraging with zero-shot BLIP achieving comparable performance with fine-tuned models. We can observe from Figure \ref{fig:svs_blip_cm} that BLIP has a general sense of the noisy intensity level of the target urban area, e.g., it mis-classifies most ``very noisy'' areas as simply ``noisy''. This implies that \blip~ understands noisy intensity levels on a different scale. For example, a ``very noisy'' place annotated by a human interviewee in Singapore might not qualify as ``very'' for BLIP, which might have seen many much noisier urban areas. To this end, \blip is generally competent for this urban perception task.
At the same time, we recognize that most of the open visual-language foundation models are still not powerful enough to connect visual features to their important yet nuanced semantics and concepts in urban studies. For example, when presented with a construction site in Figure \ref{fig:svs_4_example}, we expect a VLFM to predict that this is a very noisy neighborhood. When seeing a large vegetation coverage in Figure \ref{fig:svs_4_example}, a VLFM should associate this visual feature with the concept of `quiet' in the language space. This study highlights the fact that the current VLFMs have certain capabilities in understanding the characteristics of urban neighborhoods given visual inputs. However, their ability is still generally not as strong as the current {\llm}s on language-only tasks. Furthermore, we think the urban perception task, as a classic task in urban geography, is more challenging 
than current visual question-answering tasks commonly used in VLFM research \cite{radford2021clip,huang2023kosmos1} partly due to their partially subjective nature and the rarity of annotated datasets. 
This further emphasizes the unique challenges faced by foundation model research in GeoAI.

\subsection{Remote Sensing}  \label{sec:exp_rs}

Our final experiment focuses on a typical remote sensing (RS) task -- remote sensing image scene classification. We choose a widely-used aerial image scene classification dataset, $\aid$ \cite{xia2017aid}, which consists of 10K scenes and 30 aerial scene types. These data were collected from Google Earth imagery. Please refer to Xia et al. \cite{xia2017aid} for a detailed description of this dataset. $\aid$ does not provide an official dataset split, so 
we split the dataset into training and testing sets using stratified sampling with a ratio of 80\% for training and 20\% for testing, ensuring that both sets have similar scene type label distributions.

Similar to the street view image classification task from Section \ref{sec:exp_urban_func}, we use four CNN models (i.e., \alexnet, \rsnet18, \rsnet50, and \dsnet161) and four visual-language foundation models (i.e., \openclipl, \openclipb, \blip, and \flamingo). 
For all CNNs models, their weights are first initialized by the ImageNet-V1 pre-trained weights, and their final softmax layers are fine-tuned with full supervision on the $\aid$ training dataset. 
For the VLFMs, their model performances are highly dependent on whether their language model component can correctly comprehend the semantics of each RS image scene type. However, many RS image scene types of $\aid$ are vague such as ``center'', and ``commercial''. We find that if keeping their original scene type names, models like \openclip~ would assign no RS image to those two types. 
Therefore, we modify the names of ``center'' to ``theater'' (although only partially covers the semantics of this class), and ``commercial'' to ``commercial area'' and use them in the prompt. Models with such prompts are denoted as ``$\chg$'' while ``$\ori$'' denotes the original RS image scene type names from $\aid$ being used in the prompt. We evaluate all VLFMs in a zero-shot learning setting. Following the street view image classification task in Section \ref{sec:exp_urban_func}, similar prompt formats are used on the $\aid$ dataset. 

\begin{table}[t!]
\caption{
Evaluation results of various vision-language foundation models and baselines on the remote sensing image scene classification dataset, $\aid$ \cite{xia2017aid}. 
We use the same model set as Table \ref{tab:exp_streetview}. 
``$\ori$'' denotes we use the original remote sensing image scene class name from $\aid$ to populate the prompt while ``$\chg$''indicates we update some class names to improve its semantic interpretation for FMs.
We use accuracy and F1 score as evaluation metrics.
}
\label{tab:exp_rs}
\centering
{ 
\begin{tabular}{l|l|c|cc}
\toprule
                                     & Model             & \#Param & Accuracy       & F1             \\ \hline
\multirow{4}{*}{Supervised Finetuned CNNs} & \alexnet~ \cite{krizhevsky2017alexnet}           & 58M     & \textbf{0.831} & \textbf{0.827} \\
                                     & \rsnet18 \cite{he2016resnet}          & 11M     & 0.752          & 0.730          \\
                                     & ResNet50 \cite{he2016resnet}          & 24M     & 0.757          & 0.738          \\
                                     & \dsnet161 \cite{huang2017densenet}       & 27M     & 0.818          & 0.807          \\ \hline
\multirow{7}{*}{Zero-shot FMs}           & \openclipl~ $\ori$ \cite{radford2021learning,gabriel2021openclip,schuhmann2022laionb}    & 427M    & 0.708          & 0.688          \\
                                     & \openclipl~ $\chg$ \cite{radford2021learning,gabriel2021openclip,schuhmann2022laionb}    & 427M    & \textbf{0.710} & \textbf{0.698} \\
                                     & \openclipb~ $\ori$ \cite{radford2021learning,gabriel2021openclip,schuhmann2022laionb} & 2.5B    & 0.699          & 0.668          \\
                                     & \openclipb~ $\chg$ \cite{radford2021learning,gabriel2021openclip,schuhmann2022laionb} & 2.5B    & 0.705          & 0.686          \\
                                     & \blip~ $\ori$ \cite{li2022blip}           & 2.5B    & 0.500          & 0.473          \\
                                     & \blip~ $\chg$ \cite{li2022blip}           & 2.5B    & 0.520          & 0.494          \\
                                     & \flamingo~ \cite{awadalla2023OpenFlamingo}   & 8.3B    & 0.206          & 0.154        \\ \bottomrule 
\end{tabular}
}
\end{table}

Table \ref{tab:exp_rs} summarizes the experiment results of four fine-tuned CNNs models and zero-shot VLFMs. We can see that  \alexnet~ achieves the best accuracy and F1 score among all CNN models. 
Surprisingly, \openclipl~ $\chg$ obtains the best accuracy and F1 score among all VLFMs. We observe that bigger models do not necessarily lead to better results in this task. For example, the largest model, \flamingo~ only achieves a 0.206 accuracy. One possible reason is that these larger VLFMs might not see remote-sensing images in their training data, which usually contain general web-crawled images and texts. \openclip~, on the other hand, explicitly includes satellite images in their pre-training data\cite{gabriel2021openclip}. However, both \blip~ and \flamingo~ did not mention whether they utilized remote sensing images during the pre-training stage. Note that street view images are quite similar to Internet images which are widely used for VLFM pre-training. RS images, on the other hand, such as satellite images and UAV (unmanned aerial vehicles), are visually distinguished from Internet photos where the majority of them are captured using consumers' digital cameras at the ground level. If the visual encoders of \blip~ and \flamingo~ are not pre-trained on RS images, the features they extracted will not align well with text features that share similar semantics--this leads to poor performance on the $\aid$ dataset. Our study highlights the importance of pre-training VLFMs on a diverse set of visual inputs, including RS images, to improve their performance on remote sensing tasks.

Another important observation 
is that the semantics embedded in the prompts play a pivotal role in determining the model's performance. 
For example, when using the original scene type name ``center'', generally none of the models is able to understand the underlying ambiguous meaning. However, simply changing ``center'' to ``theater'' could help \openclip~ correctly find relevant RS scenes, although this is not a perfect name to describe this class. Nevertheless, this simple change demonstrates the importance of choosing expressive prompts while using FMs for geospatial tasks.

Compared with the results in Table \ref{tab:exp_urbanpoi_eval}, the experimental results in Table \ref{tab:exp_rs} highlight the unique challenges of remote sensing images. We will discuss the improvement of {\lpm}s~ for remote sensing in detail in Section \ref{sec:rs}.


\section{A Multimodal Foundation Model for GeoAI} \label{sec:multimodal}

Section \ref{sec:exp} explores the effectiveness of applying existing {\lpm}s on different tasks from various geospatial domains. We can see that many large language models can outperform fully-supervised task-specific ML/DL models and achieve surprisingly good performances
on several geospatial tasks such as toponym recognition, location description recognition, and 
time series forecasting of dementia. However, on other geospatial tasks (i.e., the two tested Urban Geography tasks and one RS task), especially those that involve multiple data modalities (e.g., point data, street view images, RS images, etc.), existing foundation models still underperform task-specific models.
In fact, one unique characteristic of many geospatial tasks is that they involve many data modalities such as text data, knowledge graphs, remote sensing images, street view images, trajectories, and other geospatial vector data. This will put a significant challenge on GeoAI foundation model development. 
So in this section, we discuss the challenges unique to each data modality, then propose a potential framework for future GeoAI which leverages a multimodal {\lpm}. 


\vgap
\subsection{Geo-Text Data} \label{sec:geo-text}

\begin{figure}[ht!]
	\centering 
	\vspace*{-0.2cm}
	\begin{subfigure}[b]{0.49\textwidth}  
		\centering 
		\includegraphics[width=\textwidth]{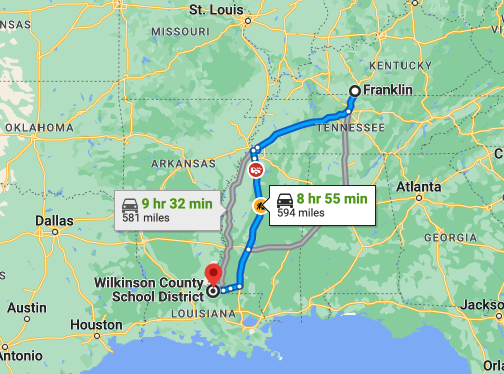}\vspace{-0.2cm}
		\caption[]%
		{{ 
		[TEXT]: \underline{Franklin} is a city in and the county seat of simpson county, ...
		}}    
		\label{fig:geosparsing_3}
	\end{subfigure}
 	\hfill
	\begin{subfigure}[b]{0.49\textwidth}  
		\centering 
		\includegraphics[width=\textwidth]{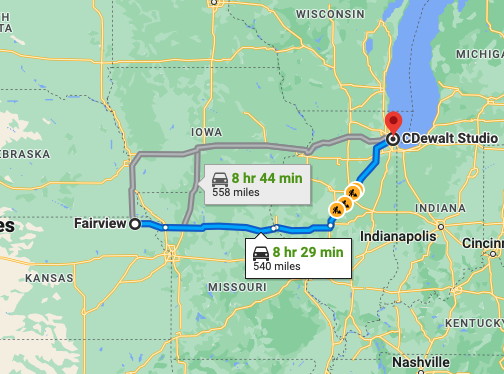}\vspace{-0.2cm}
		\caption[]%
		{{
		[TEXT]: the city of \underline{Fairview} had a population of 260 as of july 1, 2015. ...
		}}    
		\label{fig:geosparsing_4}
	\end{subfigure}
	\vspace{-0.4cm}
	\caption{  Geoparsing examples of \gptthree~ on 
	the Ju2016 dataset
	comparing the predicted coordinates (dropped pins) and the ground truth coordinates (starting points).
	The recognized toponyms are underlined in text. 
	} 
	\vspace{-0.5cm}
	\label{fig:geosparsing}
\end{figure}

Despite the promising results showed in Table \ref{tab:ner}, {\llm}s still struggle with
more complex geospatial semantics tasks such as toponym resolution/geoparsing \cite{alex2015edinburgh,gritta2018camcoder,wang2019EUPEG} 
 and geographic question answering (GeoQA) \cite{mai2020se,mai2021geoqa}, 
since {\llm}s
are unable to perform (implicit) spatial reasoning in a way that is grounded in the real world.
As a concrete example, we illustrate the shortcomings of \gptthree~ on a geoparsing task. 
Using two examples from the Ju2016 dataset, we ask \gptthree~ to both: 
1) recognize toponyms; and 2) predict their geo-coordinates. 
The prompt is shown in List \ref{ls:prompt-geoparse} while the geoparsing results are visualized in Figure \ref{fig:geosparsing}. 
We see that in both cases, \gptthree~ can correctly recognize the toponyms but the predicted coordinates are 500+ miles away from the ground truth.
Moreover, we notice that 
with a small spatial displacement of the generated geo-coordinates, \gptthree's log probability for this new pair of coordinates decreases significantly. 
In other words, the probability of coordinates generated by the {\llm} does not follow Tobler's First Law of Geography \cite{tobler1970computer}. 
\gptthree~ also generates invalid latitudinal/longitudinal coordinates,
indicating that existing {\llm}s are still far from gracefully handling complex numerical and spatial reasoning tasks.

\begin{figure}[b]
\begin{minipage}[c]{0.48\textwidth}
	\begin{lstlisting}[
	style=prompt-style, 
	basicstyle=\ttfamily\tiny,
	linewidth=\textwidth,
	breaklines=true,
	captionpos=b, 
	caption={Geoparsing with {\llm}s, e.g., GPT-3. Yellow block: the text snippet to be geoparsed. Orange box: GPT-3 outputs.
	},
	label={ls:prompt-geoparse},
	frame=tb
	]
%*\colorbox{pinkannoback}{[Instruction]}*)...
%*\colorbox{blueannoback}{Paragraph:}*) San Jose was founded in 1803 when allotments of land were made ...
%*\colorbox{greenannoback}{Q:}*) Which words in this paragraph represent named places?
%*\colorbox{redannoback}{A:}*) San Jose; New Mexico

%*\colorbox{greenannoback}{Q:}*) What is the location of San Jose?
%*\colorbox{redannoback}{A:}*) 35.39728, -105.47501
...
--
Paragraph: %*\colorbox{yellowannoback}{the city of fairview had a population of 260 as of july 1, 2015.  ...}*) 
%*\colorbox{greenannoback}{Q:}*) Which words in this paragraph represent named places?
%*\colorbox{redannoback}{A:}*) %*\colorbox{orangeannoback}{Fairview}*)

%*\colorbox{orangeannoback}{Q: What is the location of Fairview?}*)
%*\colorbox{orangeannoback}{A: 41.85003, -87.65005}*)
	\end{lstlisting}
	\vspace{-0.2cm}
\end{minipage}
\begin{minipage}[c]{0.48\textwidth}
	\centering 
	\includegraphics[width=\textwidth]{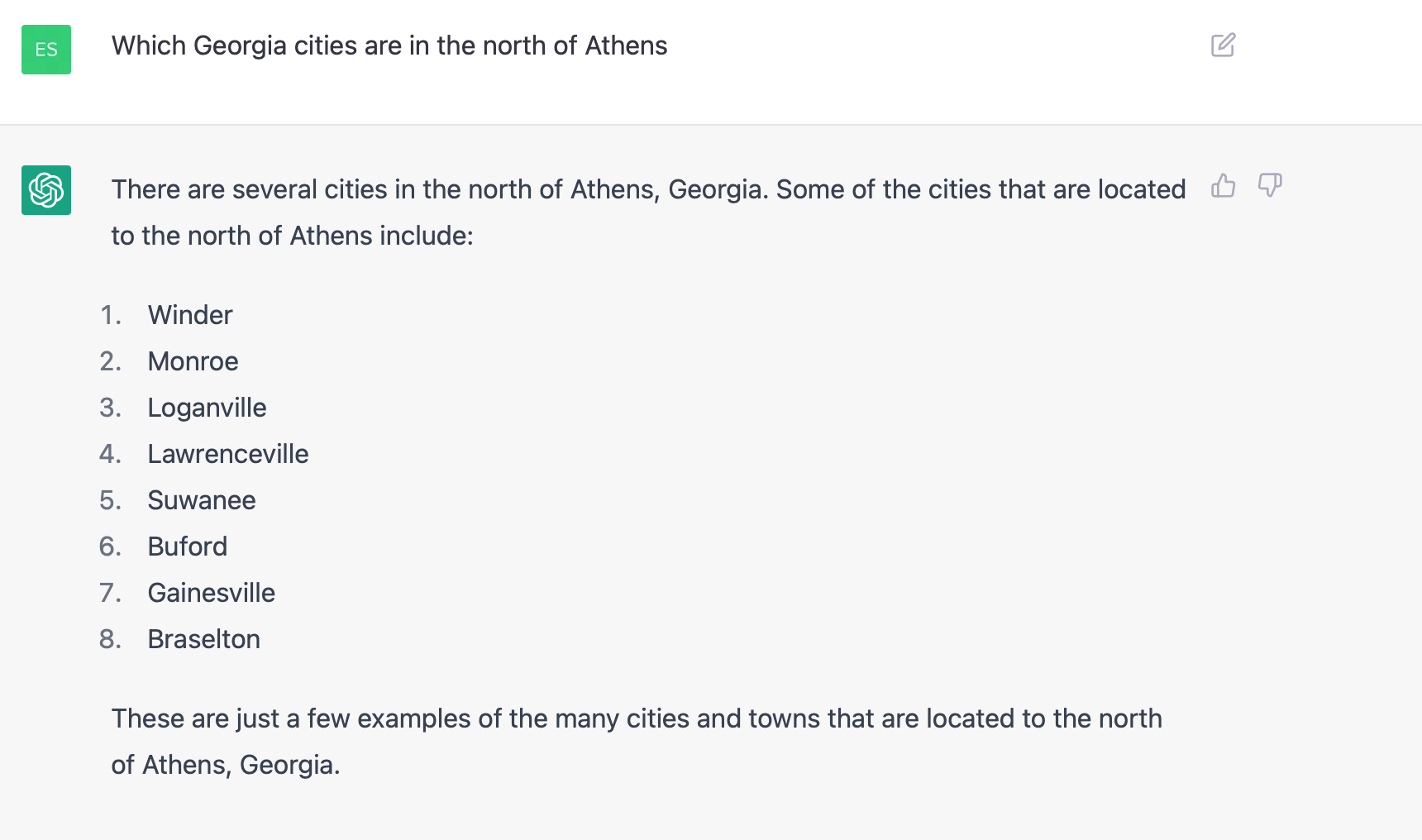}
	\caption{One example in which \chatgpt~ gives wrong answers to a geographic question about topological relations. In this example, Monore is not in the north, but the southwest of Athens, GA.
	} 
	\label{fig:chatgpt_error_topo}
\end{minipage}
\end{figure}

Figure \ref{fig:chatgpt_error_topo} provides another example of unsatisfactory results of {\llm}s in answering geographic questions related to spatial relations. 
In this example, Monore, in the generated answer by the \chatgpt~generated answer is not in the north of Athens, GA, but in the southwest of Athens. This example indicates that {\llm}s do not fully understand the semantics of spatial relation. 
The reason for this error could be that 
\chatgpt~ generates answers to this spatial relation question based on searching through its internal memory of text-based knowledge rather than performing spatial reasoning. 
One potential solution to this problem could be the use of geospatial knowledge graphs\cite{zhu2022reasoning,cai2022hyperquaternione}, which can guide the {\llm}s to perform explicit spatial relation computations. We will discuss this further in the next section.

\subsection{Geospatial Knowledge Graph} \label{sec:geokg}

Despite the superior end-to-end prediction and generation capability, {\llm}s may produce content that lacks sufficient coverage of factual knowledge or even contains non-factual information.
To address this problem, knowledge graphs (KGs) can serve as effective sources of information that complement {\llm}s. 
KGs are factual in nature because the information is usually extracted from reliable sources, with post-processing conducted by human editors to further ensure incorrect content is removed. 
As an important type of domain knowledge graphs, geospatial knowledge graphs (GeoKG) such as \textit{GeoNames} \citep{ahlers2013geonames}, \textit{ LinkedGeoData} \citep{auer2009linkedgeodata}, \textit{YAGO2} \citep{hoffart2013yago2}, \textit{GNIS-LD} \citep{regalia2018gnis}, \textit{KnowWhereGraph} \citep{janowicz2022know}, \textit{EVKG} \citep{qi2023evkg}, etc. are usually generated from authoritative data sources and spatial databases. For example, GNIS-LD was constructed based on USGS's Geographic Names Information System (GNIS). This ensures the authenticity of these geospatial data. 


In particular, developing multimodal {\lpm}s for GeoAI which jointly consider text data and geospatial knowledge graphs can lead to several advantages. 
First, from the model perspective, (geospatial) knowledge graphs could be integrated into pre-training or fine-tuning {\llm}s, through strategies such as retrieving embeddings of knowledge entities for contextual representation learning~\cite{peters2019knowledge}, fusing knowledge entities and text information~\cite{he2021klmo,zhang2022greaselm}, designing learning objectives that focus on reconstructing knowledge entities~\cite{zhang2019ernie} and triples \cite{yasunaga2022DRAGON}.
Second, from the data perspective, GeoKGs could provide contextualized semantic and spatiotemporal knowledge to facilitate prompt engineering or data generation, such as enriching prompts with contextual information from KGs~ \cite{brate2022improving,wu2023survey} and converting KG triples into natural text corpora for specific domains~\cite{agarwal2021knowledge}.
Third, from the application perspective, it is possible to convert facts in geospatial knowledge graphs into natural language to enhance text generation~\cite{yu2022survey}, to be used in scenarios such as (geographic) question answering~\cite{fan2019using,mai2019contextual} and dialogue systems~\cite{wu2019global}. 
Last, from a reasoning perspective, GeoKGs usually provide spatial footprints of geographic entities which  enable {\llm}s to perform explicit spatial reasoning
as Neural Symbolic Machine did \cite{liang2017nsm}. This can help avoid the errors we see in Figure \ref{fig:chatgpt_error_topo}.


\subsection{Street View Image} \label{sec:dis_streetview}

Section \ref{sec:exp_urban_func} has demonstrated the effectiveness of existing visual-language foundation models on a street view-based geospatial task. However, the performance gaps between the task-specific models and VLFMs shown in Table \ref{tab:exp_streetview} inform us that there are some unique characteristics of urban perception tasks we need to consider if we want to develop a \lpm~ for GeoAI. 

Although street view images are like the natural images used in common vision-language tasks, one major difference is that common vision-language tasks usually focus on factual knowledge in images (e.g., ``\textit{how many cars in this image}'') while urban perception tasks are usually related to high-level human perception of the images such as the safety, poverty, beauty, and sound intensity of a neighborhood given a street view image. Compared with factual knowledge, this kind of high-level perception knowledge is rather hard to estimate and the labels are rather rare. Moreover, many perception concepts are vague and subjective which increases the difficulties of those tasks. So in order to develop a GeoAI {\lpm} that can achieve state-of-the-art performances on various urban perception tasks, we need to conduct some domain studies to provide a concrete definition of each urban perception concept and develop some annotated datasets for GeoAI \lpm~ pre-training.

\subsection{Remote Sensing} \label{sec:rs}

With the advancement of computer vision technology, 
deep vision models have been successfully applied to
different kinds of remote sensing (RS) tasks including image classification/regression \cite{ayush2021geography,rolf2021generalizable}, land cover classification \cite{ayush2021geography}, and object detection\cite{lam2018xview}.
Unlike the usual vision tasks which usually work on RGB images, RS tasks are based on multispectral/hyperspectral images from different sensors. Most existing RS works focus on training one model for a specific RS task using data from a specific sensor \cite{lam2018xview}. Researchers often compare performances of different models using the same training datasets and decide on model implementation based on accuracy statistics. 
However, we see the trend of {\lpm}s in the CV field such as CLIP \cite{radford2021clip}, Flamingo-9B \cite{Alayrac2022Flamingodeepmind} to be further developed to meet the unique challenges of remote sensing tasks.
RS experiments in Section \ref{sec:exp_rs} demonstrate that there is still a performance gap between current visual-language foundation models and task-specific deep models. To fill this gap and develop a GeoAI \lpm~ that can achieve state-of-the-art performances on various RS tasks, we need to consider the uniqueness of RS images and tasks.

Aside from being \textbf{task-agnostic}, the desiderata for a remote sensing {\lpm} include being:
1) 
\textbf{sensor-agnostic}: 
it can seamlessly reason among RS images from different sensors with different spatial or spectral resolutions; 
2) \textbf{spatiotemporally-aware}: it can handle the spatiotemporal metadata of RS images and perform geospatial reasoning for tasks such as image geolocalization and object tracking;
3) \textbf{environmentally-invariant}: it can decompose and isolate the spectral characteristics of the objects of interest across a variety of background environmental conditions and landscape structure. 
Recent developments here include
geography-aware RS models \cite{ayush2021geography} or self-supervised/unsupervised RS models \cite{ayush2021geography,rolf2021generalizable},
all of which are task-agnostic. 
However, we have yet to develop a {\lpm} for RS tasks which can satisfy all such properties.

In summary, efforts should be focused on developing GeoAI {\lpm}s using remote sensing to address pressing environmental challenges due to climate change. It would require complex models which look beyond image classification toward modeling ecosystem functions such as forest structure, carbon sequestration, urban heat, coastal flooding, and wetland health. Traditionally remote sensing is widely used to study these phenomena but in a site-specific and sensor-specific manner. Sensor-agnostic, spatiotemporally-aware, and environmentally-invariant {\lpm}s have the potential to transform our understanding of the trends and behavior of these complex environmental phenomena.

\subsection{Trajectory and Human Mobility} \label{sec:dis_human_mobility}

Trajectory, which is a sequence of time-ordered location tuples, is another important data type in GeoAI.  The proliferation of digital trajectory data generated from various sensors (e.g., smartphones, wearable devices, and vehicle on-board devices) together with the advancement of deep learning approaches has enabled novel GeoAI models for modeling human mobility patterns, which are crucial for city management and transportation services, etc. There are four typical tasks in modeling human dynamics with deep learning~\citep{luca2021survey}, including trajectory generation~\citep{rao2020lstm,choi2021trajgail}, origin-destination (OD) flow generation~\citep{yao2020spatial,simini2021deep}, in/out population flow prediction~\citep{li2021prediction,jiang2021deepcrowd}, and next-location/place prediction~\citep{rao2021privacy,lin2021pre}. 

In order to develop GeoAI {\lpm}s for supporting human mobility analysis, we need to consider the following perspectives: 1) pre-training and generation of task-agnostic trajectory embedding \cite{wang2021survey,musleh2022let}, which represent high-level movement semantics (e.g., spatiotempporal awareness, routes, and location sequence) from various kinds of trajectories~\citep{lin2021pre}; 2) context-aware contrastive learning of trajectory: human movements are constrained from their job type, surrounding built environment, and transportation infrastructure as well as many other spatiotemporal and environmental factors~\citep{sila2016analysis,wang2018context,luca2021survey}; GeoAI {\lpm}s should be able to link trajectories to various contextual representations such as road networks (e.g., Road2Vec~\citep{liu2017road2vec}, \cite{chen2021robust}), POI composition or land use types~\citep{zhang2021traj2vec}, urban morphology~\citep{chen2021classification}, and population distribution~\citep{huang2021sensing};  3) user geoprivacy~\citep{kessler2018geoprivacy} should be protected when training such GeoAI {\lpm}s since trajectory data can reveal individuals' sensitive locations such as home and personal trips. The privacy-preserving techniques by utilizing
cryptography or differential privacy~\citep{al2019privacy} and federated learning framework may be incorporated in the GeoAI {\lpm}s training process for trajectories~\citep{rao2021privacy}.

\vgap
\subsection{Geospatial Vector Data} \label{sec:vec}
Another critical challenge in developing 
{\lpm}s for GeoAI is the complexity of geospatial vector data which are commonly used in almost all GIS and mapping platforms. 
Examples include 
the US state-level and county-level dementia data (polygon data) discussed in Section \ref{sec:exp_alz}, urban POI data (point and polygon data) introduced in Section \ref{sec:exp_urban_func}, cartographic polyline data \cite{yu2022filling}, building footprints data \cite{mai2022polygon}, road networks (composed by points and polylines), and many others. 
In contrast with NLP and CV where text (1-D) or images (2-D) are well-structured and more suitable to  
common neural network architectures, vector data exhibits more complex data structures 
in the form of points, polylines, polygons, and networks \cite{mai2022review}.
So it is particularly challenging to develop a {\lpm} which can seamlessly encode or decode different kinds of vector data. 

Noticeably, 
recently developed location encoding  \cite{mai2020space2vec,mai2022review}, polyline encoding \cite{rao2020lstm,yu2022filling}, and polygon encoding techniques\cite{mai2022polygon} can be seen as a fundamental building block for such a model. Moreover, since encoding (e.g., geo-aware image classification\cite{mai2020space2vec}) or decoding (e.g., geoparsing \cite{wang2019EUPEG}) geospatial vector data, or conducting spatial reasoning (e.g., GeoQA \cite{mai2021geoqa}) is an indispensable component for most GeoAI tasks, developing {\lpm}s for vector data is the key step towards a multimodal {\lpm} for GeoAI.
This point also differentiates GeoAI {\lpm}s from existing {\lpm}s in other domains.

\vgap
\subsection{A Multimodal {\lpm} for GeoAI} \label{sec:multimodal_dis}
Except for those data modalities, 
there are also other datasets frequently studied in GeoAI such as geo-tagged videos, spatial social networks, sensor networks, and so on. 
Given all these diverse data modalities, the question is how to develop a multimodal {\lpm} for GeoAI that 
best integrates all of them.

When we take a look at the existing multimodal {\lpm}s such as 
CLIP \cite{radford2021clip},
DALL$\cdot$E2 \cite{ramesh2022dalle2}, 
MDETR \cite{kamath2021mdetr}, 
VATT \cite{akbari2021vatt}, 
BLIP \cite{li2022blip}, 
DeepMind Flamingo \cite{Alayrac2022Flamingodeepmind}, 
KOSMOS-1 \cite{huang2023kosmos1},
we can see the following general architecture:
1) \textbf{starting with separate embedding modules to encode different modalities of data} (e.g., a Transformer for texts and ViT for images \cite{radford2021clip}); 
2) (optionally) \textbf{mixing the representations} of different modalities 
by concatenation;
3)  (optionally) \textbf{more Transformer 
layers} for across modality reasoning, which can achieve a certain degree of alignment based on semantics, e.g., the word ``hospital'' attending to a picture of a hospital; 
4) \textbf{generative or discriminative prediction modules} for different modalities to achieve self-supervised training.

\begin{figure}[t!]
	\centering 
	\includegraphics[width=1.0\textwidth]{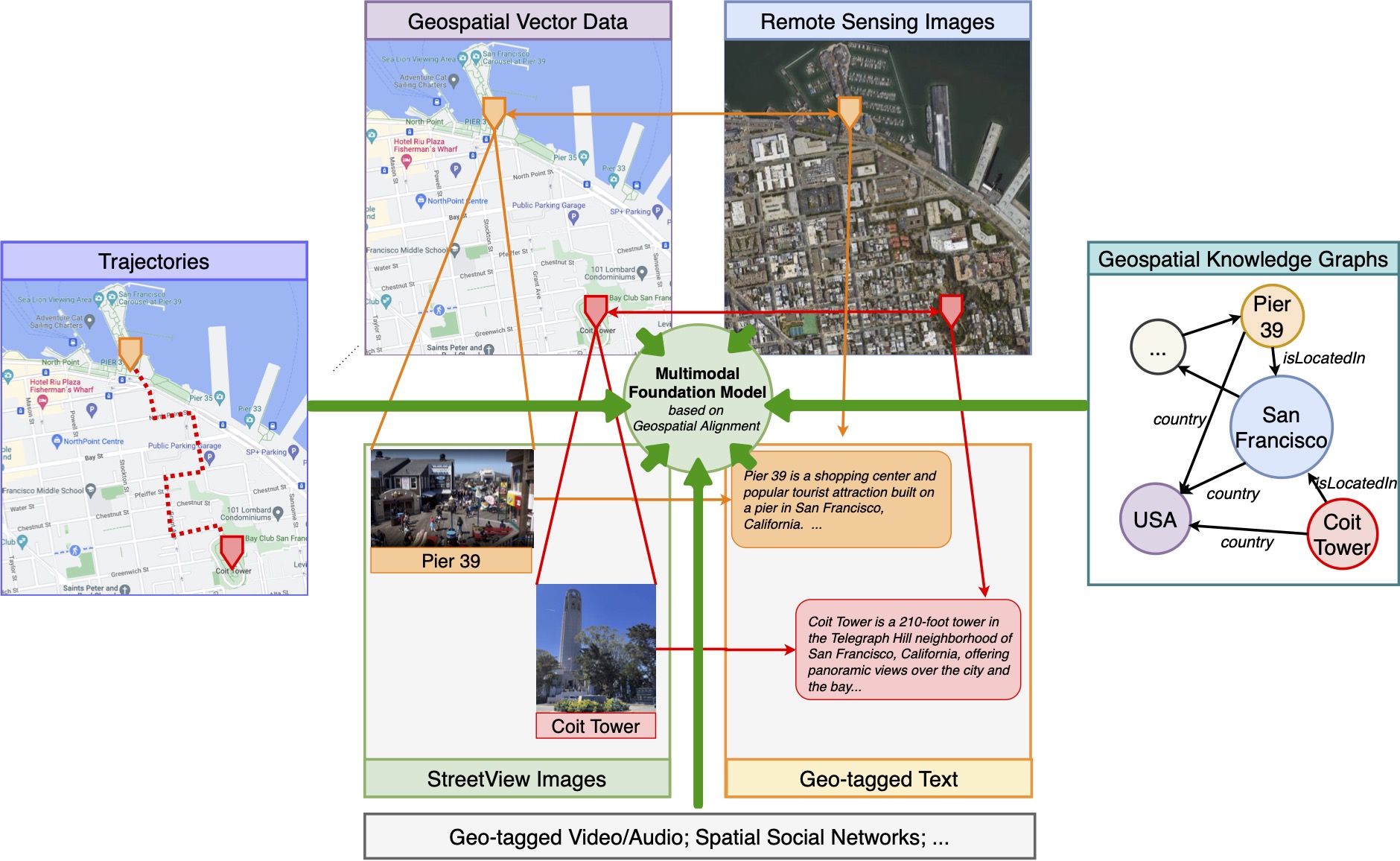}\vspace{-0.3cm}
	\caption{
	A multimodal 
	{\lpm} which 
	achieves alignment among different data sources via their geospatial relationships. 
	} 
	\vspace{-0.5cm}
	\label{fig:moltimodal_fm}
\end{figure}

One weak point of these architectures is the lack of integration with geospatial vector data, which is the backbone of spatial reasoning and helps alignment among multi-modalities in GeoAI. This is considered central and critical for GeoAI tasks. 
Therefore, we propose to replace step 2 with 
\textbf{aligning the representations} of different modalities (e.g., geo-tagged texts and RS images) by augmenting their representations with location encoding\cite{mai2020space2vec} before mixing them.
Figure \ref{fig:moltimodal_fm} illustrates this idea. 
Geo-tagged text data, 
street view images, remote sensing images, trajectories, and geospatial knowledge graphs can be easily aligned via their geographic footprints (vector data).
The key advantages of such a model are to enable spatial reasoning and knowledge transfer across modalities.

\vspace{-0.2cm}
\section{Risks and Challenges} \label{sec:risk}

Despite the recent progress, several challenges are emerging as more advanced {\lpm}s have been released~\cite{zhao2023survey}. First, as {\lpm}s continue to increase in size, there is a need to improve the computational efficiency for training and fine-tuning these models. Second, as an increasing number of {\llm}s are not open-sourced, it becomes challenging to incorporate knowledge into these models without accessing to their internal parameters. Third, as {\llm}s are increasingly deployed in remote third-party settings, protecting user privacy becomes increasingly important.

Except for these challenges for {\lpm}s in general, there are also many unique challenges and risks during the process of GeoAI {\lpm}s development.

\subsection{Geographic Fidelity} \label{sec:geo_fidelity}
\begin{figure*}
	\centering \tiny
	\vspace*{-0.2cm}
	\begin{subfigure}[b]{0.54\textwidth}  
		\centering 
		\includegraphics[width=\textwidth]{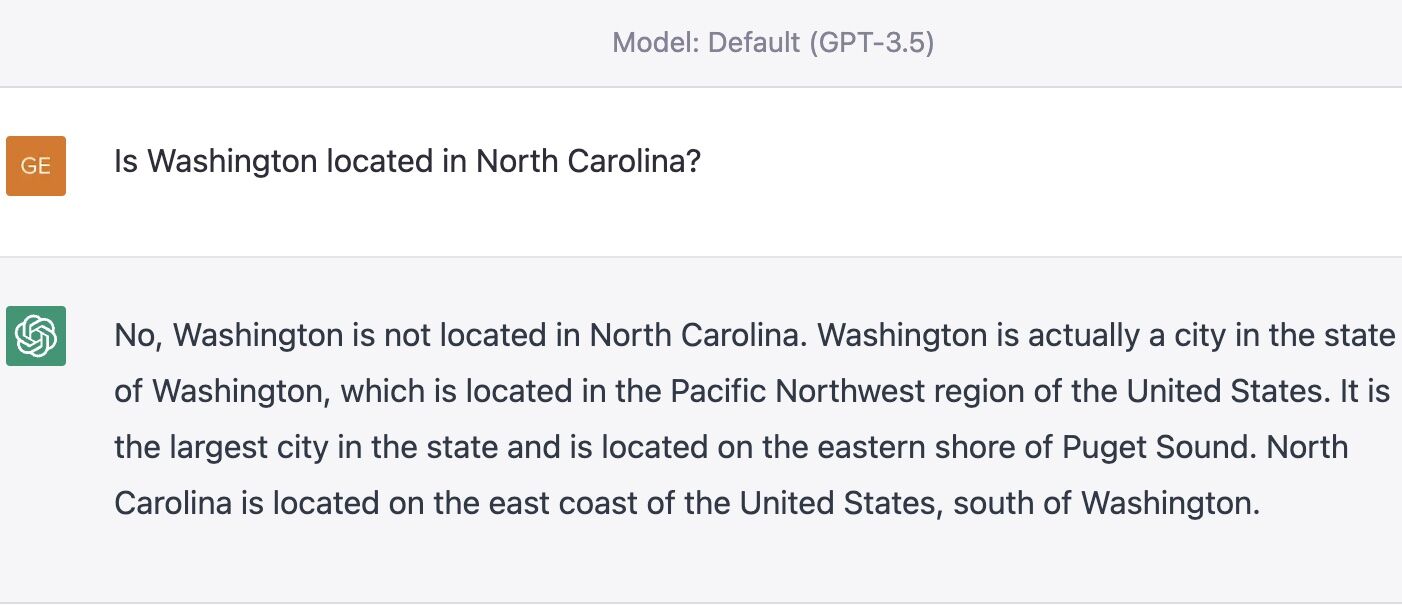}\vspace*{2cm}
		\caption[]%
		{{ 
		Geographically inaccurate results from \chatgpt
		}}    
		\label{fig:chatgpt_geoerror_washington}
	\end{subfigure}
        \begin{subfigure}[b]{0.45\textwidth}  
		\centering 
		\includegraphics[width=\textwidth]{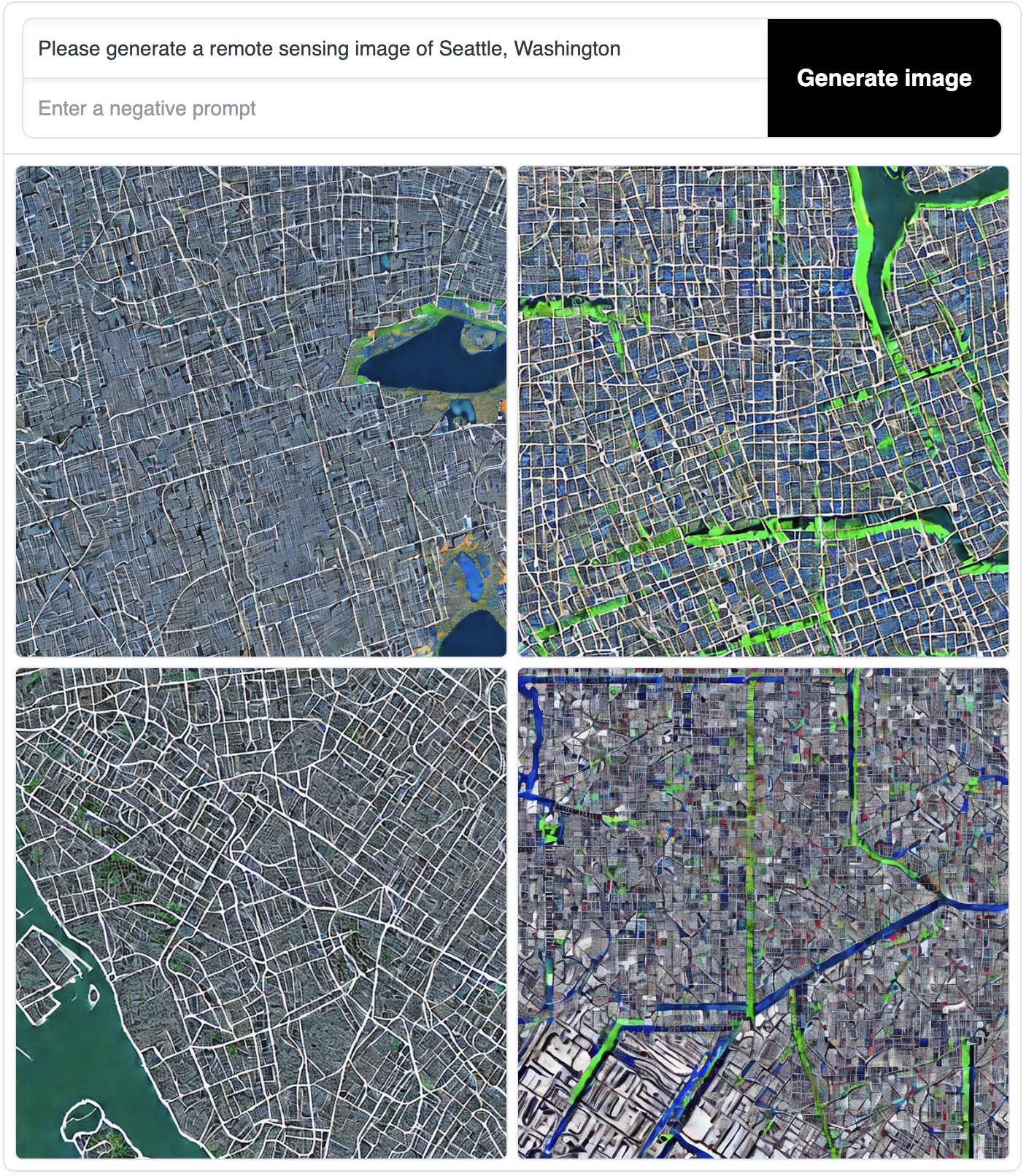}\vspace*{-0.2cm}
		\caption[]%
		{{ 
		Geographically inaccurate results from Stable Diffusion
		}}    
		\label{fig:stable_diffusion_geo_error}
	\end{subfigure}
        
	\caption{Some geographically inaccurate results generated from different language and vision foundation models.
    (a) The expected answer ``\textit{Washington, North Carolina}'' is not generated correctly. Moreover, there is no city in the state of Washington. The largest city in Washington State should be Seattle.
    (b) The generated remote sensing images from Stable Diffusion do not have correct geographic layouts such as road networks, waterbodies, etc.
 }
	\label{fig:geo_error}
    \vspace*{-0.15cm}
\end{figure*}

Many {\lpm}s are criticized for generating inaccurate and misleading results \cite{touvron2023llama,huang2023language}. In a geographic context, generating geographic faithful results is particularly important for almost all GeoAI tasks. In addition to Figure \ref{fig:chatgpt_error_topo} in Section \ref{sec:geo-text}, Figure \ref{fig:geo_error} illustrates two geographically inaccurate results generated from \chatgpt~ and Stable Diffusion. In Figure \ref{fig:chatgpt_geoerror_washington}, the expected answer should be ``\textit{Washington, North Carolina}''\footnote{\url{https://en.wikipedia.org/wiki/Washington,_North_Carolina}}. However, \chatgpt~ indicates there is no Washington in North Carolina. Moreover, the largest city in Washington State should be Seattle and there is no city in this state named Washington. Figure \ref{fig:stable_diffusion_geo_error} visualizes 4 generated remote sensing images generated by Stable Diffusion\footnote{\url{https://huggingface.co/spaces/stabilityai/stable-diffusion}}. Although those images appear similar to satellite images, it is rather easy to tell that they are fake RS images since the layouts of geographic features in these images are clearly not from any city in the world. In fact, generating faithful RS images is a popular and important RS task \cite{han2021spectral,he2021spatial} in which geometric accuracy is very important for the downstream tasks.

\vspace{-0.1cm}
\subsection{Geographic Bias} \label{sec:geo_bias}
\begin{figure*}
	\centering \tiny
	\vspace*{-0.2cm}
	\begin{subfigure}[b]{0.5\textwidth}  
		\centering 
		\includegraphics[width=\textwidth]{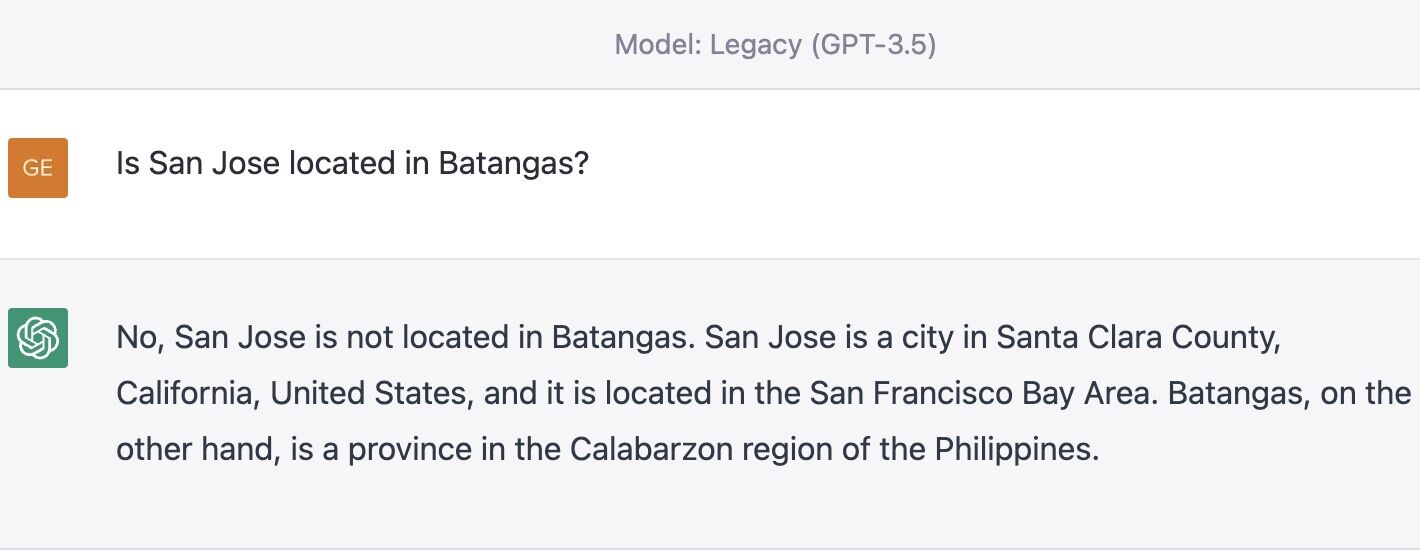} \vspace*{-0.2cm}
		\caption[]%
		{{ 
		Inaccurate results generated by \chatgpt~ due to geographic bias
		}}    
		\label{fig:chatgpt_geobias_san_jose}
	\end{subfigure}
 \hfill
        \begin{subfigure}[b]{0.49\textwidth}  
		\centering 
		\includegraphics[width=\textwidth]{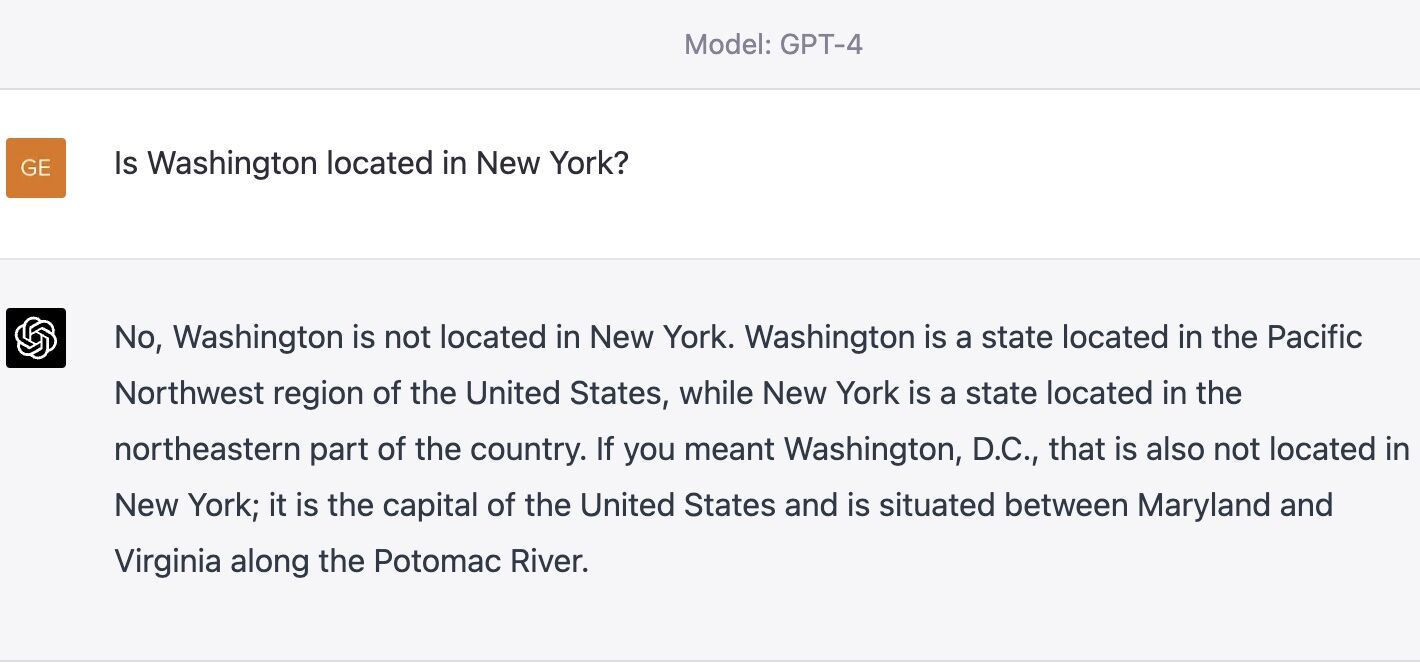}\vspace*{-0.2cm}
		\caption[]%
		{{ 
		Inaccurate results generated by GPT-4 due to geographic bias
		}}    
		\label{fig:gpt4_geobias_washington}
	\end{subfigure}
        
	\caption{Some inaccurate results generated from different \chatgpt~ and GPT-4 due to geographic bias.
    (a) \textit{San Jose, California, USA} is a more popular place name compared with \textit{San Jose, Batangas}. So \chatgpt~ interprets the name ``\textit{San Jose}'' incorrectly and leads to a wrong answer.
    (b) \textit{Washington State, USA} and \textit{Washington, D.C., USA} are two popular places with name ``\textit{Washington}''. The correct answer ``\textit{Washington, New York}'' is less popular which leads to an inaccurate answer.
 }
	\label{fig:geo_bias}
    \vspace*{-0.15cm}
\end{figure*}

It is well known that foundation models have the potential to amplify existing societal inequalities and biases present in the data \cite{bommasani2021opportunities,zhang2022opt,touvron2023llama}.
A key consideration for GeoAI in particular is 
\emph{geographic bias} \cite{liu2022geoparsing},
which is often overlooked by AI research. 
For example, Zilong et al. \cite{liu2022geoparsing} 
showed that all current geoparsers 
are highly geographically biased towards data-rich regions. 
The same issue can be observed in current {\llm}s. Figure \ref{fig:geo_bias} shows two examples in which both \chatgpt~ and GPT-4 generate inaccurate results due to the geographic bias inherited in these models. Compared with \textit{San Jose, California, USA}, \textit{San Jose, Batangas}\footnote{\url{https://en.wikipedia.org/wiki/San_Jose,\_Batangas}} is a less popular place name in many text corpus. Similarly, compared with \textit{Washington State, USA} and \textit{Washington, D.C., USA}, \textit{Washington, New York}\footnote{\url{https://en.wikipedia.org/wiki/Washington,\_New\_York}} is also a less popular place name. That is why both \chatgpt~ and GPT-4 interpret those place names incorrectly.
Compared to task-specific models, {\lpm}s suffer more from geographic bias since: 
1) the training data is collected in large-scale which is likely to be dominated by overrepresented communities or regions; 2) the huge number of learnable parameters and complex model structures make model interpretation and debiasing much more difficult; 3) the geographic bias of the {\lpm}s can be easily inherited by all the
adapted models downstream \cite{bommasani2021opportunities}, and thus bring much more harm to the society. 
This indicates a pressing need for designing proper (geographic) debiasing frameworks.

\subsection{Temporal Bias} \label{sec:time_bias}
\begin{figure*}
	\centering \tiny
	\vspace*{-0.2cm}
	\begin{subfigure}[b]{0.5\textwidth}  
		\centering 
		\includegraphics[width=\textwidth]{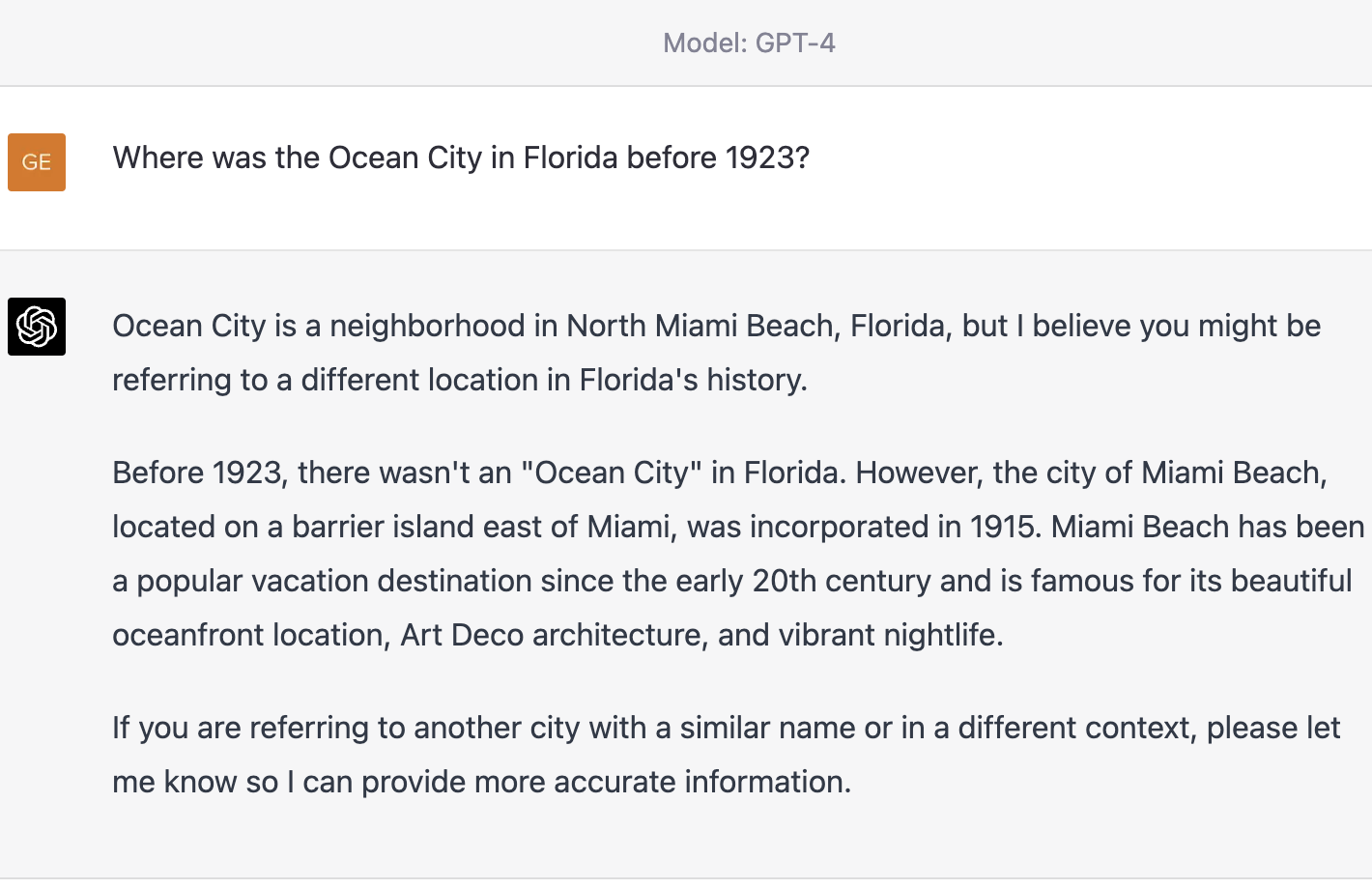} \vspace*{-0.2cm}
		\caption[]%
		{{ 
		Inaccurate results generated by \chatgpt~ due to temporal bias
		}}    
		\label{fig:gpt4_timebias_ocean}
	\end{subfigure}
        \begin{subfigure}[b]{0.49\textwidth}  
		\centering 
		\includegraphics[width=\textwidth]{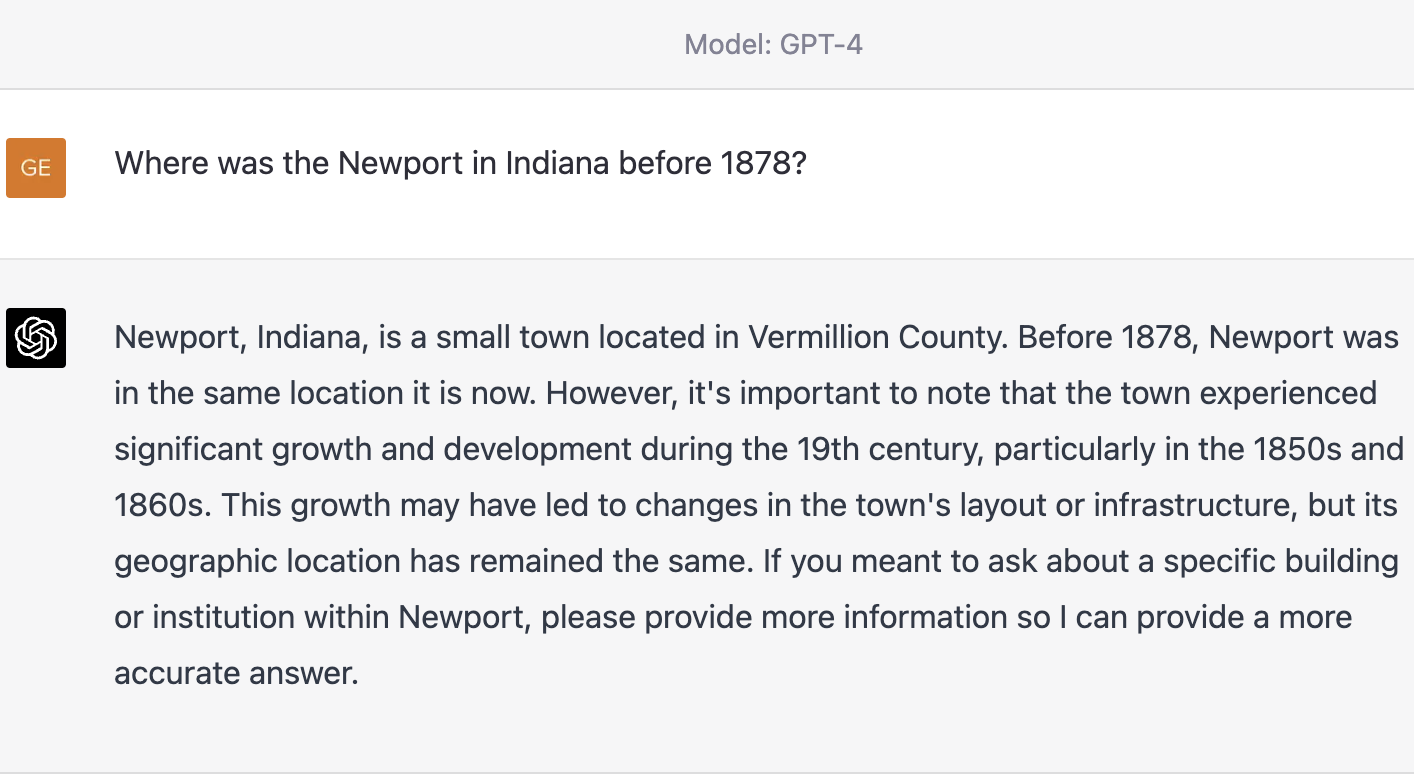}\vspace*{0.90cm}
		\caption[]%
		{{ 
		Inaccurate results generated by GPT-4 due to temporal bias
		}}    
		\label{fig:gpt4_timebias_newport}
	\end{subfigure}
        
	\caption{Some inaccurate results generated from GPT-4 due to temporal bias.
    (a) \textit{Flagler Beach, Florida} used to be named as \textit{Ocean City} during 1913 – 1923 while \textit{Ocean City, Florida} now is used to call another place in Florida. GPT-4 fails to recognize this and return a wrong answer. 
    (b) \textit{Fountain City, Indiana} was named by \textit{Newport} during 1834 - 1878 while now \textit{Newport} is used to call another city, \textit{Newport, Indiana} in \textit{Vermillion County}. GPT-4 fails to answer it correctly.
 }
	\label{fig:time_bias}
    \vspace*{-0.15cm}
\end{figure*}

Similar to geographic bias, {\lpm}s also suffer from temporal bias 
since there is much more training data available for current geographic entities than for historical ones
Temporal bias can also lead to inaccurate results. Two examples are shown in Figure \ref{fig:time_bias}. In both cases, the names of historical places are used for other places nearby. 
GPT-4 fails to answer both questions due to its heavy reliance on pre-training data which are biased towards current geographic knowledge. 
Temporal bias and geographic bias are critical challenges that need to be solved for the development of GeoAI {\lpm}s.

\vspace{-0.2cm}
\subsection{Spatial Scale} \label{sec:spa_scale}
Geographic information can be represented in different spatial scales, 
which means that 
the same geographic phenomenon/object can have completely different spatial representations (points vs. polygons) across GeoAI tasks.
For example, an urban traffic forecasting model must represent San Francisco (SF) as a complex polygon, while a geoparser usually represents SF as a single point.
Since {\lpm}s are developed for a diverse set of downstream tasks, they need to be able to 
handle geospatial information with different spatial scales, 
and infer the right spatial scale to use given a downstream task. 
Developing such a module is a critical component for an effective GeoAI FM.

\vspace{-0.2cm}
\subsection{Generalizability v.s. Spatial Heterogeneity} \label{geo_generalize}
An open problem for GeoAI is how to achieve model generalizability (``replicability'' \cite{goodchild2021replication}) across space while still allowing the model to capture spatial heterogeneity.
Given geospatial data with different spatial scales, we desire a FM that can learn general spatial trends while still memorizing location-specific details. 
Will this generalizability introduce unavoidable intrinsic model bias in downstream GeoAI tasks? Will this memorized localized information lead to an overly complicated prediction surface for a global prediction problem? With large-scale training data, this problem can be amplified and requires care .


\vspace{-0.2cm}
\section{Conclusion} \label{sec:conclude}
In this paper, we explore the promises and challenges for developing multimodal foundation models ({\lpm}s) for GeoAI. The potential of {\lpm}s is demonstrated by comparing the performance of existing {\llm}s and visual-language {\lpm}s as zero-shot or few-shot learners with fully-supervised task-specific SOTA models on seven tasks across multiple geospatial subdomains such as Geospatial Semantics, Health Geography, Urban Geography, and Remote Sensing. 
While in some language-only geospatial tasks, {\llm}s, as zero-shot or few-shot learners, can outperform task-specific fully-supervised models, existing {\lpm}s still underperform the task-specific fully-supervised models on  other geospatial tasks, especially tasks involving multiple data modalities (e.g., POI-based urban function classification, street view image-based urban noise intensity classification, and remote sensing image scene classification). 
We realize that the major challenge for developing a {\lpm} for GeoAI is the multimodality nature of geospatial tasks. After discussing the unique challenges of each geospatial data modality, 
we propose our vision for a novel multimodal {\lpm} for GeoAI that should be pre-trained based on the alignment among different data modalities via their geospatial relations. We conclude this work by discussing some unique challenges and risks for such a model.



\bibliographystyle{ACM-Reference-Format}


\newpage
\appendix

\newpage 
\appendix

\section{Appendix} \label{sec:app}

\subsection{The Full Prompts Used in Various Experiment} \label{sec:app_prompt}

\begin{minipage}[c]{0.48\textwidth}

	\begin{lstlisting}[
	style=prompt-style, 
	basicstyle=\scriptsize,
	linewidth=\textwidth,
	breaklines=true,
	captionpos=b, 
	caption={The prompt used by GPT-2 and GPT-3 models for typonym recognition on the test set of Hu2014 and Ju2016 dataset. One "Paragraph", "Q", and "A" tuple makes up one language instruction sample. \unexpanded{"[TEXT]"} will be replaced with the text to be annotated. We use in total 8 samples in this prompt while only 2 are shown here to save space. },
	label={ls:prompt-tr-all},
	frame=tbt
	]
This is a set of place name recognition problems
The `Paragraph` is a set of text containing places.
The goal is to infer which words represent named places in this paragraph, and split the named places with `;`
--
--
%*\colorbox{blueannoback}{Paragraph:}*) Alabama State Troopers say a Greenville man has died of his injuries after being hit by a pickup truck on Interstate 65 in Lowndes County.
%*\colorbox{greenannoback}{Q:}*) Which words in this paragraph represent named places?
%*\colorbox{redannoback}{A:}*) Alabama; Greenville; Lowndes
--
...
--
%*\colorbox{blueannoback}{Paragraph:}*) Settling in the Xenia area in 1856, the year after Bourbon County was organized in 1855, were John Van Syckle, Samuel Stephenson and Charles Anderson.
%*\colorbox{greenannoback}{Q:}*) Which words in this paragraph represent named places?
%*\colorbox{redannoback}{A:}*) Xenia; Bourbon
--
--
%*\colorbox{blueannoback}{Paragraph:}*) %*\colorbox{yellowannoback}{[TEXT]}*)
%*\colorbox{greenannoback}{Q:}*) Which words in this paragraph represent named places?
%*\colorbox{redannoback}{A:}*)
	\end{lstlisting}

\end{minipage}
\begin{minipage}[c]{0.04\textwidth}
\end{minipage}
\begin{minipage}[c]{0.48\textwidth}
	\begin{lstlisting}[
	style=prompt-style, 
	basicstyle=\scriptsize,
	linewidth=\textwidth,
	breaklines=true,
	captionpos=b, 
	caption={The prompt used by GPT-2 and GPT-3 models for local description recognition on the test set of HaveyTweet2017 dataset. One "Paragraph", "Q", and "A" tuple makes up one language instruction sample. \unexpanded{"[TEXT]"} will be replaced with the text to be annotated. We use in total 11 samples in this prompt while only 2 are shown here to save space. },
	label={ls:prompt-ldr-all},
	frame=tb
	]
This is a set of location description recognition problems
The `Paragraph` is a set of text containing location descriptions.
The goal is to infer which words represent location descriptions in this paragraph, and split different location descriptions with `;`.
--

--
%*\colorbox{blueannoback}{Paragraph:}*) Papa stranded in home. Water rising above waist. HELP 8111 Woodlyn Rd, 77028 #houstonflood
%*\colorbox{greenannoback}{Q:}*) Which words in this paragraph represent location descriptions?
%*\colorbox{redannoback}{A:}*) 8111 Woodlyn Rd, 77028
--
...
--
%*\colorbox{blueannoback}{Paragraph:}*) Major flooding at Clay Rd & Queenston in west Houston. Lots of rescues going on for ppl trapped.
%*\colorbox{greenannoback}{Q:}*) Which words in this paragraph represent location descriptions?
%*\colorbox{redannoback}{A:}*) Clay Rd & Queenston; west Houston
--
--
%*\colorbox{blueannoback}{Paragraph:}*) %*\colorbox{yellowannoback}{[TEXT]}*)
%*\colorbox{greenannoback}{Q:}*) Which words in this paragraph represent location descriptions?
%*\colorbox{redannoback}{A:}*)
	\end{lstlisting}
\end{minipage}

\begin{minipage}[c]{0.48\textwidth}
	\begin{lstlisting}[
	style=prompt-style, 
	basicstyle=\scriptsize,
	linewidth=\textwidth,
	breaklines=true,
	captionpos=b, 
	caption={The prompt used by GPT-3 for geoparsing on the test set of Ju2016 dataset. We use two samples as language instructions. 
	The yellow block indicate one text snippet in Ju2016 dataset and the orange block indicates the generated answers. The generated coordinates in the last line are treated as the geoparsing results.
	},
	label={ls:prompt-geoparse-all},
	frame=tb
	]
This is a set of geographical localization problems. 
The `Paragraph` is a set of text containing places.
The goal is to infer which words represent named places in this paragraph, and split the named places with `;`
Then, the next goal is to localize each named place as geographic coordinates with 5 decimal place precision.
--
--
%*\colorbox{blueannoback}{Paragraph:}*) Alabama State Troopers say a Greenville man has died of his injuries after being hit by a pickup truck on Interstate 65 in Lowndes County.
%*\colorbox{greenannoback}{Q:}*) Which words in this paragraph represent named places?
%*\colorbox{redannoback}{A:}*) Greenville; Alabama; Lowndes County

%*\colorbox{greenannoback}{Q:}*) What is the location of Greenville?
%*\colorbox{redannoback}{A:}*) 31.83283, -86.63270

%*\colorbox{greenannoback}{Q:}*) What is the location of Alabama?
%*\colorbox{redannoback}{A:}*) 32.92040, -86.83519

%*\colorbox{greenannoback}{Q:}*) What is the location of Lowndes County?
%*\colorbox{redannoback}{A:}*) 32.16314, -86.64631
--
--
%*\colorbox{blueannoback}{Paragraph:}*) San Jose was founded in 1803 when allotments of land were made to 45 men and two women by the Spanish government of New Mexico.
%*\colorbox{greenannoback}{Q:}*) Which words in this paragraph represent named places?
%*\colorbox{redannoback}{A:}*) San Jose; New Mexico

%*\colorbox{greenannoback}{Q:}*) What is the location of San Jose?
%*\colorbox{redannoback}{A:}*) 35.39728, -105.47501

%*\colorbox{greenannoback}{Q:}*) What is the location of New Mexico?
%*\colorbox{redannoback}{A:}*) 34.68965, -106.05006
--
--
Paragraph: %*\colorbox{yellowannoback}{the city of fairview had a population of 260 as of july 1, 2015. fairview ranks }*) %*\colorbox{yellowannoback}{in the lowerquartile for diversity index when compared to the other cities, towns }*) %*\colorbox{yellowannoback}{and census designated places (cdps) in .}*)
%*\colorbox{greenannoback}{Q:}*) Which words in this paragraph represent named places?
%*\colorbox{redannoback}{A:}*) %*\colorbox{orangeannoback}{Fairview}*)

%*\colorbox{orangeannoback}{Q: What is the location of Fairview?}*)
%*\colorbox{orangeannoback}{A: 41.85003, -87.65005}*)
	\end{lstlisting}
\end{minipage}

\end{document}